\documentclass[acmtog]{acmart}
\acmSubmissionID{544}

\usepackage{booktabs} %

\usepackage{multirow}
\usepackage{makecell}
\usepackage{amsmath}

\usepackage{colortbl}
\usepackage{graphicx}
\usepackage{subcaption}
\usepackage{lipsum}
\usepackage{arydshln}

\newcommand{\best}{\cellcolor{tablered}}
\newcommand{\sbest}{\cellcolor{orange}}
\newcommand{\tbest}{\cellcolor{yellow}}

\newcommand{\bc}{\mathbf{c}}

\newcommand{\bn}{\mathbf{n}}\newcommand{\bN}{\mathbf{N}}
\newcommand{\bo}{\mathbf{o}}\newcommand{\bO}{\mathbf{O}}
\newcommand{\bp}{\mathbf{p}}

\newcommand{\br}{\mathbf{r}}\newcommand{\bR}{\mathbf{R}}
\newcommand{\bS}{\mathbf{S}}

\newcommand{\bx}{\mathbf{x}}

\newcommand{\bSigma}{\boldsymbol{\Sigma}}

\newcommand{\cE}{\mathcal{E}}

\newcommand{\cG}{\mathcal{G}}

\newcommand{\secref}[1]{Section~\ref{#1}}

\newcommand{\RRR}{\mathbb{R}}

\makeatletter
\DeclareRobustCommand\onedot{\futurelet\@let@token\@onedot}
\def\@onedot{\ifx\@let@token.\else.\null\fi\xspace}

\makeatother

\definecolor{yellow}{rgb}{1, 1, 0.7}
\definecolor{orange}{rgb}{1, 0.85, 0.7}
\definecolor{tablered}{rgb}{1, 0.7, 0.7}
\definecolor{red}{rgb}{1, 0, 0}

\definecolor{wincolor}{rgb}{0.85, 0.0, 0.0}

\definecolor{darkyellow}{rgb}{0.8, 0.8, 0.5}
\definecolor{darkred}{rgb}{0.7, 0.3, 0.3}
\definecolor{darkgreen}{rgb}{0.3, 0.7, 0.3}
\definecolor{green}{rgb}{0, 1.0, 0}
\definecolor{pink}{rgb}{1, 0.4, 0.7}

\definecolor{realred}{rgb}{0.95, 0.1, 0.0}

\newcommand{\revise}[1]{{#1}}
\newcommand{\update}[1]{{#1}}

\newcommand{\boldparagraph}[1]{\vspace{0.1cm}\noindent{\bf #1:}}

\usepackage{xcolor}
\usepackage{wrapfig}

\usepackage{algorithm}
\usepackage{algpseudocode}

\acmJournal{TOG}

\begin{document}

\setcopyright{acmlicensed}
\acmJournal{TOG}
\acmYear{2024} \acmVolume{43} \acmNumber{6} \acmArticle{} \acmMonth{12}\acmDOI{10.1145/3687937}

\title{Gaussian Opacity Fields: Efficient Adaptive Surface Reconstruction in Unbounded Scenes}

\author{Zehao Yu}
\affiliation{
\institution{University of Tübingen, Tübingen AI Center}
\country{Germany}
}

\author{Torsten Sattler}
\affiliation{
\institution{Czech Technical University in Prague}
\country{Czech Republic}
}

\author{Andreas Geiger}
\affiliation{
\institution{University of Tübingen, Tübingen AI Center}
\country{Germany}
}

\begin{abstract}
Recently, 3D Gaussian Splatting (3DGS) has demonstrated impressive novel view synthesis results, while allowing the rendering of high-resolution images in real-time. However, leveraging 3D Gaussians for surface reconstruction poses significant challenges due to the explicit and disconnected nature of 3D Gaussians. In this work, we present Gaussian Opacity Fields (GOF), a novel approach for efficient, high-quality, and adaptive surface reconstruction in unbounded scenes. Our GOF is derived from ray-tracing-based volume rendering of 3D Gaussians, enabling direct geometry extraction from 3D Gaussians by identifying its levelset, without resorting to Poisson reconstruction or TSDF fusion as in previous work. We approximate the surface normal of Gaussians as the normal of the ray-Gaussian intersection plane, enabling the application of regularization that significantly enhances geometry. Furthermore, we develop an efficient geometry extraction method utilizing Marching Tetrahedra, where the tetrahedral grids are induced from 3D Gaussians and thus adapt to the scene's complexity. Our evaluations reveal that GOF surpasses existing 3DGS-based methods in surface reconstruction and novel view synthesis. Further, it compares favorably to or even outperforms, neural implicit methods in both quality and speed.
\end{abstract}



\begin{teaserfigure}
\begin{flushleft}
  \vspace{-0.2cm}
  {\large \textcolor{magenta}{\texttt{\href{https://niujinshuchong.github.io/gaussian-opacity-fields}{https://niujinshuchong.github.io/gaussian-opacity-fields}}}}\\
  \vspace{0.15cm}
    \centering
    \includegraphics[width=1.0\linewidth]{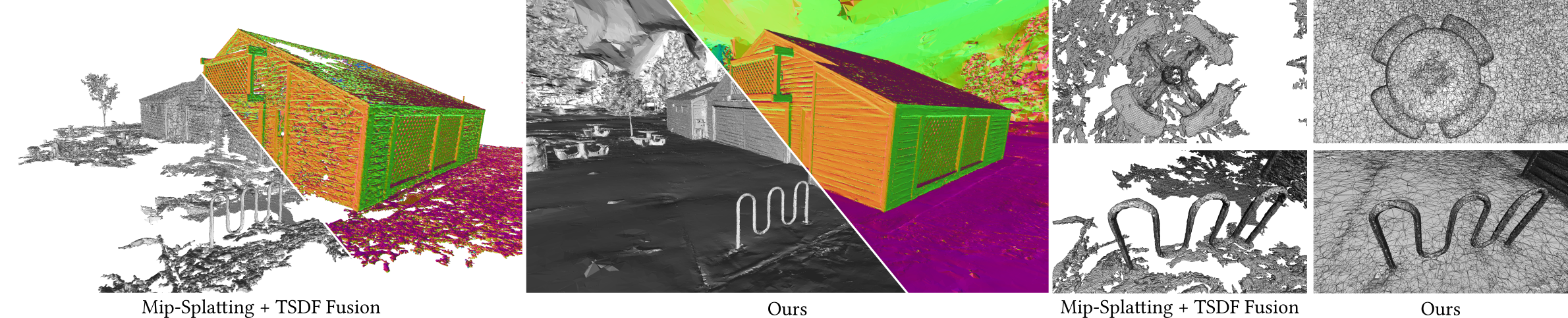}
    \caption{Applying TSDF fusion with rendered depth maps from the state-of-the-art Mip-Splatting~\cite{Yu2023MipSplatting} models results in noisy and incomplete meshes, while meshes extracted with our method are complete, smooth, and detailed. This is achieved by establishing Gaussian opacity fields from 3D Gaussians, which enables geometry extraction by directly identifying its level-set. Moreover, we generate tetrahedral meshes from 3D Gaussians and utilize Marching Tetrahedra to extract adaptive and compact meshes.}
    \label{fig:teaser}
\end{flushleft}
\end{teaserfigure}

\maketitle

\section{Introduction}
3D Reconstruction from multi-view images has been a long-standing goal in computer vision, with various applications in robotics, graphics, animation, virtual reality, and more. Since Neural Radiance Field (NeRF)~\cite{mildenhall2020nerf} demonstrated impressive novel view synthesis (NVS) results with implicit representations~\cite{Mescheder2019CVPR,Park_2019_CVPR} and volume rendering~\cite{drebin1988volume,levoy1990efficient,kajiya1984ray}, it has been extended to surface reconstruction with occupancy networks~\cite{Oechsle2021ICCV} and Signed Distance Functions (SDF)~\cite{wang2021neus,yariv2021volume}. While recent advancements~\cite{Yu2022MonoSDF,li2023neuralangelo,yariv2023bakedsdf,Yu2022SDFStudio} have shown impressive reconstruction results, these methods are mostly limited to reconstructing foreground objects~\cite{rosu2023permutosdf} and computationally expensive to optimize~\cite{yariv2023bakedsdf}. For instance, Neuralangelo~\cite{li2023neuralangelo} models the background separately with NeRFs and necessitates approximately 128 GPU hours to reconstruct a single scene. 

Another research avenue focuses on directly extracting surfaces from NeRF's opacity field for real-time rendering~\cite{chen2023mobilenerf,tang2022nerf2mesh,Rakotosaona2023THREEDV,Reiser2024ARXIV}. These methods employ a modular workflow including opacity field training, mesh extraction, simplification, and refinement. 
Particularly noteworthy is Binary Opacity Grids (BOG)\cite{Reiser2024ARXIV}, which excels at capturing intricate details in unbounded scenes through super-sampling. To extract detailed surfaces, \revise{it renders depth maps to generate a sparse high-resolution voxel grid and applies a heuristic fusion technique to label the voxels as inside or outside}. Marching cube algorithm~\cite{lorensen1998marching} is applied to extract a high-resolution mesh with hundreds of millions of points and thousands of millions of triangles, which is then simplified using slow post-processing techniques~\cite{garland1997surface}. Notably, as it focuses on NVS rather than surface reconstruction, the extracted meshes are noisy and contain fewer details in the background region, probably due to lack of regularization and contracted space~\cite{barron2022mipnerf360} fusion. 

More recently, 3D Gaussian Splatting (3DGS)~\cite{kerbl3Dgaussians} represents complex scenes as a set of 3D Gaussians, demonstrating photorealistic NVS results while trained efficiently and rendered in real-time. It has been quickly extended to surface reconstruction~\cite{guedon2023sugar,Huang2024ARXIV,Yu2024GSDF,chen2023neusg}. Notably, SuGaR~\cite{guedon2023sugar} regularizes the 3D Gaussians to align with surfaces and employs Poisson surface reconstruction~\cite{kazhdan2013screened} to extract a mesh from rendered depth maps. 2D Gaussian Splatting (2DGS)~\cite{Huang2024ARXIV} uses 2D Gaussians instead of 3D Gaussians as a scene representation for better surface representation and utilizes TSDF fusion to reconstruct a mesh. While these methods have shown improved reconstruction, they struggle with extracting fine-grained geometry~\cite{guedon2023sugar} and reconstructing background regions~\cite{Huang2024ARXIV}. A primary challenge is the inconsistency between mesh extraction and volume rendering during training. Specifically, Poisson reconstruction ignores the opacity and scale of Gaussian primitives and rendered depth maps are not sufficiently reliable. Moreover, TSDF fusion struggles to accurately model thin structures and to reconstruct unbounded scenes. Resorting to high-resolution voxel grids for TSDF leads to the creation of large meshes, similar to BOG~\cite{Reiser2024ARXIV}, due to the lack of adaptivity in the grid resolution relative to the scene's geometric complexity.

\boldparagraph{Contributions} In this paper, we propose \textit{Gaussian Opacity Fields} (GOF), a novel approach to achieve efficient, high-quality, and adaptive surface reconstruction from 3D Gaussians directly. Our key insights are threefold: 
\textbf{First}, we establish a Gaussian opacity field from a set of 3D Gaussians. Specifically, unlike projection-based volume rendering, our method leverages an explicit ray-Gaussian intersection to determine a Gaussian's contribution during volume rendering. Our ray-tracing-inspired formula facilitates the evaluation of opacity values for any point along a ray. 
We then define the opacity of any 3D point as the minimal opacity among all training views that observed the point. \update{Taking the minimum opacity across all views achieves view independence, making the opacity field solely a function of position. }
Our GOF is consistent with volume rendering during training and enables surface extraction from 3D Gaussians by directly identifying a level set, without resorting to Poisson reconstruction or TSDF fusion.
\textbf{Second}, we approximate the surface normals of 3D Gaussians as the normals of intersection planes between the ray and Gaussians. This technique allows for the incorporation of regularizations~\cite{Huang2024ARXIV} during training, thus enhancing the fidelity of geometry reconstruction.
\textbf{Third}, we propose an efficient surface extraction technique based on tetrahedra-grids. Recognizing that 3D Gaussians effectively indicate potential surface locations, we focus opacity evaluations on these areas. In particular, we use the center and corners of 3D bounding boxes around the 3D Gaussian primitives as vertex sets for the tetrahedral mesh. Upon assessing the opacity at tetrahedral points, we utilize the Marching Tetrahedra algorithm for triangle mesh extraction. Given that our opacity fields challenge the assumption that opacity changes linearly, we further implement a binary search algorithm to accurately identify the opacity field's level set, substantially enhancing the quality of the resulting surfaces.

To demonstrate the effectiveness and efficiency of GOF, we carry out extensive experiments across three challenging datasets~\cite{jensen2014large,barron2022mipnerf360,Knapitsch2017}. Our results indicate that GOF not only matches but, in some cases, surpasses the performance of existing SDF-based methods, while being much faster. Moreover, GOF outperforms all other 3DGS-based methods in both surface reconstruction and novel view synthesis.

\section{Related work}
\subsection{Novel view synthesis}
NeRF~\cite{mildenhall2020nerf} utilizes a multi-layer perception (MLP) for scene representation, including geometry and view-dependent appearances. The MLP is optimized via a photometric loss through volume rendering~\cite{kajiya1984ray,drebin1988volume,levoy1990efficient,max1995optical}. Subsequent enhancements have focused on optimizing NeRF's training using feature-grid representations~\cite{muller2022instant,Chen2022ECCV,yu2022plenoxels,Sun2022CVPR,kulhanek2023tetranerf} and improving rendering speed via baking~\cite{Reiser2021ICCV,Hedman2021ICCV,yariv2023bakedsdf,reiser2023merf}. Moreover, NeRF has been adapted to address challenges in anti-aliasing~\cite{barron2022mip,Barron2023ICCV} and unbounded scene modeling~\cite{barron2022mipnerf360,kaizhang2020}. More recently, 3D Gaussian splatting~\cite{kerbl3Dgaussians} represents complex scenes with 3D Gaussians. It demonstrated impressive NVS results while being optimized efficiently and rendering high-resolution images in real-time. \revise{Subsequent works improved its rendering quality via anti-aliasing~\cite{Yu2023MipSplatting} or extended it to dynamic scenes modeling~\cite{zhou2024hugs}, and more~\cite{chen2024survey}}. In this work, we extend 3DGS for high-quality surface reconstruction through the development of Gaussian Opacity Fields. We further introduce an efficient tetrahedron grid-based mesh extraction algorithm to extract scene adaptive and compact meshes.

\subsection{3D reconstruction}
3D Reconstruction from multi-view images is a fundamental problem in computer vision. Multi-view stereo methods~\cite{schoenberger2016mvs,yao2018mvsnet,Yu_2020_fastmvsnet} often employ complex multi-stage pipelines that include feature matching, depth estimation, point clouds fusion, and ultimately, surface reconstruction from aggregated point clouds~\cite{kazhdan2013screened}. In contrast, neural implicit methods~\cite{Oechsle2021ICCV,wang2021neus,yariv2021volume} significantly simplify the pipeline by optimizing an implicit surface representation via volume rendering. After optimization, triangle meshes can be extracted easily with Marching Cubes~\cite{lorensen1998marching} at any resolution. Notable advancements have been made through the adoption of more expressive scene representations~\cite{li2023neuralangelo}, advanced training strategies~\cite{li2023neuralangelo}, and the integration of monocular priors~\cite{Yu2022MonoSDF}. Despite these advances, these methods are mostly limited in reconstructing foreground objects~\cite{rosu2023permutosdf} and are computationally expensive to optimize~\cite{li2023neuralangelo,yariv2023bakedsdf}. Furthermore, the use of high-resolution grids for capturing fine details often results in excessively large meshes. By contrast, we establish Gaussian Opacity Fields using 3DGS~\cite{kerbl3Dgaussians}, which facilitates fast training. We utilize Marching Tetrahedra~\cite{doi1991efficient,shen2021dmtet} to extract adaptive, compact, and detailed meshes in unbounded scenes.

\subsection{Surface Reconstruction with Gaussians}
Inspired by the impressive NVS performance of 3D Gaussian Splatting (3DGS)~\cite{kerbl3Dgaussians}, researchers have proposed to utilize 3D Gaussians for surface reconstruction. Recent efforts~\cite{chen2023neusg,Yu2024GSDF} have integrated 3D Gaussians with neural implicit surfaces, optimizing a Signed Distance Function (SDF) network and 3D Gaussians jointly. While these approaches mark some advancements, they also inherit the shortcomings associated with implicit surfaces as previously discussed. Other studies have focused on surface extraction from optimized Gaussian primitives through post-processing techniques~\cite{guedon2023sugar,Huang2024ARXIV,Dai2024GaussianSurfels,Turkulainen2024ARXIV}.
Notably, SuGaR~\cite{guedon2023sugar} and GaussianSurfels~\cite{Dai2024GaussianSurfels} adopt Poisson surface reconstruction~\cite{kazhdan2013screened} to extract meshes from rendered depth maps. Meanwhile, 2D Gaussian Splatting (2DGS)~\cite{Huang2024ARXIV} employs TSDF fusion for this purpose. Though these methods achieve improved reconstructions, they face challenges in capturing fine-grained geometry and adequately reconstructing background areas. %
In this work, we derive Gaussian Opacity Fields~\textit{directly} from 3D Gaussians. Our GOF is consistent with the volume rendering process for rendering RGB images. It enables direct surface extraction by identifying a level set, without resorting to Poisson reconstruction and TSDF fusion. Additionally, we propose a tetrahedron grid-based technique for mesh extraction, resulting in adaptive and detailed meshes.

\section{Method}
Given multiple posed and calibrated images, our goal is to reconstruct the 3D scene efficiently while allowing detailed and compact surface extraction and photorealistic novel view synthesis.
To this end, we first construct Gaussian Opacity Fields (GOF) from 3D Gaussians, enabling geometry extraction directly by identifying a level set, eliminating the need for Poisson reconstruction or TSDF fusion. Next, we extend two effective regularizers from 2DGS~\cite{Huang2024ARXIV} to our 3D Gaussians, improving reconstruction quality. Finally, we propose a novel tetrahedra-based method to extract detailed and compact meshes from GOFs with marching tetrahedra. 

\subsection{Modeling} 
\label{sec:modeling}
Similar to prior works~\cite{zwicker2001ewa,kerbl3Dgaussians}, we represent the scene with a set of 3D Gaussian primitives $\{\cG_k | k=1, \cdots, K\}$. The geometry of each 3D Gaussian $\cG_k$ is parameterized by center $\bp_k \in \RRR^{3}$, \update{scaling matrix $\bS_k \in \RRR^{3 \times 3}$}, and rotation $\bR_k \in \RRR^{3 \times 3}$ parameterized by a quaternion:
\begin{equation}
\cG_k(\bx) = e^{-\frac{1}{2} (\bx-\bp_k)^T \bSigma_k^{-1}(\bx-\bp_k)}
\end{equation}
where $\bSigma_k \in \RRR^{3\times3}$ is a covariance matrix defined as $\bSigma_k = \bR_k \bS_k\bS_k^T \bR_k^T$.

\boldparagraph{Ray Gaussian Intersection} Instead of projecting 3D Gaussians to 2D screen space and evaluating the Gaussian in 2D, we evaluate the contribution of a Gaussian to a ray with explicit ray-Gaussian intersection~\cite{keselman2022fuzzy,R3DG2023}.
As we will see in \secref{sec:gaussian_opacity_fields}, this crucially enables evaluation of opacity values of \textit{arbitrary} 3D points in contrast to the projection-based Gaussian splatting mechanism of \cite{kerbl3Dgaussians} where 3D information is lost during the 3D-to-2D projection step.
The ray-Gaussian intersection is defined as the point where the Gaussian reaches its maximum value along the ray. Specifically, given a camera center at $\bo \in \RRR^{3}$ and a ray direction $\br \in \RRR^{3}$, any point $\bx \in \RRR^{3}$ along the ray can be written as $\bx = \bo + t \br$, where $t$ is depth of the ray. We first transform the point $\bx$ to the local coordinate system of the 3D Gaussian and normalize the point by its scale:
\update{
\begin{align}
    \bo_g &= \bS^{-1}_k \, \bR_k (\bo -\bp_k) \\
    \label{eq:transform}
    \br_g &= \bS^{-1}_k \, \bR_k \br \\
    \bx_g &= \bo_g + t \br_g
\end{align}

In this local coordinate system, the Gaussian value at any point along the ray becomes a 1D Gaussian, which is described by:
\begin{equation}
\cG_k^{1D}(t) = e^{-\frac{1}{2} \bx_g^T\bx_g} = e^{-\frac{1}{2} (\br_g^T\br_g t^2 + 2\bo_g^T\br_g t + \bo_g^T\bo_g)}
\label{eq:one_d_gaussian}
\end{equation}
The maximum of Eq.~\ref{eq:one_d_gaussian} is achieved when the quadratic term %
reaches maximum. Therefore, the closed-form solution for Eq.~\ref{eq:one_d_gaussian} is}
\begin{equation}
t^* = -\frac{B}{A}
\label{eq:depth_t}
\end{equation}
where $A = \br_g^T\br_g$ and $B = \bo_g^T\br_g$. 
Note that our formula is equivalent to the one presented in~\cite{keselman2022fuzzy}, where the ray-Gaussian intersection is computed directly in world space. However, using the normalized Gaussian coordinate by transforming the ray offers a clearer geometric interpretation and simplifies defining the normal of the intersection plane, as we will show in Sec.~\ref{subsec:normal}.

Now, we define the contribution $\cE$ of a Gaussian $\cG_k$ for a given camera center $\bo$ and ray direction $\br$ as:
\begin{equation}
\cE(\cG_k, \bo, \br) =\cG_k^{1D}(t^*)
\end{equation}

\boldparagraph{Volume Rendering}
Similar to 3DGS~\cite{kerbl3Dgaussians}, the color of a camera ray is rendered via alpha blending according to the primitive's depth order $1,\dots,K$:
\begin{equation}
\label{eq:volume_rendering}
    \bc(\bo, \br) = \sum^K_{k=1} \bc_k\,\alpha_k\,\cE(\cG_k, \bo, \br) \prod_{j=1}^{k-1} (1 - \alpha_j\,\cE(\cG_j, \bo, \br))
\end{equation}
where $\bc_k$ is the view-dependent color, modeled with spherical harmonics, and $\alpha_k\in\left[0,1\right]$ is an additional parameter that influences the opacity of Gaussian $k$. To efficiently render an image, we utilize the same tile-based rendering process as in 3DGS~\cite{kerbl3Dgaussians}.

\subsection{Gaussian Opacity Fields}
\label{sec:gaussian_opacity_fields}
\begin{figure}[t]
    \centering
    \includegraphics[width=0.99\linewidth]{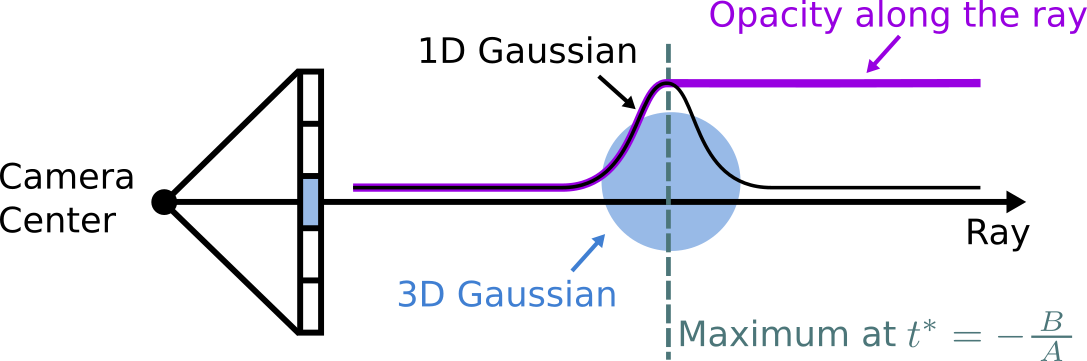}
    \caption{\textbf{Illustration of ray tracing volume rendering}. \revise{Evaluating a 3D Gaussian along a ray results in a 1D Gaussian, which has a closed-form solution for when it reaches maximal value. We define the \textit{opacity} along the ray as monotonously increasing until it reaches the maximal value and constant afterward.}
    }
    \label{fig:modeling}
\end{figure}

One significant benefit of using explicit ray-Gaussian intersection instead of projection is that it allows evaluating the opacity value or transmittance of \textit{any} 3D point $\bx$ along the ray. Let's first consider the case when there is only a single Gaussian $\cG_k$ along the ray. In this case, we define the opacity of any 3D point along the ray as
\[
\bO_k(\cG_k, \bo, \br, t) = 
\begin{cases} 
\cG^{1D}_k(t) \ & \text{if } t \leq t^* \\
\cG^{1D}_k(t^*) & \text{if } t > t^*
\end{cases}
\]
where $\bx = \bo + t \br$. Intuitively, the opacity (reverse of transmittance) increases until it reaches its maximal value, and remains constant afterwards as illustrated in Figure~\ref{fig:modeling}. Therefore, the opacity at any point along a ray given a set of Gaussians can be defined similar to the volume rendering process in Eq.~\ref{eq:volume_rendering} as:
\begin{equation}
    \bO(\bo, \br, t) = \sum^K_{k=1} \alpha_k\,\bO_k(\cG_k, \bo, \br, t) \prod_{j=1}^{k-1} (1 - \alpha_j\,\bO_j(\cG_j, \bo, \br, t))
\end{equation}
As a 3D point might be visible by \textit{any} training view, we define the opacity of a 3D point $\bx$ as the minimal opacity value among all training views or viewing directions:
\begin{equation}
    \bO(\bx) = \min_{(\bo, \br)} \bO(\bo, \br, t)
\end{equation}
We refer to $\bO(\bx)$ as the \textit{Gaussian Opacity Field} (GOF) since it is an opacity field derived from 3D Gaussians. Our GOF shares similarities to the visual hull~\cite{laurentini1994visual} or space carving~\cite{kutulakos2000theory}. But instead of using a silhouette where the opacity value of a ray (all points on the ray) is either 1 or 0, GOF uses volume rendering to evaluate opacity for each point from 3D Gaussians. 

Our GOF is consistent with the volume rendering process for RGB rendering during training. With GOF, we can extract surfaces directly by identifying their level sets, similar to UNISURF~\cite{Oechsle2021ICCV}, without resorting to Poisson reconstruction~\cite{guedon2023sugar} or TSDF fusion~\cite{Huang2024ARXIV}. We will discuss our method for efficient and adaptive mesh extraction in~\secref{sec:surface_extraction}. 

We also note that GOFs are a general formula as long as the scene representation is a set of 3D Gaussians. For example, we can use it to extract a mesh from a pre-trained 3DGS~\cite{kerbl3Dgaussians} or Mip-Splatting~\cite{Yu2023MipSplatting} model, where projection-based volume rendering is used for training, as we will show in the experiments. %

\subsection{Optimization}
Optimizing 3D Gaussians with pure photometric loss leads to noisy results as 3D reconstruction from multi-views is an underconstrained problem~\cite{barron2022mipnerf360,kaizhang2020}. Therefore, we extend the regularization terms in 2DGS~\cite{Huang2024ARXIV} to optimize our 3D Gaussians, including a depth distortion loss and a normal consistency loss.

\boldparagraph{Depth Distortion} 
We apply the depth distortion loss~\cite{Huang2024ARXIV}, \update{which is originally proposed in Mip-NeRF 360~\cite{barron2022mipnerf360},} to the ray-Gaussian intersection to concentrate Gaussians along the ray:
\begin{equation}
\mathcal{L}_{d} = \sum_{i,j}\omega_i\omega_j|t_i-t_j|
\end{equation}
where $i,j$ index over Gaussians contributed to the ray and $\omega_i = \alpha_k\,\cE(\cG_k, \bo, \br) \prod_{j=1}^{k-1} (1 - \alpha_j\,\cE(\cG_j, \bo, \br))$ is the blending weight of the $i-$th Gaussian and $t_i$ is the depth of the intersection point in Eq.~\ref{eq:depth_t}. 
However, the distortion loss minimizes both the distance between Gaussians and the weights of each Gaussian whereas minimizing the weights of the Gaussians could lead to an increase in alpha values for Gaussians that are blended first, which results in exaggerated Gaussians, resulting in floaters. Therefore, we detach the gradient of weights $\omega_i$ and only minimize the distance between Gaussians.

\boldparagraph{Normal Consistency} 
\label{subsec:normal}
A key challenge of applying 2DGS's normal consistency regularization~\cite{Huang2024ARXIV} to 3D Gaussians is that the gradient of 3D Gaussians always points outwards from the centers. \update{Consider a simple case when we render a single isotropic 3D Gaussian to image space. The rendered result is a 2D Gaussian in the image plane. The gradient of this 2D Gaussian always points outward from the projected 2D center, meaning the rendered normals at two different pixels will differ as long as the directions from the projected 2D Gaussian center to the pixel coordinates vary. Moreover, the normal at the projected 2D center is not well-defined. This ambiguity makes the optimization difficult.}

\begin{figure}[t]
    \centering
    \includegraphics[width=0.99\linewidth]{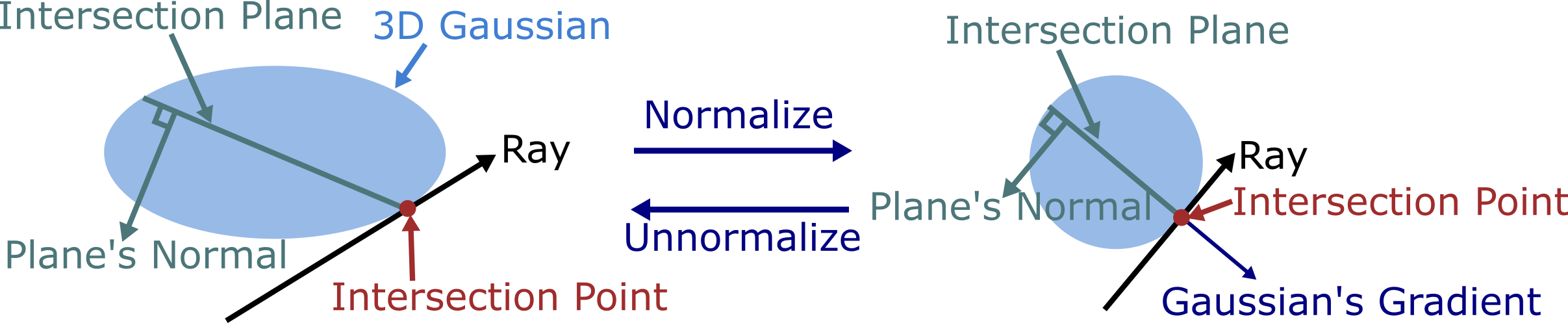}
    \caption{\update{\textbf{Definition of Gaussian's normal}. We approximate the Gaussian's normal as the normal of the ray-Gaussian intersection plane. First, we transform the ray into the Gaussian coordinate system and normalize it using the Gaussian's scales. In this normalized coordinate system, the ray-Gaussian intersection plane is perpendicular to the ray, making the normal the inverse of the ray direction. Finally, we transform the intersection plane back to world space by reversing the normalization process. %
    }}
    \label{fig:gaussian_normal}
\end{figure}

\newcommand{\densifywidth}{0.24\textwidth}

\begin{figure*}[t]
    \centering
    \setlength{\tabcolsep}{0.1em}
    \renewcommand{\arraystretch}{0.4}
    \scriptsize
    \begin{tabular}{cccc}
    \includegraphics[width=\densifywidth]{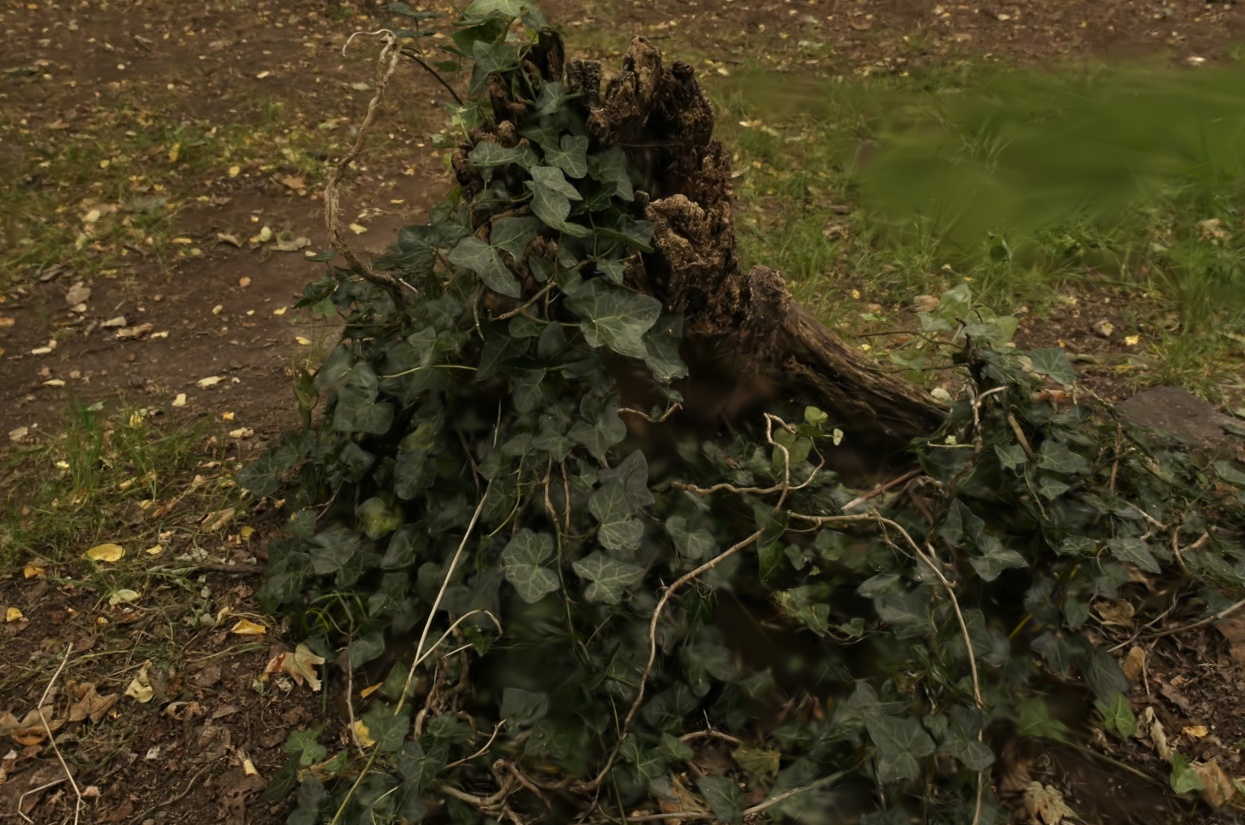}&    
    \includegraphics[width=\densifywidth]{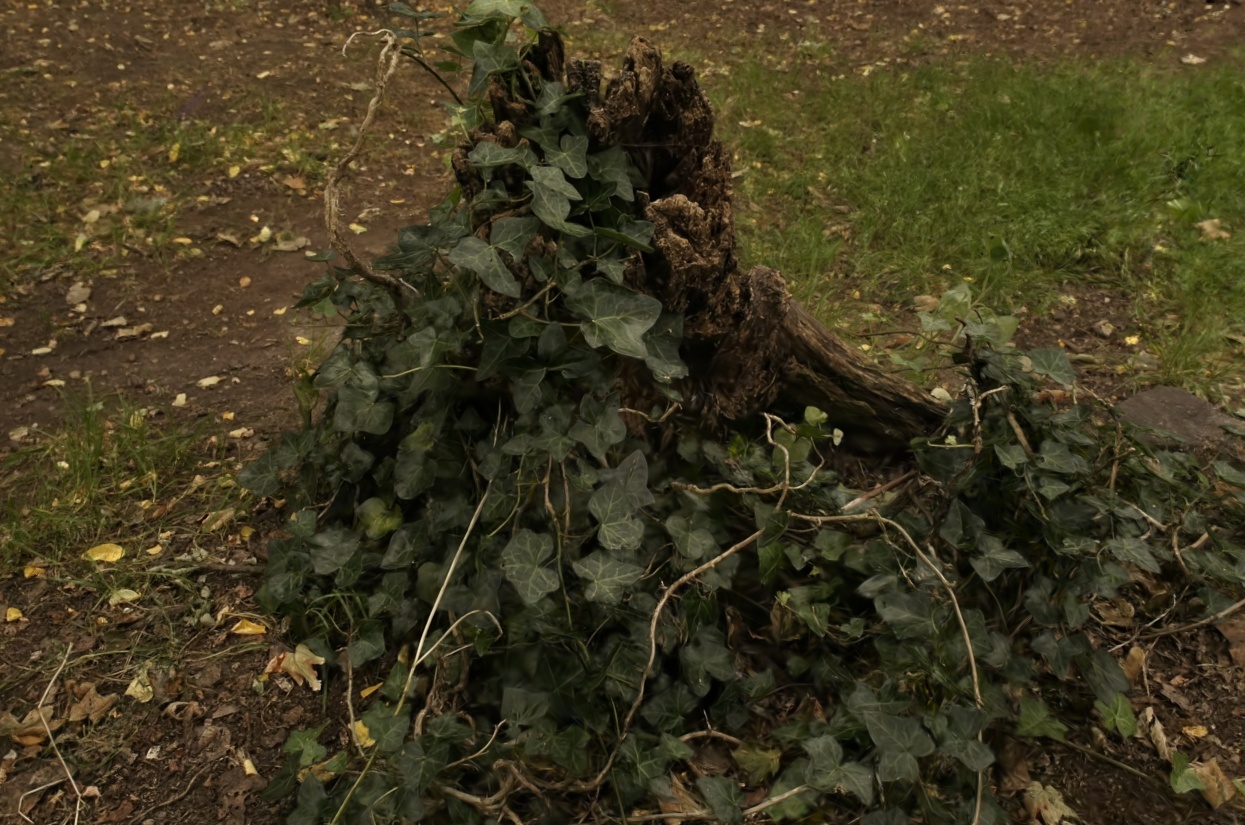}&    
    \includegraphics[width=\densifywidth]{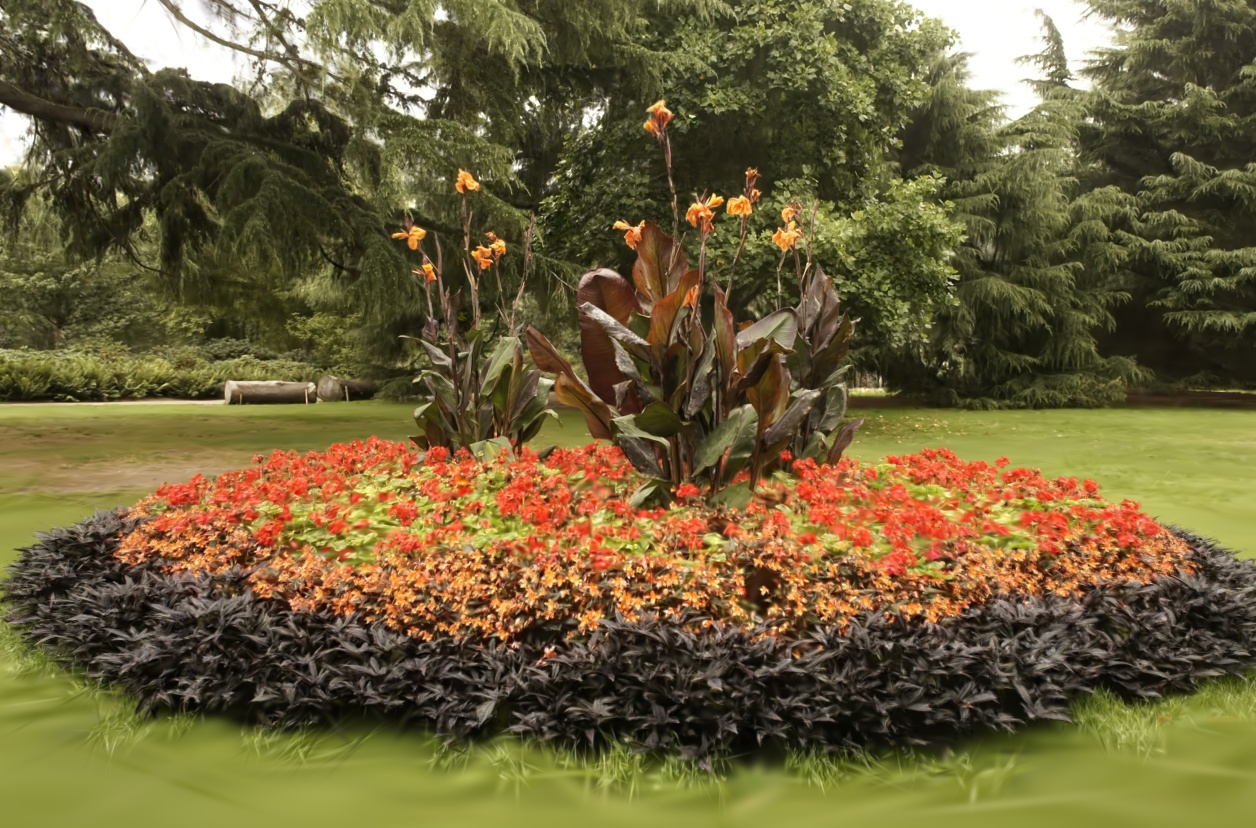}&    
    \includegraphics[width=\densifywidth]{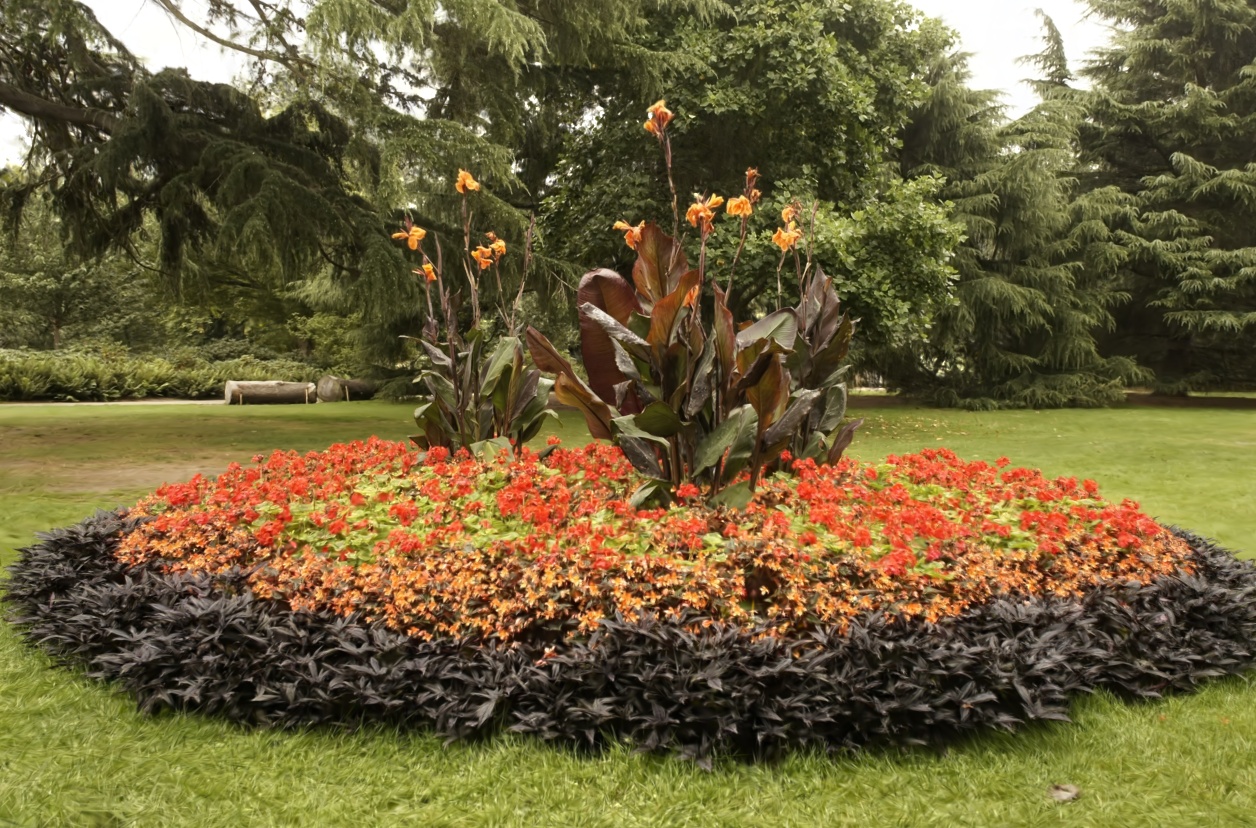} \\
    3DGS~\cite{kerbl3Dgaussians} & 3DGS + Our Densification & Mip-Splatting~\cite{Yu2023MipSplatting} & Mip-Splatting + Our Densification
    \end{tabular}
    \vspace{-0.1in}
    \caption{\textbf{Comparison of densification strategy on the Mip-NeRF 360 Dataset~\cite{barron2022mipnerf360}.} 
    Applying our densification to 3DGS~\cite{kerbl3Dgaussians} and Mip-Splatting~\cite{Yu2023MipSplatting} significant improves the NVS results. Please note that the glass regions are blur in both 3DGS and Mip-Splatting, while our method renders the image faithfully.
    }
    \label{fig:densify}
\end{figure*}

To mitigate this issue, we define the normal of a 3D Gaussian as the normal of the intersection plane given a ray direction. \update{An illustration is shown in Figure~\ref{fig:gaussian_normal}. Specifically, we begin by transforming the ray into the Gaussian coordinate system and normalizing it using the Gaussian's scales, as described in Eq.~\ref{eq:transform}. In the normalized coordinate system, the ray-Gaussian intersection plane is perpendicular to the ray. Therefore, the normal is the inverse ray direction $-\br_g$. Next, we reverse the normalization to transform the intersection plane back to world space, where the plane's normal is given by $-\bS^{-1}_k\br_g$. Finally, we apply the inverse of the world-to-Gaussian rotation to transform the normal back to world space, yielding the normal direction of the intersection plane as $\bn_i = -\bR^T_k \, \bS^{-1}_k\br_g $, which should then be normalized to ensure it is a valid unit vector. It is important to note that in world space, the intersection plane is not necessarily perpendicular to the ray direction, but the plane's normal is always perpendicular to the intersection plane.
}

After defining the normal of a Gaussian, we apply the depth-normal consistency regularisation: 
\begin{equation}
\mathcal{L}_{n} = \sum_{i} \omega_i (1-\bn_i^\top\bN)
\end{equation}
where $i$ indexes over intersected Gaussians along the ray, $\omega$ denotes the blending weight, and $\bN$ is the normal estimated by the gradient of the depth map~\cite{Huang2024ARXIV}.

\boldparagraph{Final Loss} Finally, we optimize our model from an initial sparse point cloud using multiple posed images with the following loss:
\begin{equation}
\mathcal{L} = \mathcal{L}_{c} + \alpha \mathcal{L}_{d}  + \beta  \mathcal{L}_{n}
\end{equation}
where $\mathcal{L}_c$ is an RGB reconstruction loss combining $\mathcal{L}_1$ with the D-SSIM term from \cite{kerbl3Dgaussians}, while $\mathcal{L}_{d}$ and $\mathcal{L}_{n}$ are regularization terms. Note that, we utilize the decoupled appearance modeling proposed in VastGaussian~\cite{lin2024vastgaussian} to model the uneven illumination for the Tanks and Temples dataset~\cite{Knapitsch2017}, where a small convolutional neural network is used to predict image-dependent colors such that the model will not fake inconsistent illumination with geometry. \update{While other variants could also be applicable, we use VastGaussian's solution due to its demonstrated improvements in large-scale scenarios and its ease of reimplementation.}

\update{
\boldparagraph{Improved Densification}
As the optimization starts from a sparse point cloud, it is necessary to increase the number of 3D Gaussians to better reconstruct the scene. We follow the densification strategy in 3DGS~\cite{kerbl3Dgaussians}. Specifically, the densification of a Gaussian (either by cloning or splitting) is guided by the magnitude of the view-space position gradient $\frac{dL}{d\bx}$, where $\bx$ is the center of projected Gaussian. Mathematically, $\frac{dL}{d\bx}$ sums over pixels $\bp_i$ that the Gaussian contributed to:
\begin{equation}
    \frac{dL}{d\bx} = \sum_{i} \frac{dL}{d\bp_i} \frac{d\bp_i}{d\bx}
\end{equation}
If the norm of the gradient $\|\frac{dL}{d\bx}\|_2$ is above a predefined threshold $\tau_\bx$, the Gaussian is chosen as the candidate for densification. 

However, we found that this metric is not effective in identifying overly blurred areas, due to its inability to distinguish between well-reconstructed regions and those where the gradient signals from different pixels negate each other, leading to minimal overall gradient magnitudes. Therefore, we propose a simple modification to the metric that accumulates the norms of the individual pixel gradients instead:
\begin{equation}
    M = \sum_{i} \|\frac{dL}{d\bp_i} \frac{d\bp_i}{d\bx}\|
\end{equation}
Our metric $M$ better indicates regions with significant reconstruction errors, resulting in better reconstruction and novel view synthesis results, as shown in Figure~\ref{fig:densify}.
}

\subsection{Surface Extraction}
\label{sec:surface_extraction}
Post-training, the conventional step towards surface or triangle mesh extraction involves densely evaluating the opacity values within regions of interest, a technique well-suited to simple scenarios like the DTU Dataset~\cite{jensen2014large}, as done in previous work~\cite{Oechsle2021ICCV,yariv2021volume,wang2021neus}. For a large-scale unbounded scene, some have adopted dense evaluation in a contracted space~\cite{yariv2023bakedsdf}. However, dense evaluation for grids incurs substantial computational costs due to the cubic growth of complexity with grid resolution. Unfortunately, capturing fine details necessitates high-resolution grids, leading to significant overhead. Alternative sparse grids may reduce dense evaluation but still result in huge meshes, often comprising hundreds of millions of points and billions of faces. Simplifying such large and complex meshes typically requires a slow post-processing step. For instance, mesh simplification in BOG~\cite{Reiser2024ARXIV} requires approximately 4 hours. To circumvent these challenges, we introduce a novel method for extracting adaptive and compact meshes using tetrahedral grids and marching tetrahedra. 

\boldparagraph{Tetrahedral Grids Generation}
Our primary insight is that the position and scale of 3D Gaussian primitives serve as reliable indicators for the presence of surfaces. To capitalize on this, \update{we define a 3D bounding box around each Gaussian primitive, where the extent of the 3D box is 3 times the Gaussian scales}. The 3D box's center has the highest opacity and its corners have the smallest opacity. \update{Note that we do not consider the opacity $\alpha$ of Gaussian primitives but it could be used to filter out Gaussians with low opacity value.} We create tetrahedral grids with the center and corners of 3D bounding boxes. Inspired by Tetra-NeRF~\cite{kulhanek2023tetranerf}, we employ the CGAL Library~\cite{cgal:pt-tds3-24a} for Delaunay triangulation to construct tetrahedral cells. The generated tetrahedral cell might connect points across significant distances. Therefore, we employ a filtering step for the tetrahedral cells, removing any cell whose edges connect non-overlapping Gaussians. Gaussians are considered non-overlapping when the length of the edge connecting them exceeds the sum of their maximum scales, i.e., maximum 3-sigma extent dimension. %

\boldparagraph{Efficient Opacity Evaluation}
To efficiently evaluate the opacity of the vertices of the tetrahedral grid, we design a tile-based evaluation algorithm, inspired by 3DGS~\cite{kerbl3Dgaussians}. Specifically, we first project the vertices to image space and identify the corresponding tiles for these projections. These points are then organized according to their tile ID. For each tile, we retrieve the list of points that are projected within it, project these points again to identify the pixel that it falls inside and identify the Gaussians that contribute to this pixel. Finally, we enumerate all points to evaluate their opacity based on the pre-filtered list of Gaussians. \update{Note that this process is iterated over all training images. Then we take the minimum opacity over all training images as the opacity for the vertices of the tetrahedral grid.} An overview of the algorithm is provided in the supplementary material.

\newcommand{\mcwidth}{0.23\textwidth}

\begin{figure}[t]
    \centering
    \setlength{\tabcolsep}{0.1em}
    \renewcommand{\arraystretch}{0.4}
    \scriptsize
    \begin{tabular}{cc}
    \includegraphics[width=\mcwidth]{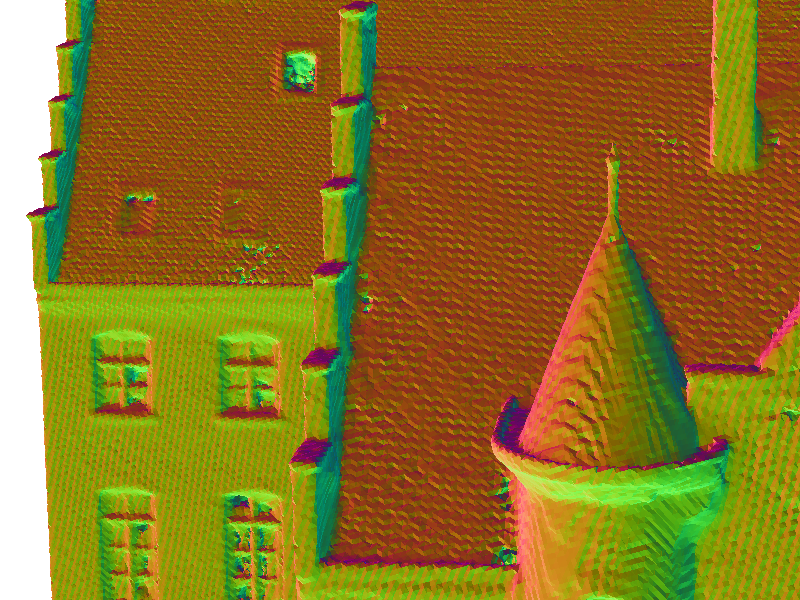}&    
    \includegraphics[width=\mcwidth]{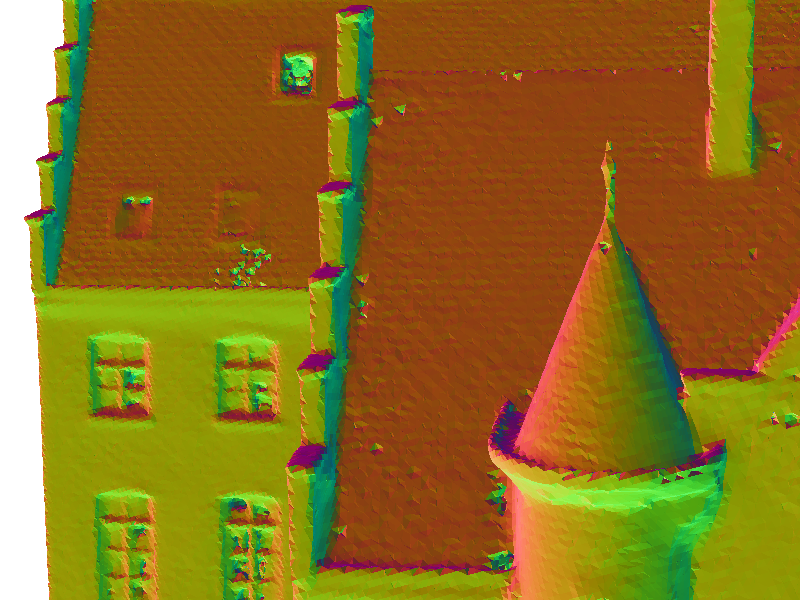} \\
    (a) Marching Cubes & (b) Marching Cubes with Binary Search 
    \end{tabular}
    \vspace{-0.1in}
    \caption{\textbf{Comparison of applying binary search to Marching cubes}. Strong step artifacts can be observed in Marching cubes results since the linear assumption does not hold in our GOF. Applying our binary search algorithm eliminates this artifact.
    }
    \label{fig:mc_binary}
\end{figure}
 
\newcommand{\tntwidth}{0.192\textwidth}

\begin{figure*}[t]
    \centering
    \setlength{\tabcolsep}{0.1em}
    \renewcommand{\arraystretch}{0.4}
    \scriptsize
    \begin{tabular}{ccccc}
    \includegraphics[width=\tntwidth]{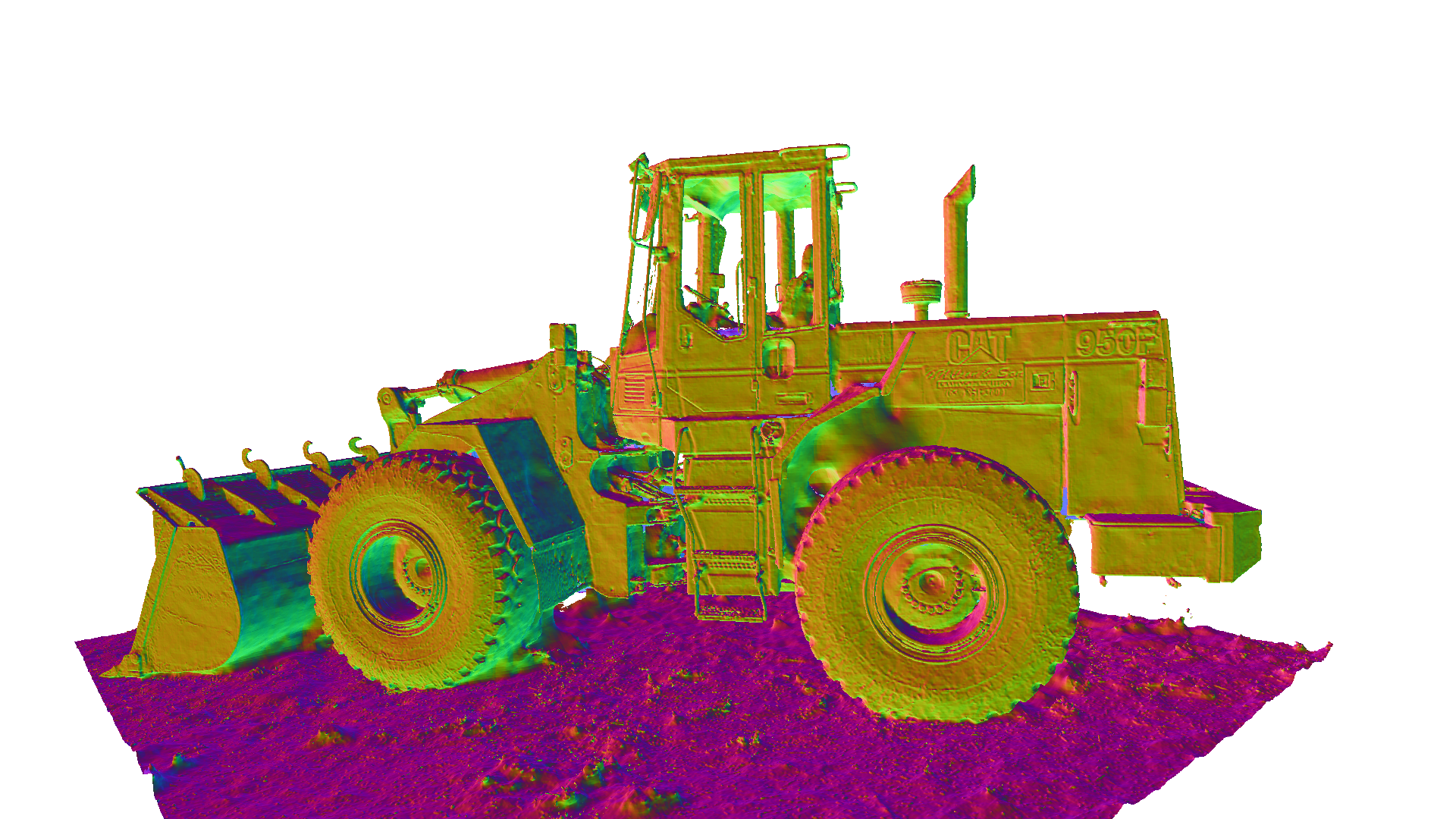}&  
    \includegraphics[width=\tntwidth]{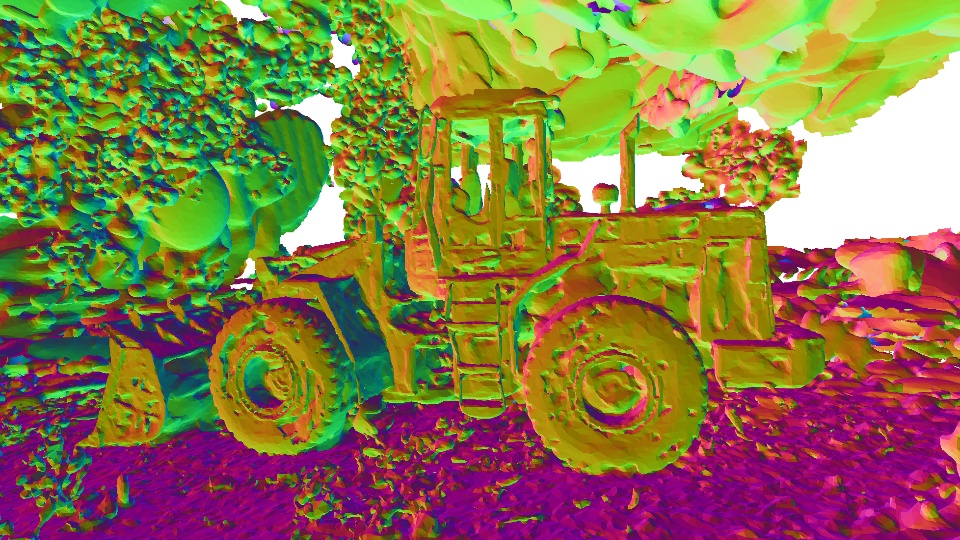}&    
    \includegraphics[width=\tntwidth]{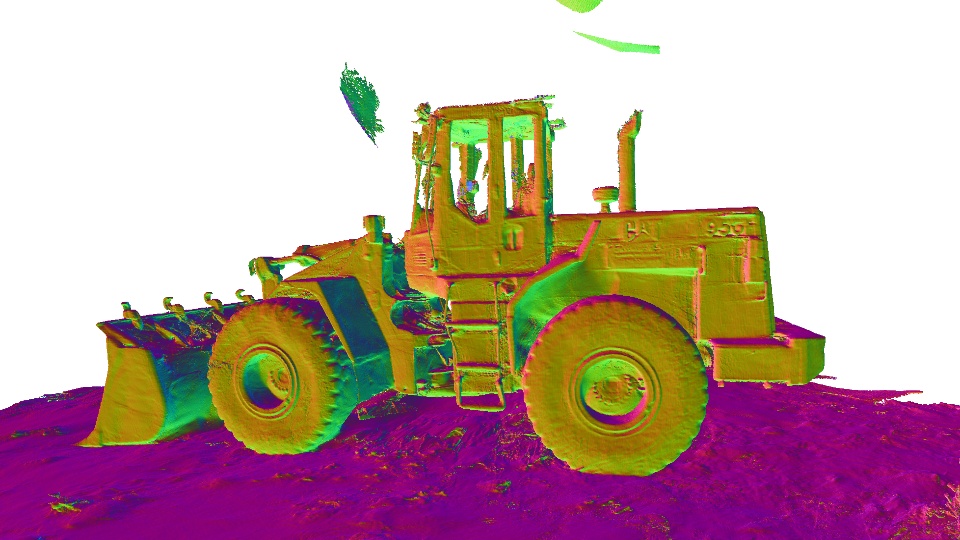}&    
    \includegraphics[width=\tntwidth]{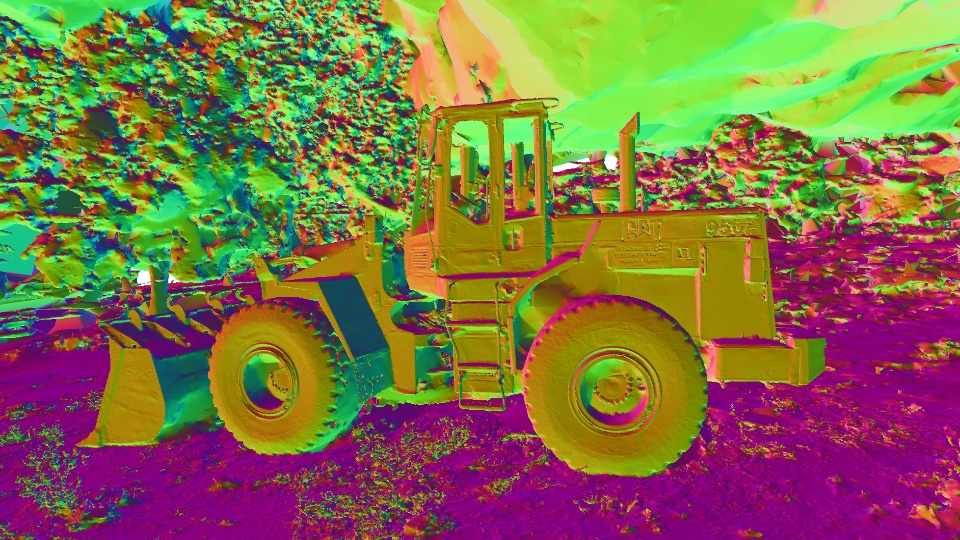}&  
    \includegraphics[width=\tntwidth]{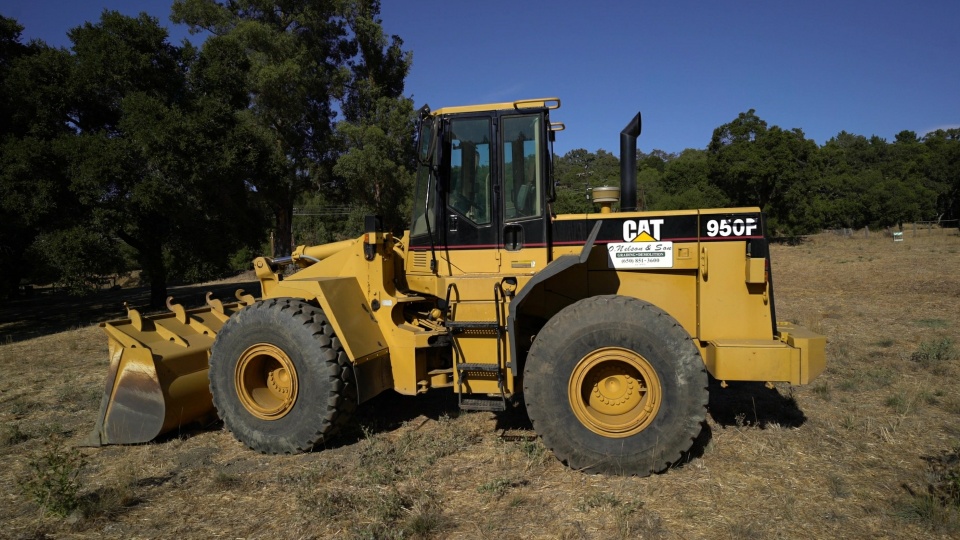} \\
    \includegraphics[width=\tntwidth]{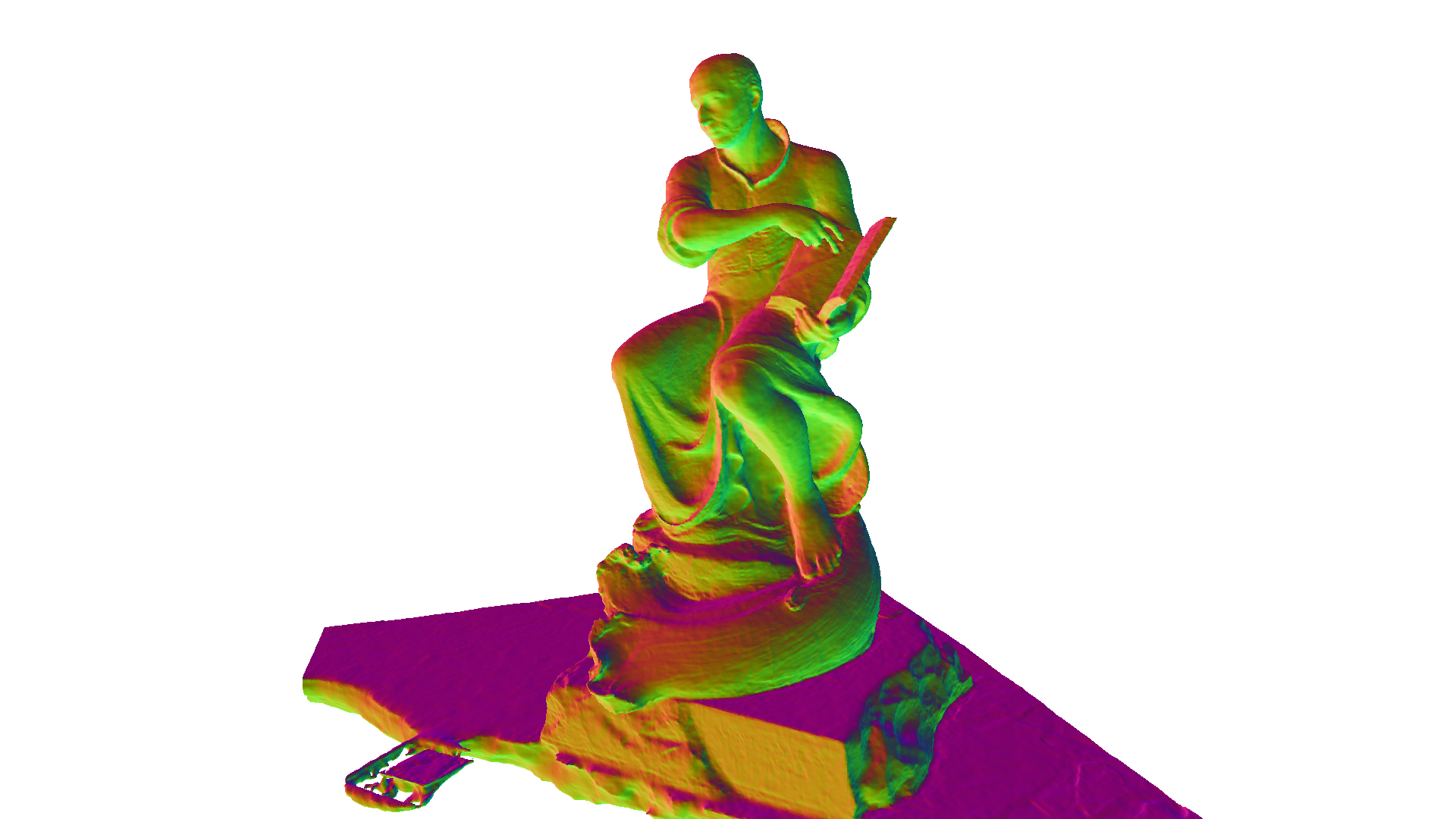}&    
    \includegraphics[width=\tntwidth]{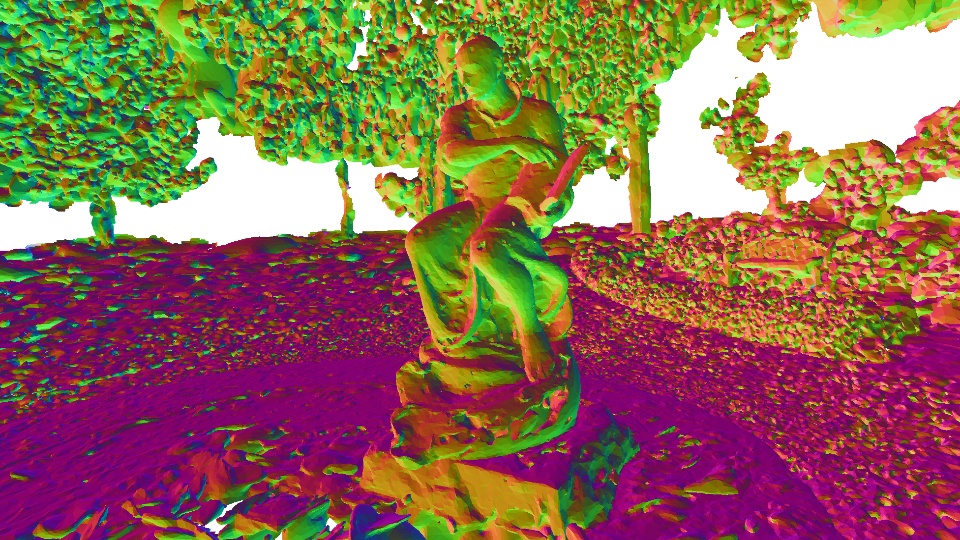}&    
    \includegraphics[width=\tntwidth]{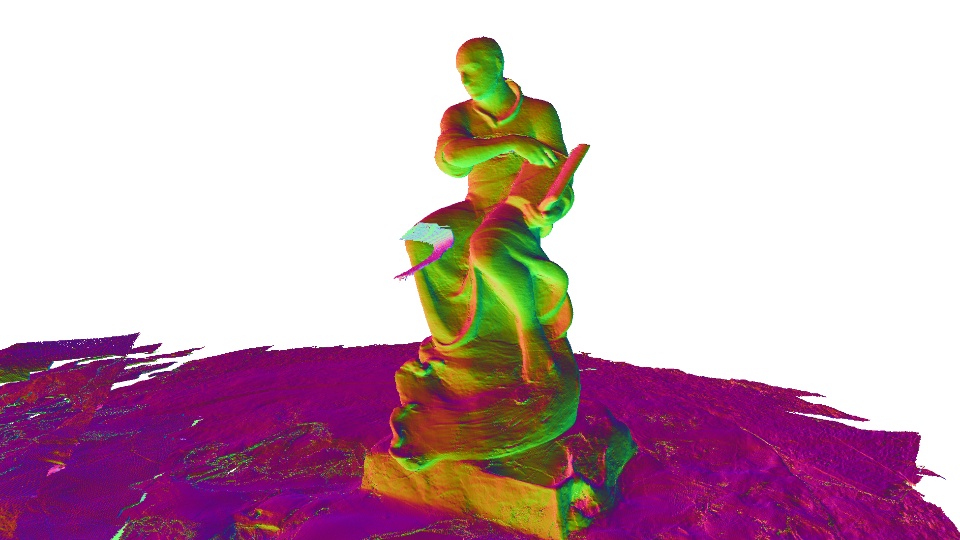}&    
    \includegraphics[width=\tntwidth]{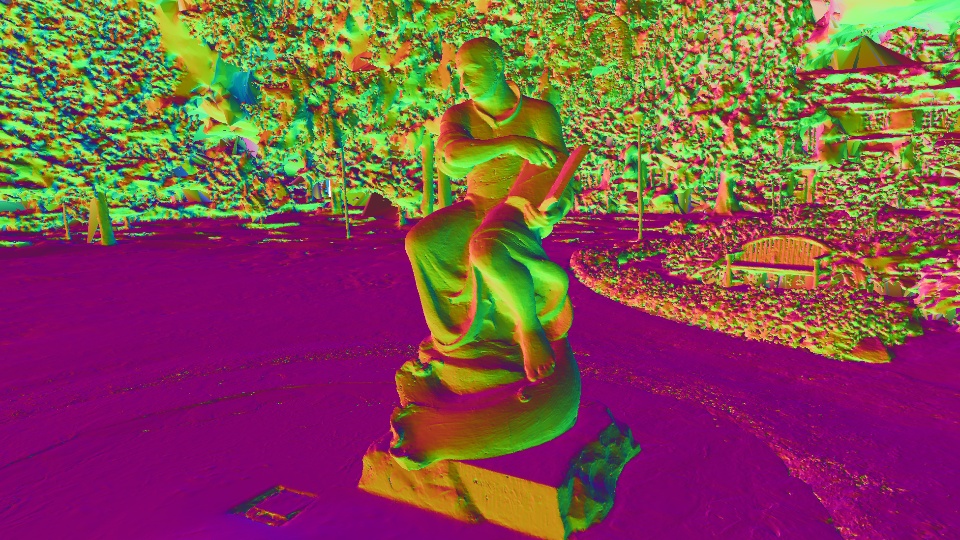}&  
    \includegraphics[width=\tntwidth]{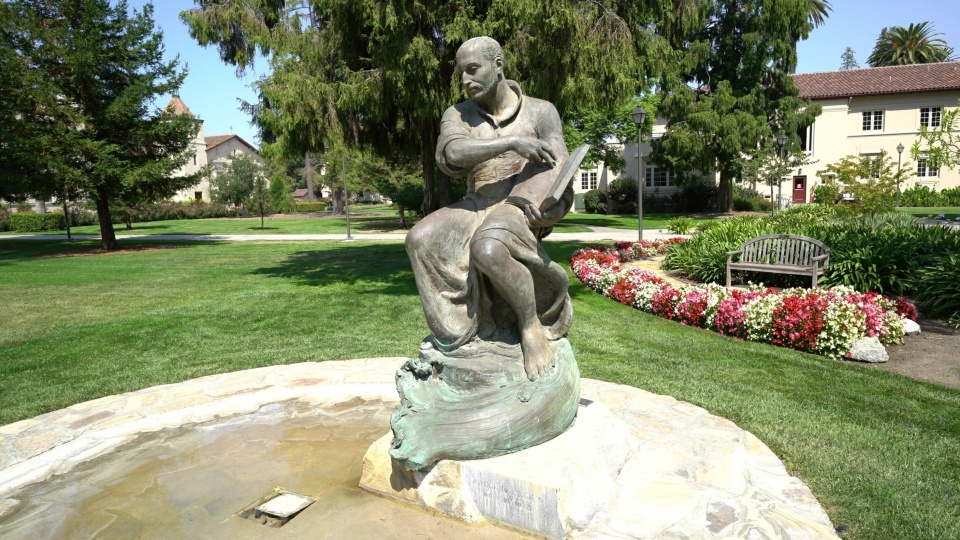} \\
    \includegraphics[width=\tntwidth]{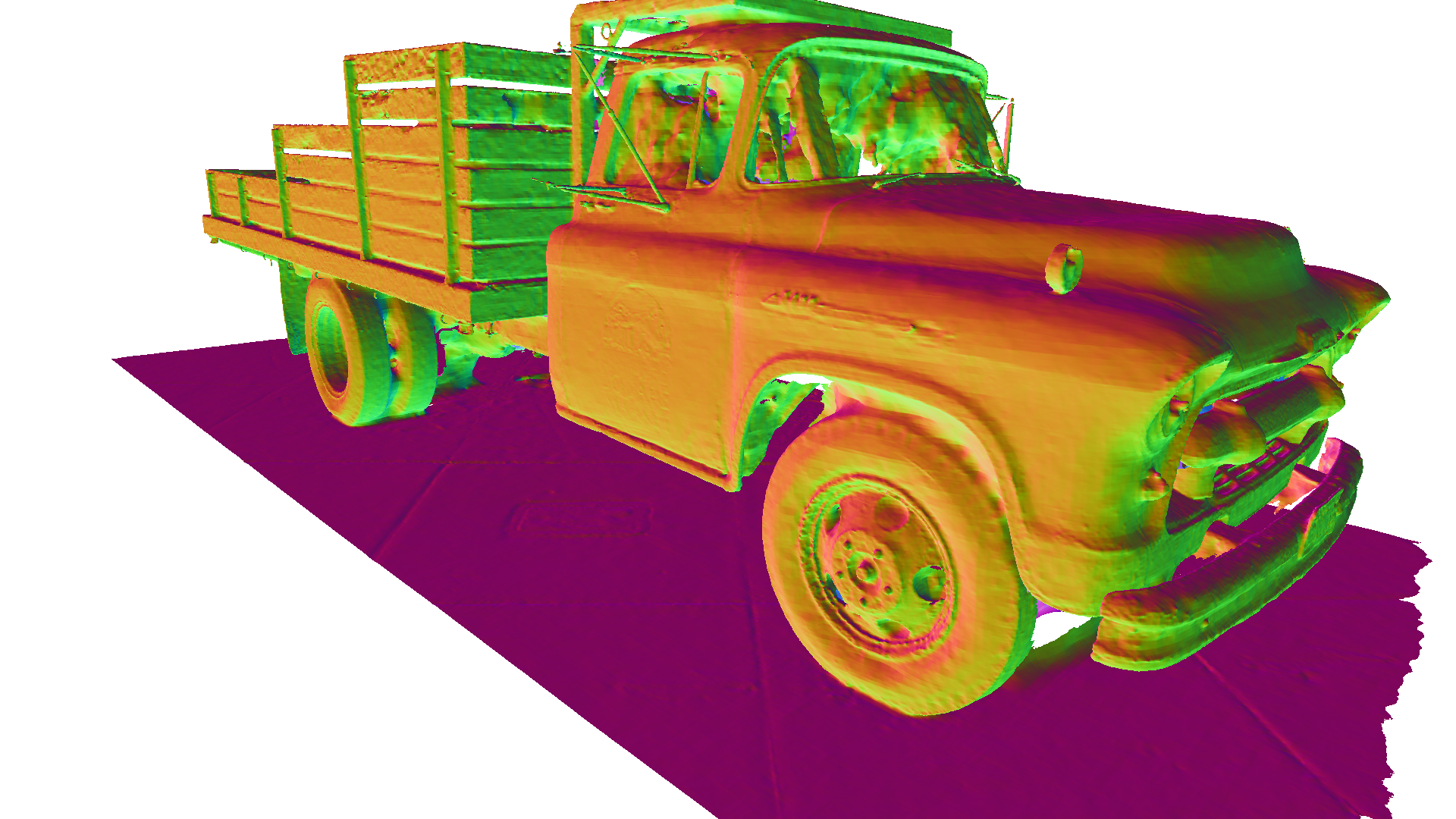}&  
    \includegraphics[width=\tntwidth]{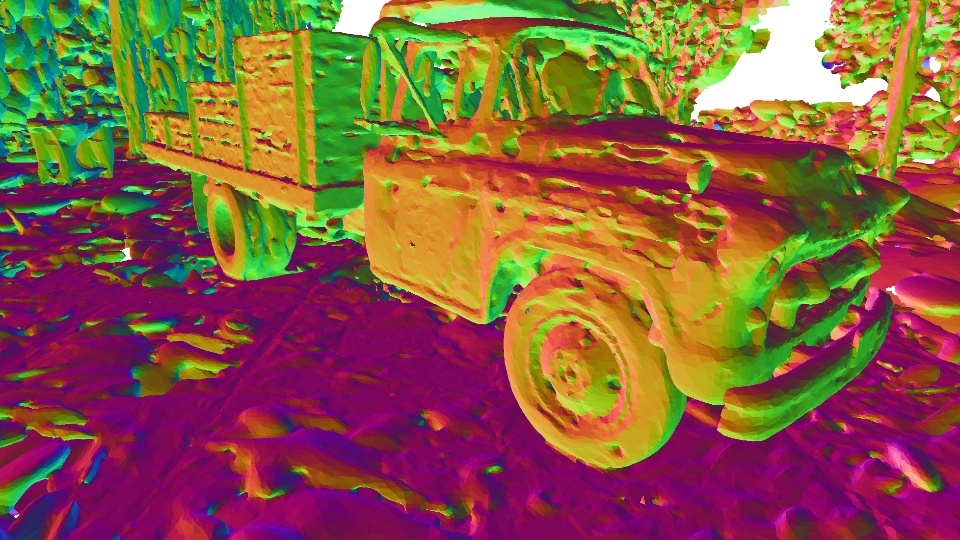}&    
    \includegraphics[width=\tntwidth]{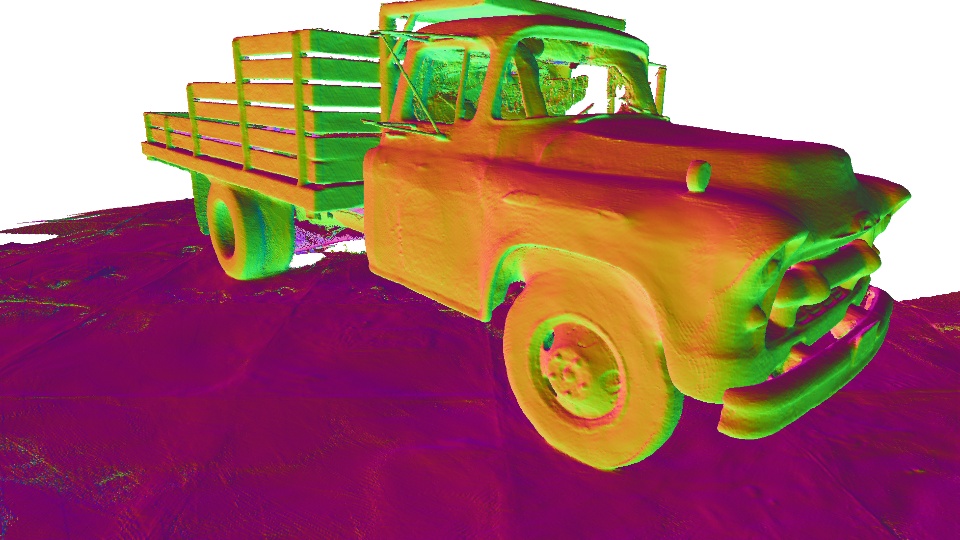}&    
    \includegraphics[width=\tntwidth]{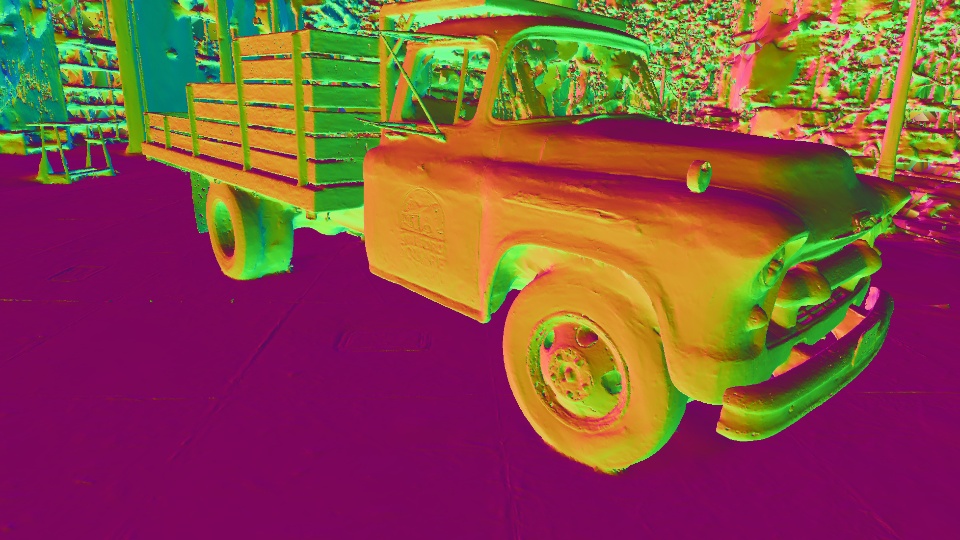}&  
    \includegraphics[width=\tntwidth]{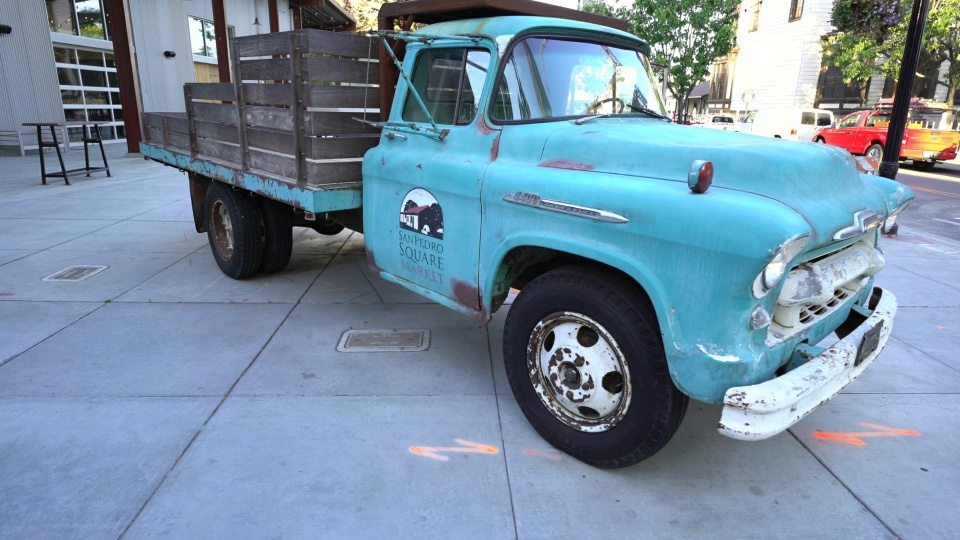} \\
    Neuralangelo~\cite{li2023neuralangelo}&
    SuGar~\cite{guedon2023sugar} & 2DGS~\cite{Huang2024ARXIV} & Ours & GT
    \end{tabular}
    \vspace{-0.1in}
    \caption{\textbf{Surface Reconstruction on the Tanks and Temples Dataset~\cite{Knapitsch2017}}. We show the rendered normal maps from extract meshes and GT images for reference. \update{Neural implicit methods, such as Neuralangelo~\cite{li2023neuralangelo}, model the foreground with SDF and the background with NeRF. Therefore, only the foreground mesh is extracted.} SuGaR's mesh is noisy~\cite{guedon2023sugar} and 2DGS fails at reconstructing background regions~\cite{Huang2024ARXIV}. In contrast, our method can reconstruct detailed surfaces for both foreground objects and background regions.
    }
    \label{fig:tnt}
\end{figure*}

\boldparagraph{Binary Search of Level Set}
Upon determining the opacity values for the tetrahedral grid, we proceed to extract triangle meshes using the Marching Tetrahedra method~\cite{shen2021dmtet}. Traditional algorithms, including Marching Cubes~\cite{lorensen1998marching} and Marching Tetrahedra, typically rely on linear interpolation to approximate the level set, presuming the underlying field's linearity. This assumption, however, misaligns with the characteristics of our Gaussian Opacity Field, leading to artifacts due to linear interpolation, as shown in Figure~\ref{fig:mc_binary} (a). To overcome this discrepancy and accurately identify the level set within our non-linear opacity field, we relax the linear assumption to a monotonically increasing assumption. This adjustment allows for the implementation of a binary search algorithm to precisely identify the level set. In practice, we found that conducting 8 iterations of binary search—effectively simulating 256 dense evaluations—yields consistent and reliable outcomes. A comparison highlighting the improvements of binary search is shown in Figure~\ref{fig:mc_binary}.

\section{Experiments}
We conduct a thorough evaluation of our Gaussian Opacity Fields (GOF), comparing its surface reconstruction and novel view synthesis against leading methods. We further validate the effectiveness of its key components through ablation studies.

\begin{table}[t]
\centering
\caption{\textbf{Quantitative results on the Tanks and Temples Dataset~\cite{Knapitsch2017}}. Reconstructions are evaluated with the official evaluation scripts and we report F1-score and average optimization time. \update{The results of implicit methods are taken from Neuralangelo~\cite{li2023neuralangelo} and the results of explicit methods are taken from 2DGS~\cite{Huang2024ARXIV}, where the mesh of 3DGS~\cite{kerbl3Dgaussians} and 2DGS~\cite{Huang2024ARXIV} are extracted with TSDF fusion and SuGaR uses Poisson reconstruction.} GOF outperforms all 3DGS-based surface reconstruction methods by a large margin and performs comparably with the SOTA neural implicit methods while optimizing significantly faster.}
\vspace{-0.1cm}
\resizebox{0.98\columnwidth}{!}{
\begin{tabular}{@{}l|ccc|cccc}
 & \multicolumn{3}{c@{}|}{Implicit} & \multicolumn{3}{c@{}}{Explicit} \\ 
 & NeuS & Geo-Neus & Neuralangelo & SuGaR & 3DGS & 2DGS & Ours\\ 
 \hline
Barn & 0.29 &  0.33 &  \best 0.70  & 0.14 & 0.13 & \tbest 0.41 & \sbest 0.51\\
Caterpillar &  \tbest 0.29 & 0.26 &  \sbest 0.36 & 0.16 & 0.08 & 0.23 & \best 0.41\\
Courthouse &  \tbest 0.17 & 0.12 &  \best 0.28 & 0.08 & 0.09 & 0.16 & \best 0.28\\
Ignatius &   \sbest 0.83 & \tbest 0.72 &  \best 0.89 & 0.33 & 0.04 & 0.51 & 0.68\\
Meetingroom &   \tbest 0.24 & 0.20 &  \best 0.32 &  0.15 & 0.01 & 0.17 & \sbest0.28\\
Truck &  \tbest0.45 &  \tbest0.45 &  \sbest 0.48 &  0.26 & 0.19 & \tbest0.45 & \best 0.59\\ 
\hline
Mean &  \tbest 0.38 & 0.35 &  \best 0.50 & 0.19 & 0.09 & 0.32 & \sbest0.46\\
Time & >24h & >24h & >24h & >1h & \best 14.3~m & \sbest 15.5 ~m & \tbest 24.2~m\\
\end{tabular}
}
\label{tab:tnt}
\vspace{-0.1cm}
\end{table}

\subsection{Implementation Details}
We build GOF upon the open-source 3DGS code base~\cite{kerbl3Dgaussians} and implement custom CUDA kernels for ray-tracing-based volume rendering, regularizations, and opacity evaluation. Regularization parameters are set to $\alpha=1000$ for bounded scenes, $\alpha=100$ for unbounded scenes, and $\beta=0.05$ for all scenes, following 2DGS~\cite{Huang2024ARXIV}. 
\update{As our improved densification strategy increases the number of primitives for the same hyperparameters, we use a higher opacity threshold 0.05 instead of 0.005 for Gaussian pruning, resulting in a similar number of primitives post-training for a fair comparison.
}
Similar to 3DGS, we stop densification at 15k iterations and optimize all of our models for 30k iterations. For mesh extraction, we adapt the Marching Tetrahedra algorithm~\cite{shen2021dmtet} from the Kaolin library~\cite{KaolinLibrary} with our binary search algorithm and extract the mesh for the 0.5 level-set. All experiments are conducted on an NVIDIA A100 GPU.

\setlength\tabcolsep{0.5em}
\begin{table*}[t]
\centering
\caption{\textbf{Quantitative comparison on the DTU Dataset~\cite{jensen2014large}}. We report the Chamfer distance and average optimization time. \update{The results of 3DGS~\cite{kerbl3Dgaussians} and SuGaR~\cite{guedon2023sugar} are taken from 2DGS~\cite{Huang2024ARXIV}.} Our method achieves the highest reconstruction accuracy among other explicit methods.}
\vspace{-0.2cm}
\resizebox{.98\textwidth}{!}{
\begin{tabular}{@{}llcccccccccccccccclcc}
\hline
 \multicolumn{3}{c}{} & 24 & 37 & 40 & 55 & 63 & 65 & 69 & 83 & 97 & 105 & 106 & 110 & 114 & 118 & 122 & & Mean & Time \\ \cline{4-18} \cline{20-21}
\multirow{4}{*}{\rotatebox[origin=c]{90}{implicit}} & NeRF~\cite{mildenhall2021nerf} & & 1.90 & 1.60 & 1.85 & 0.58 & 2.28 & 1.27 & 1.47 & 1.67 & 2.05 & 1.07 & 0.88 & 2.53 & 1.06 & 1.15 & 0.96 & & 1.49 & > 12h \\
 & VolSDF~\cite{yariv2021volume} & &  1.14 &  1.26 &  0.81 & 0.49 & 1.25 &  \tbest0.70 &  \tbest0.72 &  \sbest 1.29 & \tbest 1.18 &  \sbest 0.70 & \tbest 0.66 & \tbest 1.08 &  0.42 &  \tbest 0.61 &  0.55 & & 0.86 & >12h \\
 & NeuS~\cite{wang2021neus} & &  1.00 & 1.37 & 0.93 &  0.43 & 1.10 &  \sbest 0.65 &   \sbest 0.57 &  1.48 &  \sbest 1.09 &  0.83 &  \sbest 0.52 &  1.20 & \sbest 0.35 &  \sbest 0.49 &  0.54 & &  0.84 & >12h \\
 & Neuralangelo~\cite{li2023neuralangelo} & & \best 0.37 & \best 0.72 & \best 0.35 & \best 0.35 & \best 0.87 & \best 0.54 & \best 0.53 & \sbest 1.29 & \best 0.97 & \tbest 0.73 & \best 0.47 & \best 0.74 & \best 0.32 & \best 0.41 & \best 0.43 & & \best 0.61 & > 12h \\ 
 \cline{2-2} \cline{4-18} \cline{20-21}
\multirow{4}{*}{\rotatebox[origin=c]{90}{explicit}} 
&  3DGS~\cite{kerbl3Dgaussians} & & 2.14 & 1.53 & 2.08 & 1.68 & 3.49 & 2.21 & 1.43 & 2.07 & 2.22 & 1.75 &  1.79 & 2.55 & 1.53 & 1.52 & 1.50 & & 1.96 & \sbest {11.2~m} \\
 &  SuGaR~\cite{guedon2023sugar} & & 1.47 & 1.33 & 1.13 & 0.61 & 2.25 & 1.71 & 1.15 & 1.63 & 1.62 & 1.07 & 0.79 & 2.45 & 0.98 & 0.88 & 0.79 & & 1.33 & $\sim$~1h \\
 & GaussianSurfels~\cite{Dai2024GaussianSurfels} & & 0.66 & 0.93 & 0.54 & 0.41 & \tbest1.06 & 1.14 & 0.85 & \sbest1.29 & 1.53 & 0.79 & 0.82 & 1.58 & 0.45 & 0.66 & 0.53 && 0.88 & \best 6.7~m \\
 & 2DGS~\cite{Huang2024ARXIV} &&  \sbest 0.48 &  \tbest0.91 &  \tbest 0.39 &  \tbest 0.39 &  \sbest 1.01 &  0.83 &  0.81 &  1.36 &  1.27 &  0.76  &  0.70 &  1.40 &   \tbest 0.40 &   0.76 &  \tbest 0.52 &&  \tbest 0.80 &  \tbest 10.9~m \\
 & Ours & & \tbest 0.50 & \sbest 0.82 & \sbest 0.37 & \sbest 0.37 & 1.12 &  0.74 & 0.73 & \best 1.18 & 1.29 & \best 0.68 & 0.77 & \sbest 0.90 & 0.42 & 0.66 & \sbest 0.49 && \sbest 0.74 & 18.4~m\\
 \hline
\end{tabular}
}
\label{tab:dtu_result}
\vspace{-0.1cm}
\end{table*}

\newcommand{\mipwidth}{0.24\textwidth}

\begin{figure*}[t]
    \centering
    \setlength{\tabcolsep}{0.1em}
    \renewcommand{\arraystretch}{0.4}
    \scriptsize
    \begin{tabular}{cccc}
    \includegraphics[width=\mipwidth]{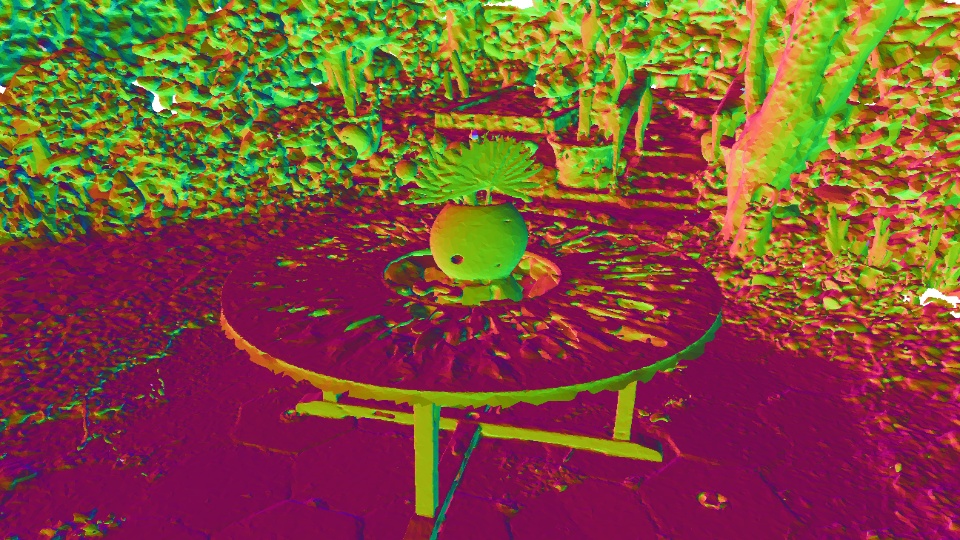}&    
    \includegraphics[width=\mipwidth]{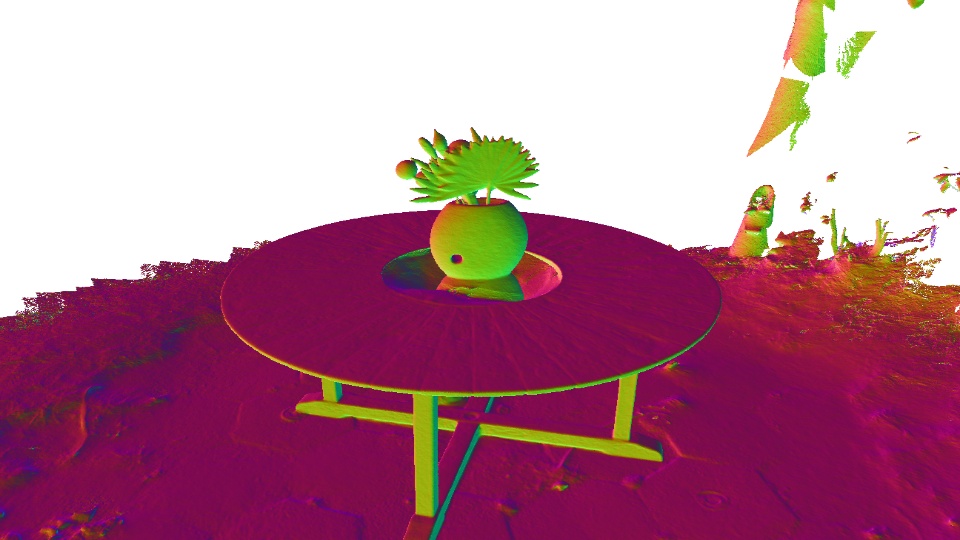}&    
    \includegraphics[width=\mipwidth]{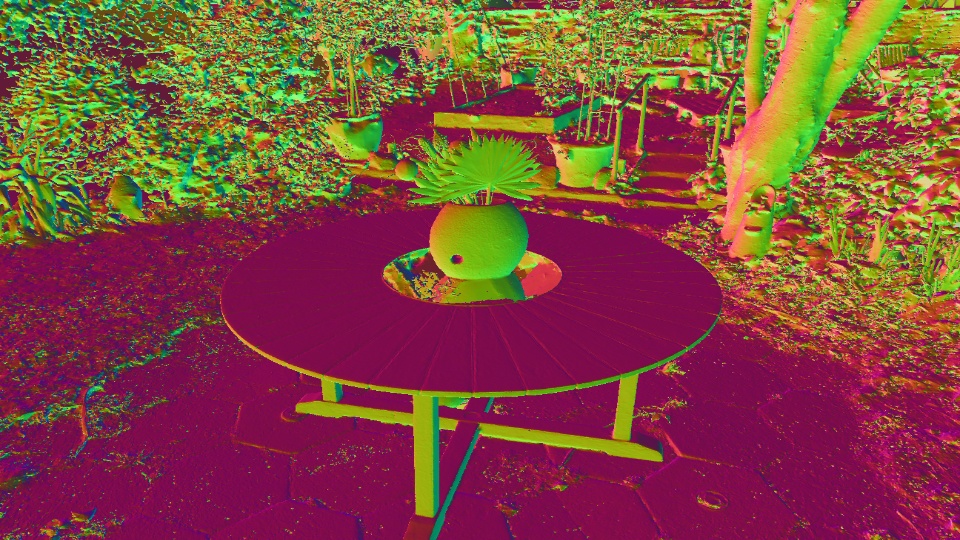}&  
    \includegraphics[width=\mipwidth]{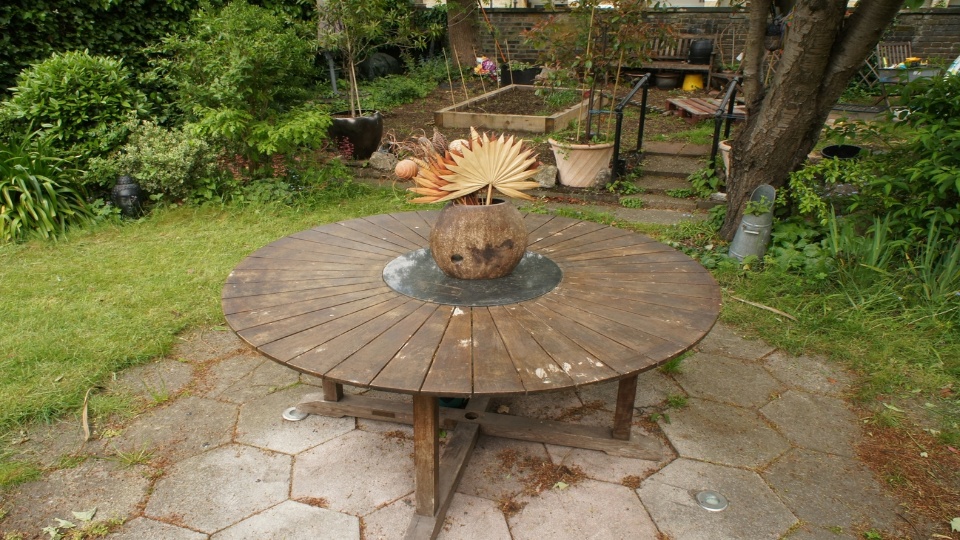} \\
    \includegraphics[width=\mipwidth]{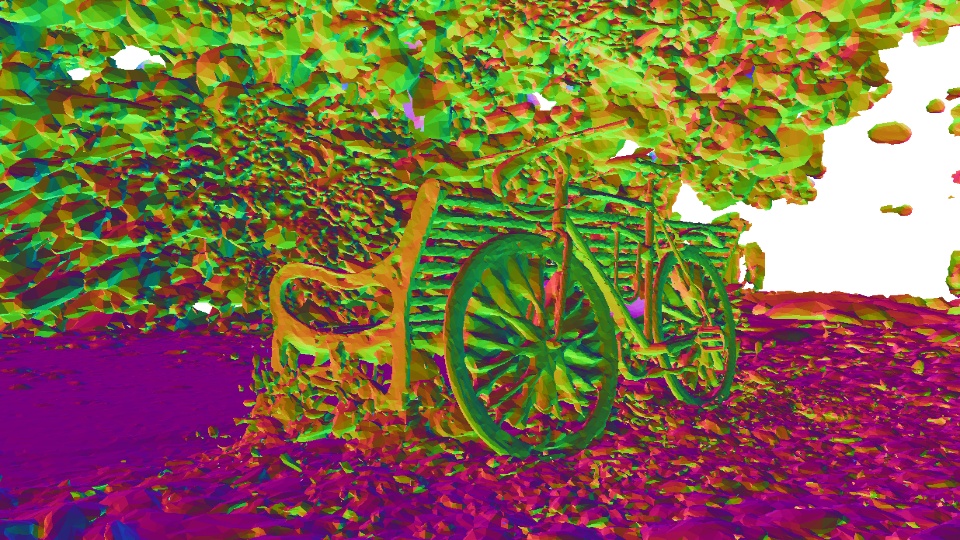}&    
    \includegraphics[width=\mipwidth]{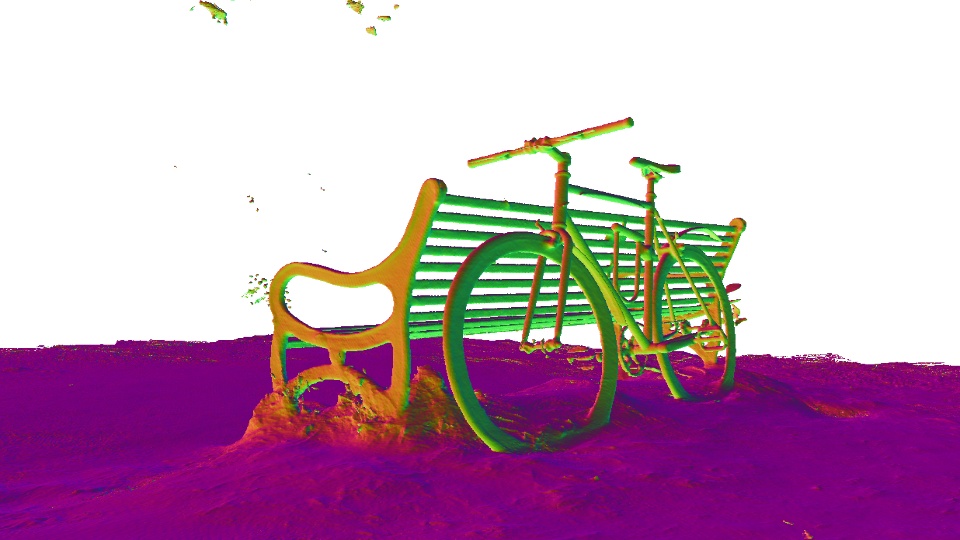}&    
    \includegraphics[width=\mipwidth]{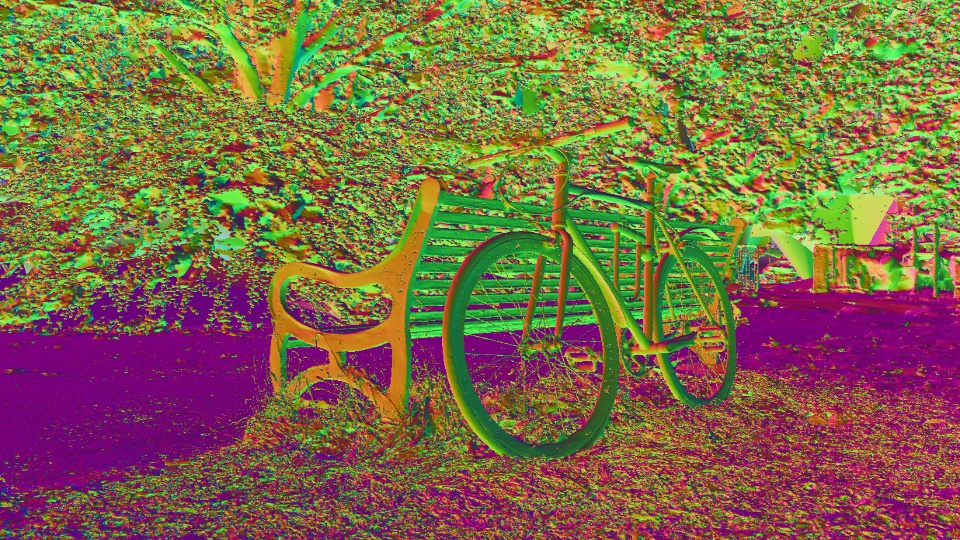}&  
    \includegraphics[width=\mipwidth]{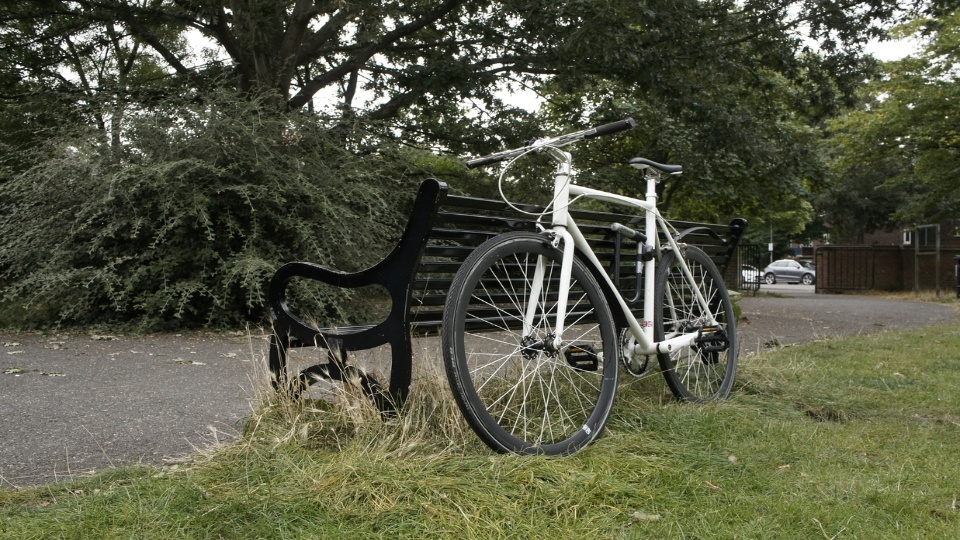} \\
    \includegraphics[width=\mipwidth]{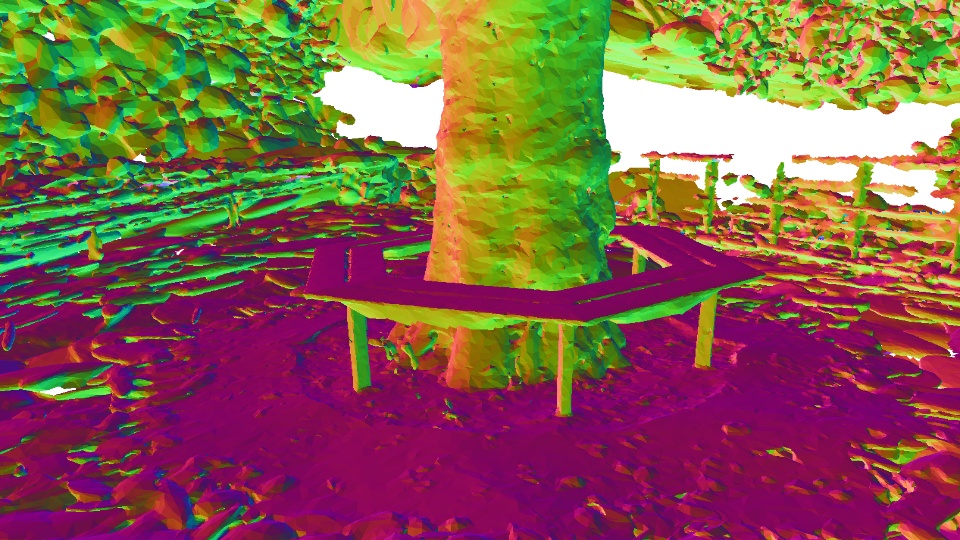}&    
    \includegraphics[width=\mipwidth]{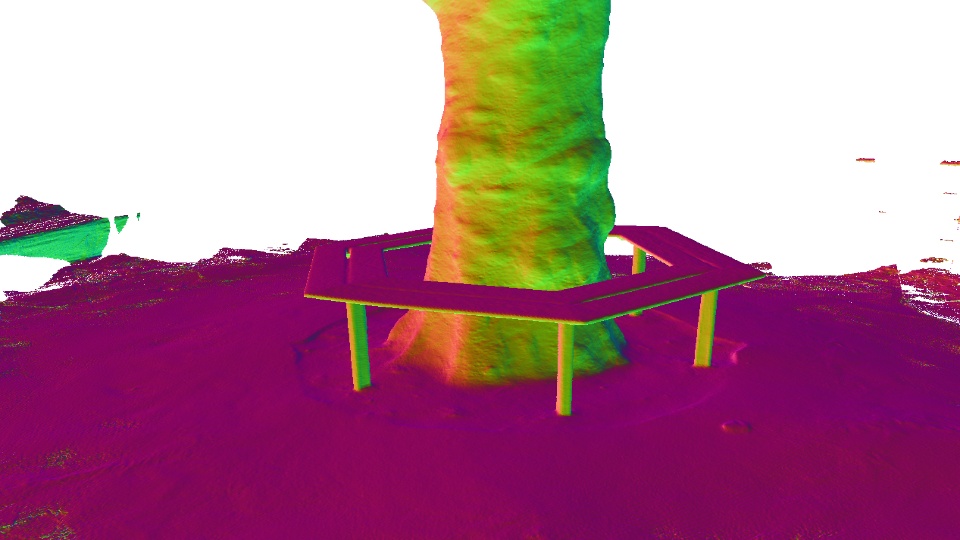}&    
    \includegraphics[width=\mipwidth]{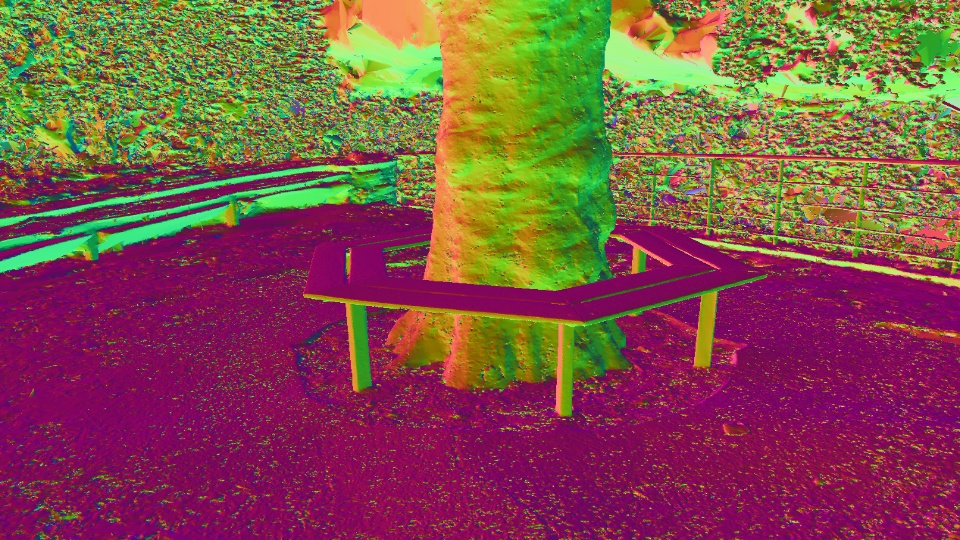}&  
    \includegraphics[width=\mipwidth]{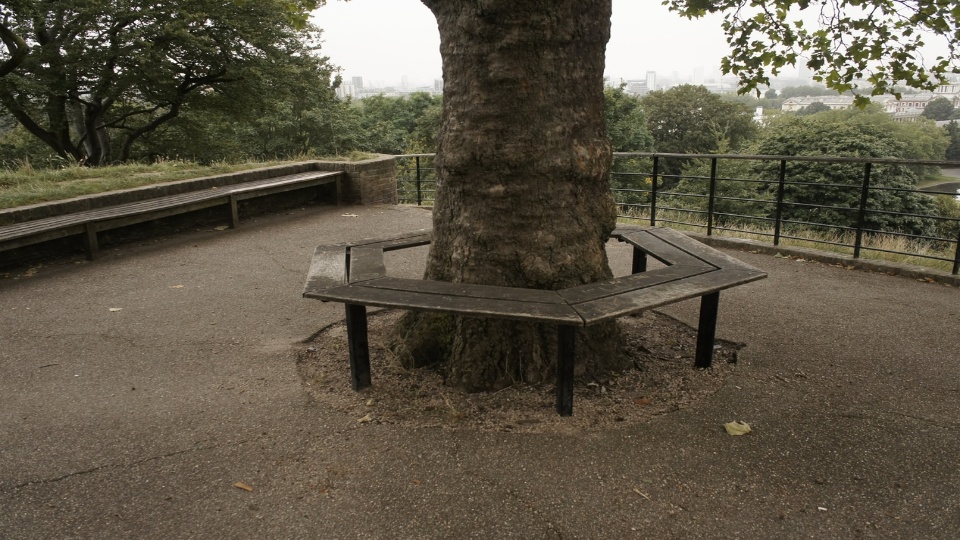} \\
    SuGaR~\cite{guedon2023sugar} & 2DGS~\cite{Huang2024ARXIV} & Ours & GT
    \end{tabular}
    \vspace{-0.1in}
    \caption{\textbf{Reconstructions on the Mip-NeRF 360 Dataset~\cite{barron2022mipnerf360}}. We show the rendered normal maps from extract meshes together with GT images for reference. Our method can reconstruct detailed surfaces for both foreground objects and background regions while meshes from previous work are noisy~\cite{guedon2023sugar} or fail to reconstruct background regions and thin structures, such as the spokes in the bicycle scene~\cite{Huang2024ARXIV}.
    }
    \label{fig:360}
    \vspace{-0.1cm}
\end{figure*}

\subsection{Geometry Evaluation}
We first compare against both SOTA implicit and explicit surface reconstruction methods on the Tanks and Temples Dataset. Reconstructions are evaluated only for the foreground objects since the ground truth point clouds do not contain background regions. As shown in Table~\ref{tab:tnt}, our method is competitive with the leading implicit approaches~\cite{li2023neuralangelo} while being much more efficient to be optimized. Note that most implicit approaches~\cite{li2023neuralangelo,wang2021neus} only reconstruct the foreground objects, while our method can reconstruct detailed meshes also for the background regions, which is of great importance for mesh-based real-time rendering~\cite{Reiser2024ARXIV}. Furthermore, while our method is slightly slower than 3DGS~\cite{kerbl3Dgaussians} and 2DGS~\cite{Huang2024ARXIV} due to the ray-Gaussian intersection computation, it significantly outperforms all SOTA 3DGS-based methods in terms of reconstruction quality. A qualitative comparison is shown in Figure~\ref{fig:tnt}. GOF reconstructs fine-detailed surfaces for both foreground objects and the background regions. By contrast, meshes extracted from SuGaR~\cite{guedon2023sugar} are noisy, while 2DGS~\cite{Huang2024ARXIV} fails to extract geometry for the background regions.

We further compare against SOTA surface reconstruction methods on the DTU dataset~\cite{jensen2014large}. As shown in Table~\ref{tab:dtu_result}, our method outperforms all other 3DGS-based methods~\cite{guedon2023sugar,Dai2024GaussianSurfels,Huang2024ARXIV}. Despite a performance gap with the leading implicit reconstruction method~\cite{li2023neuralangelo}, GOF's optimization is much faster. This performance disparity is attributed to the DTU dataset's strong view-dependent appearance. Utilizing a better view-dependent appearance modeling~\cite{verbin2022refnerf} or a coarse-to-fine training strategy~\cite{li2023neuralangelo} could potentially improve the reconstructions.

\subsection{Novel view Synthesis}
\begin{table}[t]
\centering
\caption{\textbf{Quantitative results on Mip-NeRF 360~\cite{barron2022mip} dataset.}
\update{All results of the baseline methods are taken from their papers whenever available and we rerun SuGaR~\cite{guedon2023sugar} with the same setting as ours.} Our method achieved SOTA NVS results, especially in the outdoor scenes in terms of LPIPS.
}
\resizebox{0.98\columnwidth}{!}{
\begin{tabular}{@{}l|ccc|ccc}
 & \multicolumn{3}{c@{}|}{Outdoor Scene} & \multicolumn{3}{c@{}}{Indoor scene} \\ 
& PSNR~$\uparrow$ & SSIM~$\uparrow$ & LPIPS~$\downarrow$ & PSNR~$\uparrow$ & 
SSIM~$\uparrow$ & LPIPS~$\downarrow$ \\
\hline
NeRF & 21.46 & 0.458 & 0.515 & 26.84 &  0.790 & 0.370 \\
Deep Blending & 21.54 &0.524 & 0.364 & 26.40 & 0.844 & 0.261 \\
Instant NGP & 22.90 & 0.566 & 0.371 & 29.15 & 0.880 & 0.216 \\
MERF & 23.19 & 0.616 &   0.343 & 27.80 & 0.855 & 0.271 \\
MipNeRF360 &   24.47 &  0.691 &  0.283 &   \best 31.72 &  0.917 &  \best 0.180 \\
\hline
\hline
Mobile-NeRF & 21.95 & 0.470 & 0.470 & - & - & - \\
BakedSDF & 22.47 & 0.585 &  0.349 & 27.06 & 0.836 & 0.258 \\
SuGaR &  22.93 & 0.629 & 0.356 & 29.43 & 0.906 & 0.225 \\
BOG & 23.94 & 0.680 & 0.263 & 27.71 & 0873 & 0.227 \\
\hline
\hline
3DGS &  \tbest 24.64 &  \sbest 0.731 &  \sbest 0.234 &   \tbest 30.41 &  \tbest 0.920 &  \tbest 0.189 \\
Mip-Splatting & \sbest 24.65 & \tbest 0.729 & \tbest 0.245 & \sbest 30.90 & \sbest 0.921 &  0.194  \\
2DGS &  24.34 & 0.717 & 0.246 & 30.40 & 0.916 & 0.195 \\
Ours & \best 24.82 & \best 0.750 &\best  0.202 & \tbest 30.79 & \best 0.924 &\sbest  0.184
\end{tabular}
}
\label{tab:mipnerf360}
\end{table}

\newcommand{\mipablationwidth}{0.33\textwidth}
\begin{figure*}[t]
    \centering
    \setlength{\tabcolsep}{0.1em}
    \renewcommand{\arraystretch}{0.2}
    \scriptsize
    \begin{tabular}{ccc}
    \includegraphics[width=\mipablationwidth]{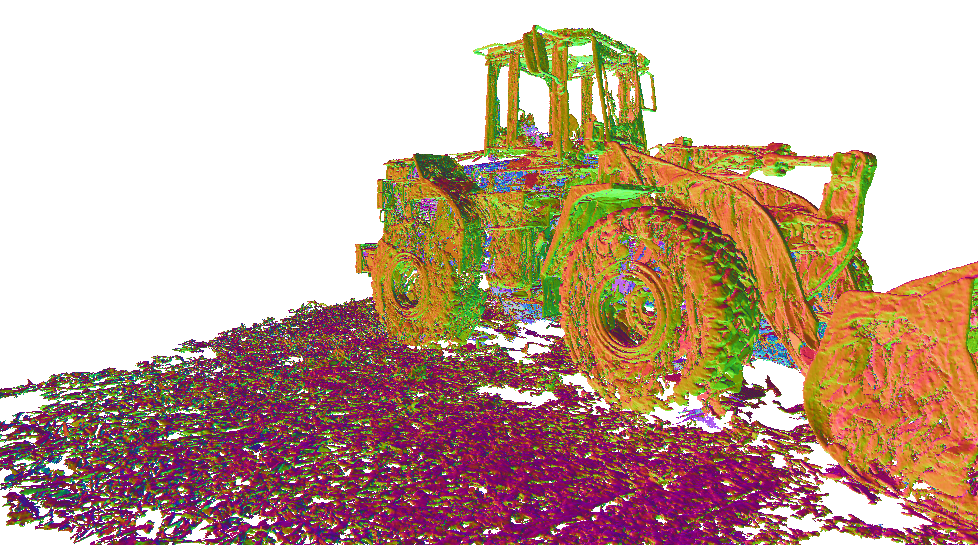}&    
    \includegraphics[width=\mipablationwidth]{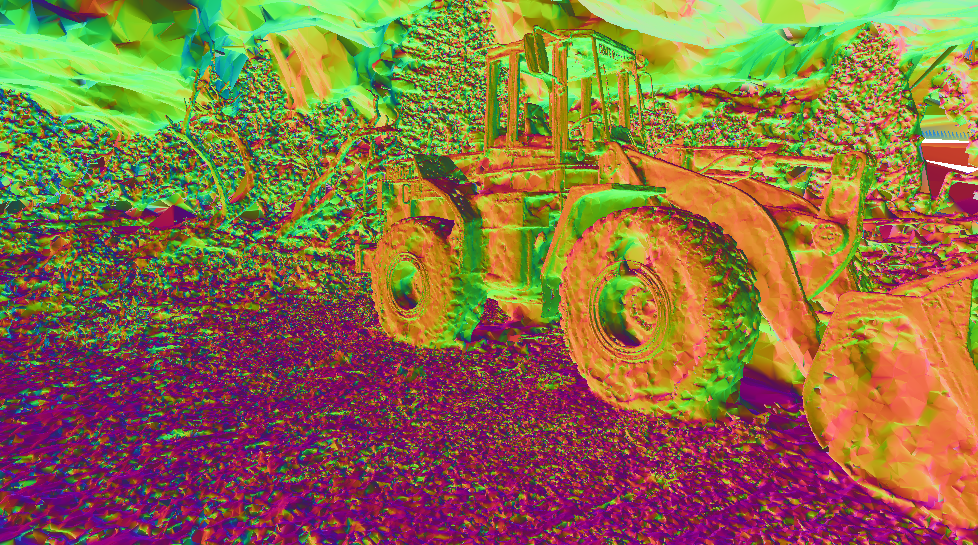} &
    \includegraphics[width=\mipablationwidth]{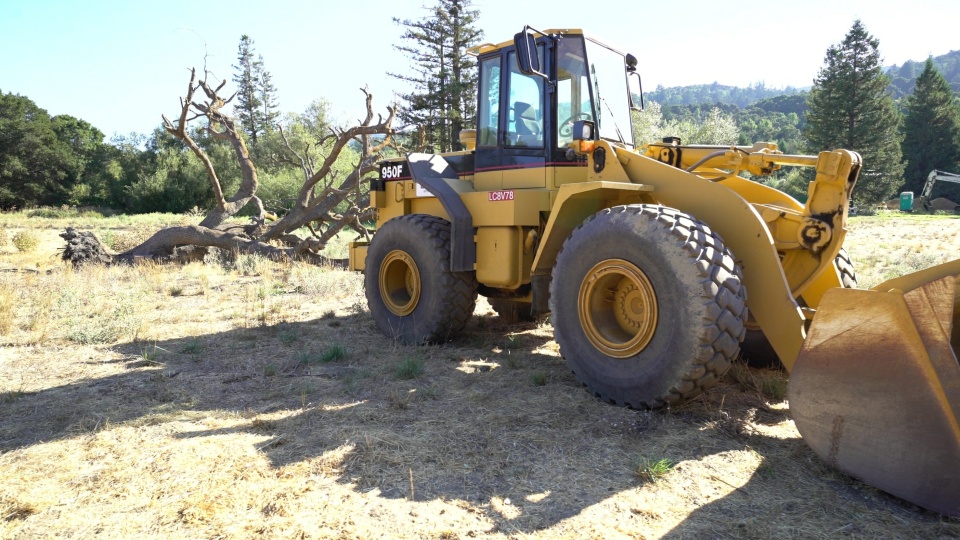} \\
    Mip-Splatting + TSDF Fusion & Mip-Splatting + GOF & GT Image
    \end{tabular}
    \vspace{-0.1in}
    \caption{\textbf{Comparison of mesh extraction methods.} Applying GOF to SOTA Mip-Splatting~\cite{Yu2023MipSplatting} yields significant improvements over TSDF, enabling complete mesh extraction for background regions.
    }
    \vspace{-0.3cm}
    \label{fig:mip_gof}
\end{figure*}

To evaluate the NVS results of GOF, we further compare against SOTA NVS methods on the Mip-NeRF 360 dataset~\cite{barron2022mipnerf360}. The quantitative results are shown in Table~\ref{tab:mipnerf360}. GOF not only performs slightly better than all other 3DGS-based methods in terms of PSNR, but also outperforms all other methods significantly in terms of LPIPS~\cite{zhang2018perceptual} in the outdoor scenes. The main improvements come from our improved densification strategy. In the ablation, we show that adding our densification strategy to 3DGS~\cite{kerbl3Dgaussians} and Mip-Splatting~\cite{Yu2023MipSplatting} improves the NVS results by a large margin. For the indoor scenes, our results are similar to Mip-Splatting~\cite{Yu2023MipSplatting}, with less than 0.1 PSNR difference, which we attribute to our regularizations that trade-off NVS and surface reconstruction. Our method also outperforms Sugar~\cite{guedon2023sugar} and 2DGS~\cite{Huang2024ARXIV} in all metrics. We show a qualitative comparison of extracted meshes in Figure~\ref{fig:360}. Similar to our observations on the Tanks and Temples dataset~\cite{Knapitsch2017}, GOF can reconstruct detailed surfaces for both foreground objects and background regions, while SuGaR's~\cite{guedon2023sugar} meshes are noisy and have fewer details and 2DGS fails to extract meshes for the background regions. More qualitative results of our method can be found in Figure~\ref{fig:pro} and Figure~\ref{fig:our_one}.

\subsection{Ablation Study}
\begin{table}[t]
\centering
\caption{\textbf{Ablation on the Tanks and Temples Dataset~\cite{Knapitsch2017}.} \update{The metrics are computed with the official script from the dataset.} GOF significantly improves surface reconstruction quality. Our regularization, decoupled appearance, and densification strategy modeling further improve the results.}
\vspace{-3pt}
\resizebox{0.92\columnwidth}{!}{
\begin{tabular}{@{}l|ccc}
 & Precision~$\uparrow$ & Recall~$\uparrow$ & F-score~$\uparrow$\\ 
 \hline
A. Mip-Splatting w/ TSDF & 0.15 & 0.25 & 0.16 \\
B. Mip-Splatting w/ GOF & 0.40 & 0.33 & 0.36 \\
\hline
C. Ours w/o GOF & 0.37 & 0.45 & 0.39 \\
D. Ours w/o normal consistency & 0.41 & 0.35 & 0.37 \\
E. Ours w/o decoupled appearance & 0.49 & 0.39 & 0.43 \\
F. Ours w/ minimal axis's normal & 0.46 & 0.36 & 0.40 \\
G. Ours w/o improved densification & 0.52 & 0.39 & 0.44 \\
\hline
H. Ours & 0.54 & 0.42 & 0.46 \\
\end{tabular}
}
\label{tab:tnt_ablation}
\vspace{-0.3cm}
\end{table}

\newcommand{\mtwidth}{0.19\textwidth}

\begin{figure*}[t]
    \centering
    \setlength{\tabcolsep}{0.1em}
    \renewcommand{\arraystretch}{0.4}
    \scriptsize
    \begin{tabular}{ccccc}
    \includegraphics[width=\mtwidth]{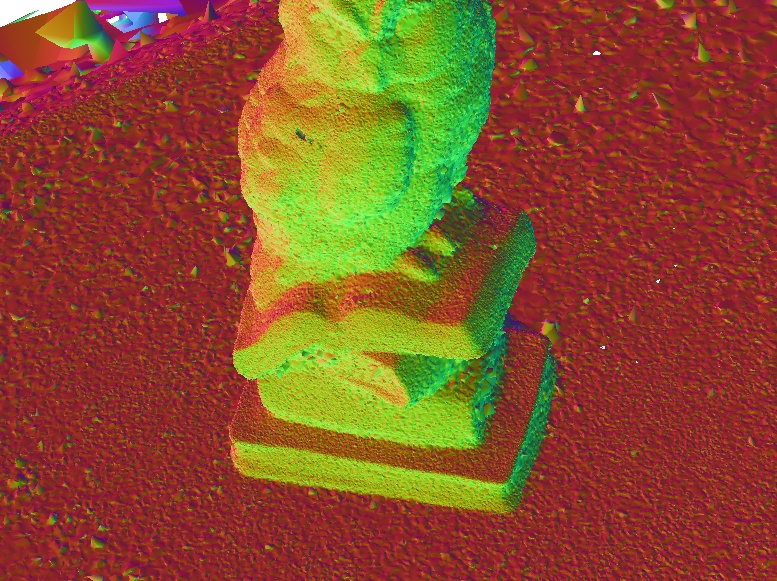}&    
    \includegraphics[width=\mtwidth]{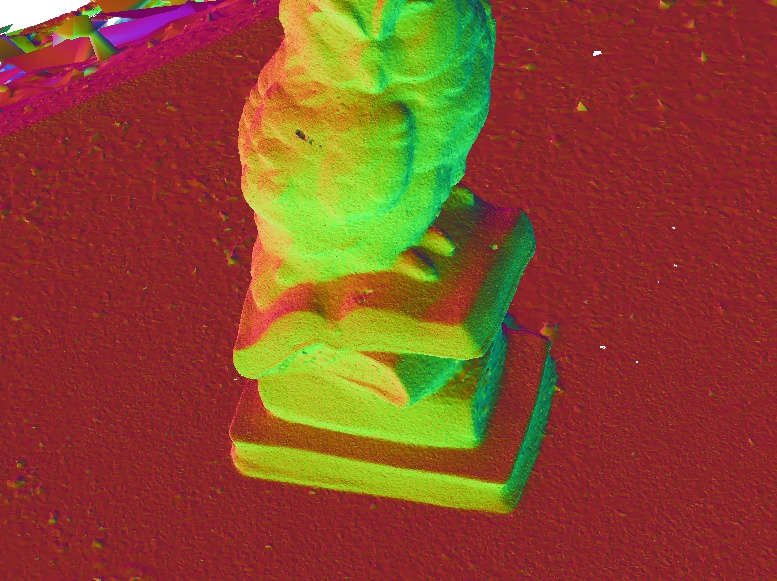}&    
    \includegraphics[width=\mtwidth]{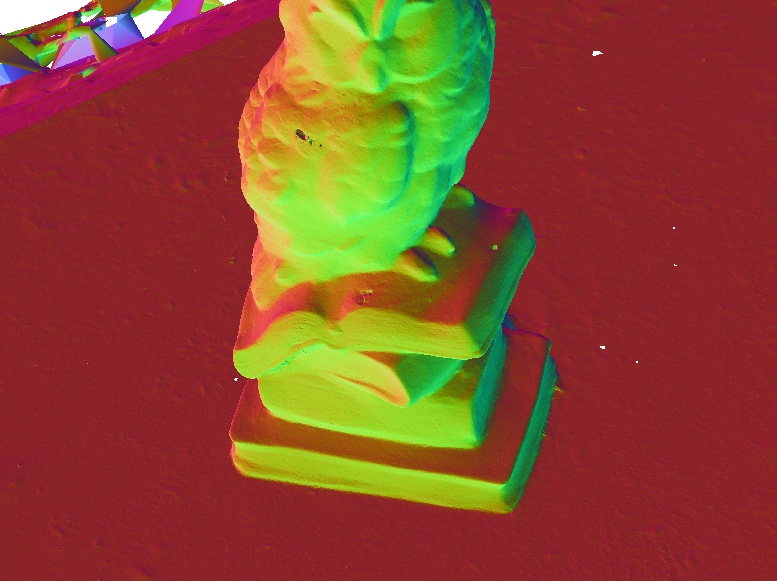}&    
    \includegraphics[width=\mtwidth]{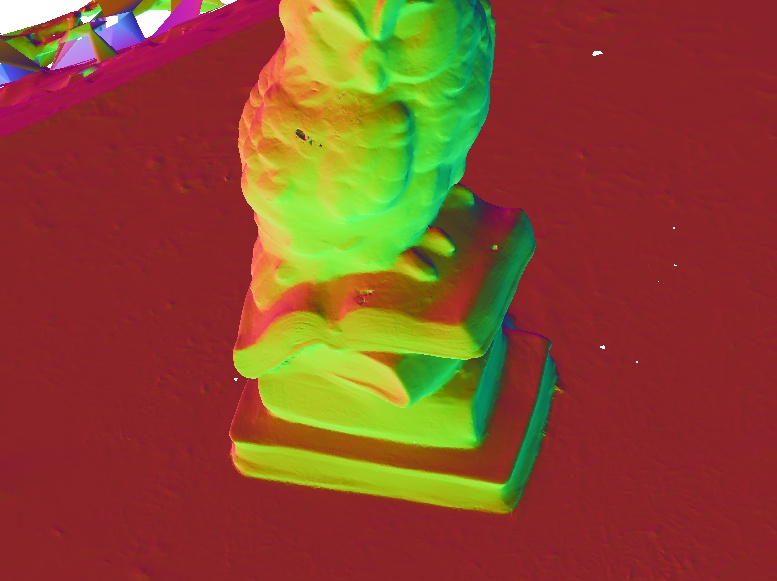}&
    \includegraphics[width=\mtwidth]{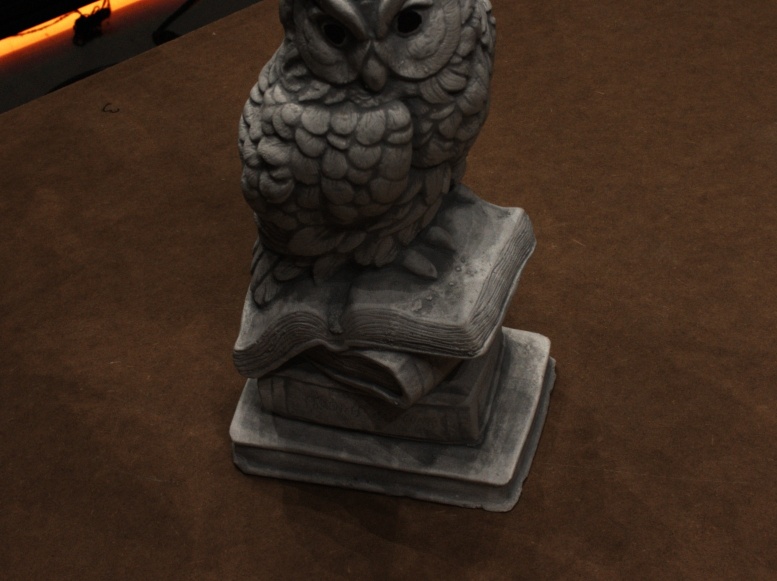} \\
    Step 0 & Step 1 & Step 3 & Step 7 & GT Image
    \end{tabular}
    \vspace{-0.1in}
    \caption{\textbf{Different number of binary search steps with marching tetrahedra on the DTU dataset~\cite{jensen2014large}.} 
    Applying our binary search to Marching Tetrahedra~\cite{shen2021dmtet} significantly improves the quality of extracted meshes in just few steps.
    }
    \label{fig:mt_binary}
\end{figure*}

\newcommand{\levelsetwidth}{0.19\textwidth}

\begin{figure*}[t]
    \centering
    \setlength{\tabcolsep}{0.1em}
    \renewcommand{\arraystretch}{0.4}
    \scriptsize
    \begin{tabular}{ccccc}
    \includegraphics[width=\levelsetwidth]{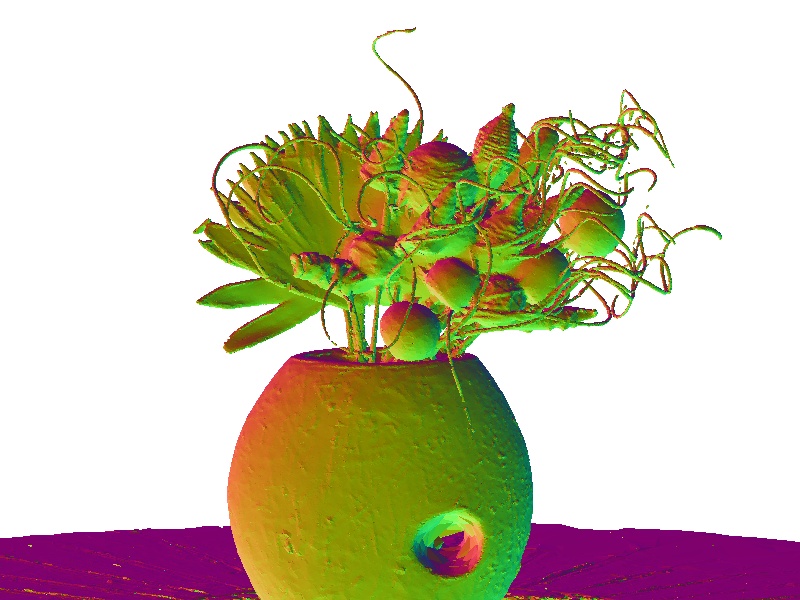}&    
    \includegraphics[width=\levelsetwidth]{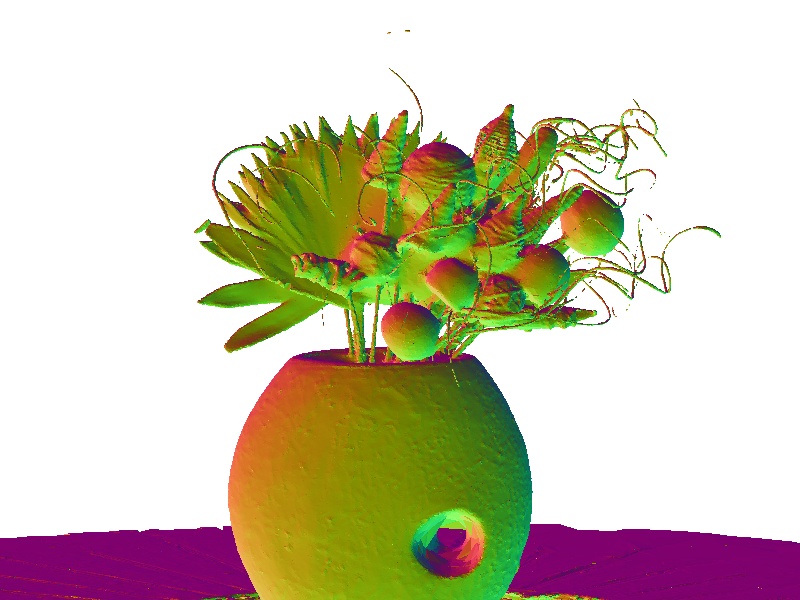}&    
    \includegraphics[width=\levelsetwidth]{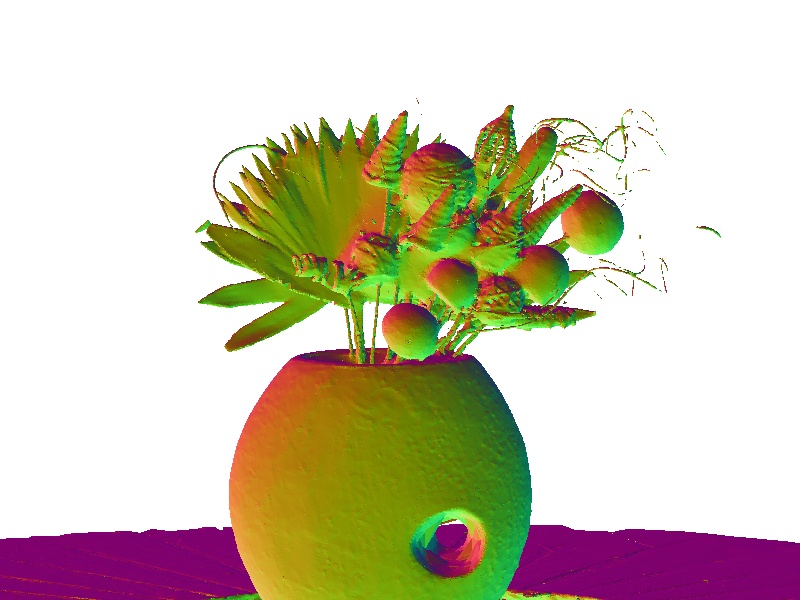}&    
    \includegraphics[width=\levelsetwidth]{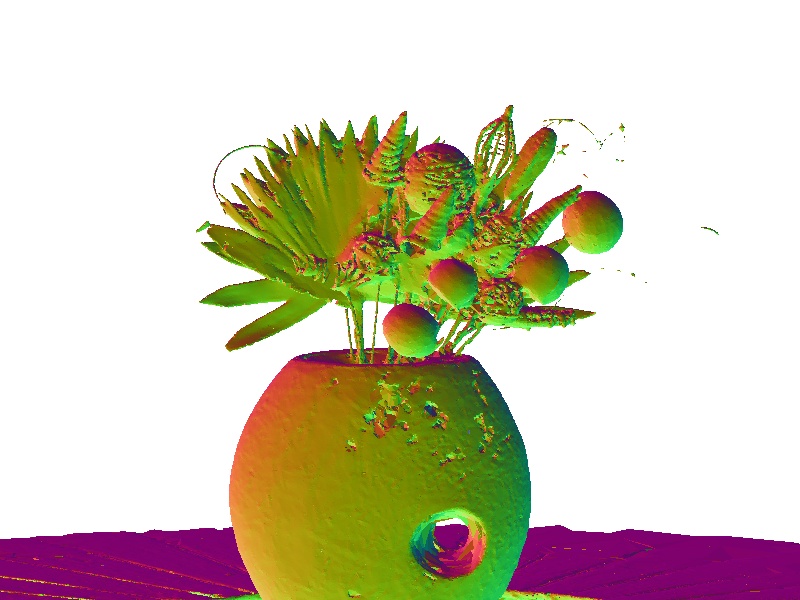}&
    \includegraphics[width=\levelsetwidth]{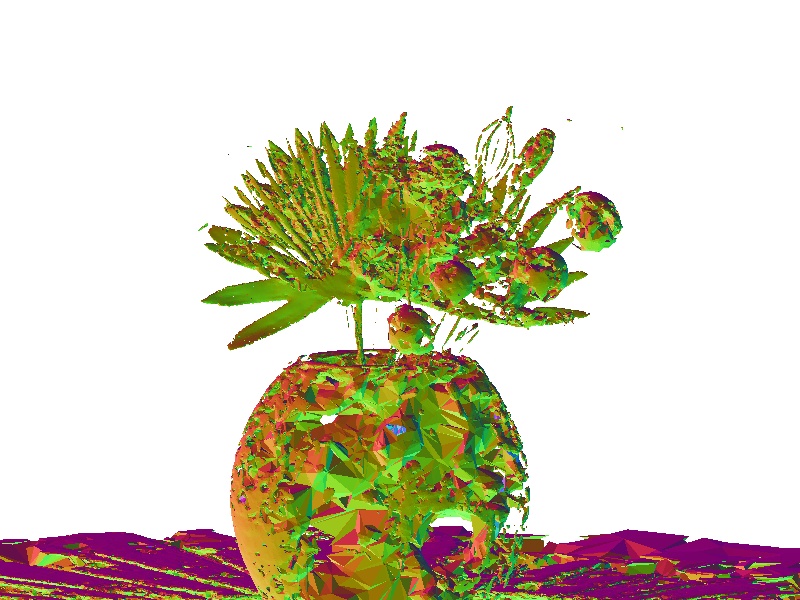} \\
    0.1 & 0.3 & 0.5 & 0.7 & 0.9
    \end{tabular}
    \vspace{-0.1in}
    \caption{\textbf{Meshes extracted with different level sets on the Mip-NeRF 360 dataset~\cite{barron2022mipnerf360}.} 
    Our method supports multi-layer meshes extraction by using different level sets for marching tetrahedra.
    }
    \label{fig:levelsets}
\end{figure*}

\boldparagraph{Mesh Extraction}
Our GOF enables mesh extraction from 3D Gaussians directly by identifying a level set without resorting to Poisson reconstruction or TSDF fusion. \update{Compared to using TSDF fusion for mesh extraction (Table~\ref{tab:tnt_ablation}, C), our tetrahedra-based mesh extraction (Table~\ref{tab:tnt_ablation}, H) significantly improves the quality of the extracted meshes.}

To further demonstrate the effectiveness and generalizability of our mesh extraction strategy, we apply it to a SOTA 3DGS-based model, Mip-Splatting~\cite{Yu2023MipSplatting}, and compare it with TSDF fusion. A comparison of the extracted meshes is shown in Figure~\ref{fig:mip_gof}. The mesh extracted from TSDF fusion is noisy and has a lot of holes on the ground, due to the inconsistency of depth. By contrast, our extracted mesh is more complete. Our method also extracts detailed surfaces for the background regions. Quantitative results in Table~\ref{tab:tnt_ablation} (A vs. B) further indicate the effectiveness of our method.

\boldparagraph{Regularization and Appearance Modeling}
As shown in Table~\ref{tab:tnt_ablation} (D vs. H), using the normal regularization during training significantly improves reconstruction quality, which is consistent with the observation in 2DGS~\cite{Huang2024ARXIV}. Including the decoupled appearance modeling~\cite{lin2024vastgaussian} further improves the reconstruction results as shown in Table~\ref{tab:tnt_ablation} (E vs. H), since the model is less likely to model view-dependent appearance with geometry.

\update{
\boldparagraph{Normal Definition}
We approximate the normal of a Gaussian as the normal of the ray-Gaussian intersection plane. By contrast, SuGaR~\cite{guedon2023sugar} approximates the normal as one of Ganssian's axis directions with minimal scale. We experimented by replacing our normal definition with the axis direction of the minimum scale and found that the F1-score decreases from 0.46 to 0.40 on the TNT dataset, as shown in Table~\ref{tab:tnt_ablation} (F vs. H), validating the effectiveness of our normal definition.

\boldparagraph{Densification on Geometry}
As shown in Table~\ref{tab:tnt_ablation} (G vs. H), disabling our improved densification strategy leads to a decrease in the F1-score from 0.46 to 0.44 in the TNT dataset~\cite{Knapitsch2017}, showing that our densification metric contributes positively to geometry reconstruction.

\boldparagraph{Overall Improvements}
When using the same mesh extraction method, such as TSDF fusion (Table~\ref{tab:tnt_ablation}, A vs. C) or our tetrahedra-based mesh extraction (Table~\ref{tab:tnt_ablation}, B vs. H), the improvements of our method over Mip-Splatting~\cite{Yu2023MipSplatting} are due to additional regularization, appearance modeling, and our improved densification strategy. Without these elements, the reconstruction quality between our method and rasterization-based methods is comparable. For instance, disabling our normal regularization results in similar reconstruction quality compared to Mip-Splatting with GOF (Table~\ref{tab:tnt_ablation}, D vs. B).

\begin{table}[t]
\centering
\caption{\textbf{Ablation on the Mip-NeRF 360~\cite{barron2022mip} dataset.} Our densification strategy improves the novel view synthesis results for 3DGS~\cite{kerbl3Dgaussians} and Mip-Splatting~\cite{Yu2023MipSplatting} significantly.}
\resizebox{0.98\columnwidth}{!}{
\begin{tabular}{@{}l|ccc|ccc}
 & \multicolumn{3}{c@{}|}{Outdoor Scene} & \multicolumn{3}{c@{}}{Indoor scene} \\ 
& PSNR~$\uparrow$ & SSIM~$\uparrow$ & LPIPS~$\downarrow$ & PSNR~$\uparrow$ & 
SSIM~$\uparrow$ & LPIPS~$\downarrow$ \\
\hline
3DGS &  \textbf{24.64} &  0.731 &  0.234 &   30.41 &  0.920 &  0.189 \\
 w/ our densification & 24.62 & \textbf{0.743} & \textbf{0.199} & \textbf{31.10} & \textbf{0.928} & \textbf{0.174} \\
\hline
Mip-Splatting & 24.65 & 0.729 & 0.245 & 30.90 & 0.921 & 0.194  \\
w/ our densification & \textbf{24.77} & \textbf{0.745} & \textbf{0.205} & \textbf{31.18} & \textbf{0.926} & \textbf{0.180}   \\
\hline
Ours & 24.82 & 0.750 & 0.202 & 30.79 & 0.924 & 0.184
\end{tabular}
}
\label{tab:mipnerf360_densify}
\end{table}

\boldparagraph{Densification on View Synthesis}
\label{sec:ab_densification}To further evaluate the effectiveness of our improved densification metric on novel view synthesis, we apply it to 3DGS~\cite{kerbl3Dgaussians} and Mip-Splatting~\cite{Yu2023MipSplatting}. The quantitative results, as shown in Table~\ref{tab:mipnerf360_densify}, demonstrate that our strategy improves NVS results significantly, especially in terms of LPIPS~\cite{zhang2018perceptual}. Qualitative comparisons are also provided in Figure~\ref{fig:densify},where the glass regions are rendered with high fidelity using our densification metric.

It is important to note that our ray-Gaussian intersection formula has similar rendering quality to rasterization-based methods~\cite{kerbl3Dgaussians,Yu2023MipSplatting}. For instance, our NVS results are similar to those of Mip-Splatting when augmented with our densification strategy, as shown in Table~\ref{tab:mipnerf360_densify}. However, the ray-Gaussian intersection formula enables the establishment of opacity fields, improving mesh extraction compared to TSDF fusion. Additionally, our regularization and appearance modeling further improve the reconstruction quality, as shown previously.

\boldparagraph{Binary Search for Marching Tetrahedra}
To demonstrate the effectiveness of applying binary search to Marching Tetrahedra~\cite{shen2021dmtet}, we apply binary search with different numbers of steps. As shown in Figure~\ref{fig:mt_binary}, using binary search significantly improves the quality of reconstructed meshes in just a few iterations.

\boldparagraph{Multi-layer Meshes}
While we extract meshes for the 0.5 level set in the main paper, our method also supports extracting meshes with different level sets. As a proof of concept, we extract meshes with different level sets 0.1, 0.3, 0.5, 0.7, and 0.9 and show the rendered meshes in Figure~\ref{fig:levelsets}. Finer structures can be extracted with a smaller level set, but it might result in expanded meshes.

}

\section{Limitations}
We now discuss some limitations and potential future extensions of our method. 

\boldparagraph{Delaunay Triangulation Efficiency} We employ the CGAL Library~\cite{cgal:pt-tds3-24a} for Delaunay triangulation to construct tetrahedral cells, which has \(O(N \log N)\) complexity. It becomes a bottleneck particularly when the number of points increases. For example, it takes around 8 minutes to construct the tetrahedral cells for the bicycle scene in the Mip-NeRF dataset~\cite{barron2022mipnerf360}. This process could potentially be optimized by considering the spatial locality since the points are generated from 3D Gaussians, or by employing parallel processing techniques on GPUs. Additionally, optimizing the selection and number of Gaussian primitives used for constructing tetrahedral grids could further improve efficiency.

\boldparagraph{Opacity Evaluation Optimization} During the binary search of marching tetrahedra, our method evaluates the opacity of points using all training views, which may lead to redundant computations. Recognizing that a single view can determine a point’s minimal opacity value suggests a more efficient approach could be developed by associating points with their respective influential training views.

\boldparagraph{View Dependent Appearance Modeling} Using spherical harmonics for view-dependent appearance modeling has limitations, such as potentially inaccurately representing reflections as geometric features. Incorporating a better view-dependent appearance model~\cite{verbin2022refnerf} could potentially enhance the quality of reconstructions.

\boldparagraph{Mesh-based Rendering} While the current focus of GOF is on surface reconstruction and novel view synthesis, leveraging the extracted meshes for real-time rendering is an interesting future direction~\cite{Reiser2024ARXIV}. This could potentially improve the quality of mesh-based rendering given the detailed and adaptive meshes extracted with GOF.

\section{Conclusion}
We have presented Gaussian Opacity Fields (GOF), a novel method for efficient, high-quality, and adaptive surface reconstruction in unbounded scenes. Our GOF is derived from ray-tracing-based volume rendering of 3D Gaussians, maintaining consistency with RGB rendering. Our GOF enables geometry extraction directly from 3D Gaussians by identifying its level set, without Poisson reconstruction or TSDF. We approximate the surface normal of Gaussians as the normal of the ray-Gaussian intersection plane and apply depth-normal consistency regularization to enhance geometry reconstructions. Furthermore, we propose an efficient and adaptive mesh extraction method utilizing Marching Tetrahedra, where the tetrahedral grids are induced from 3D Gaussians. %
\update{Our evaluations reveal that GOF surpasses existing explicit methods in both surface reconstruction and novel view synthesis. Further, GOF achieves comparable surface reconstruction results with leading implicit methods while being able to reconstruct detailed meshes for the background regions in unbounded scenes.}

\boldparagraph{Acknowledgement} We thank Christian Reiser and Binbin Huang for insightful discussions and valuable feedback throughout the project and proofreading. ZY and AG are supported by the ERC Starting Grant LEGO-3D (850533) and DFG EXC number 2064/1 - project number 390727645. TS is supported by a Czech Science Foundation (GACR) EXPRO grant (UNI-3D, grant no. 23-07973X).
\bibliographystyle{ACM-Reference-Format}
\bibliography{sample-bibliography}

\begin{appendix}
\section{Additional Implementation Details}
In this section, we describe our implementation details, including a minor modification to 3DGS's densification strategy to overcome the clustered issues resulting from clone operation and details for surface extraction.

\boldparagraph{Clone with Sampling}
We found that the position of Gaussian is relatively stable, which is also observed in Mip-Splatting~\cite{Yu2023MipSplatting}. Therefore, Gaussians cloned from the same parents remain clustered. To address this issue, instead of using the same position for the cloned Gaussian, we sample a new Gaussian according to the Gaussian's parameter similar to the procedures for Gaussian split~\cite{kerbl3Dgaussians}. We found this simple strategy leads to more uniformly distributed Gaussians, as shown in Figure~\ref{fig:clone}. However, we do not observe a significant impact in terms of NVS results.
\newcommand{\clonewidth}{0.24\textwidth}

\begin{figure*}[t]
    \centering
    \setlength{\tabcolsep}{0.1em}
    \renewcommand{\arraystretch}{0.4}
    \scriptsize
    \begin{tabular}{cccc}
    \includegraphics[width=\clonewidth]{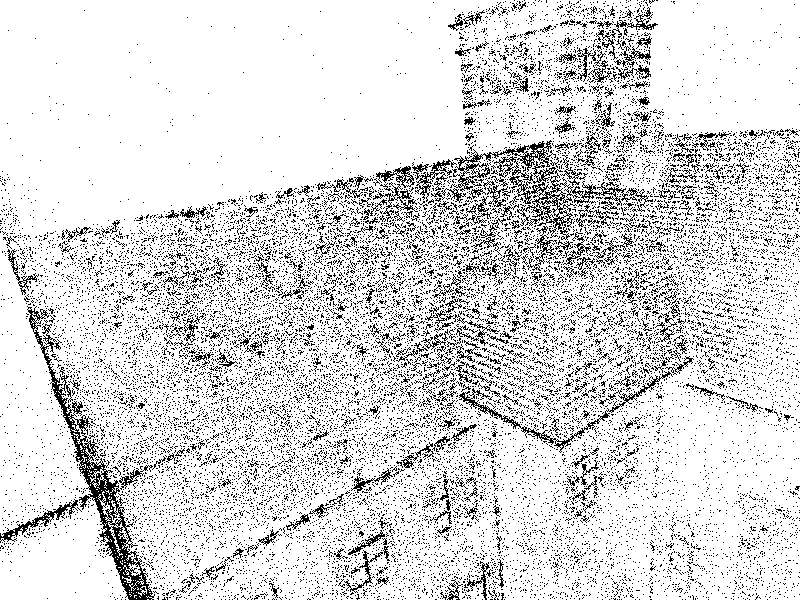}&    
    \includegraphics[width=\clonewidth]{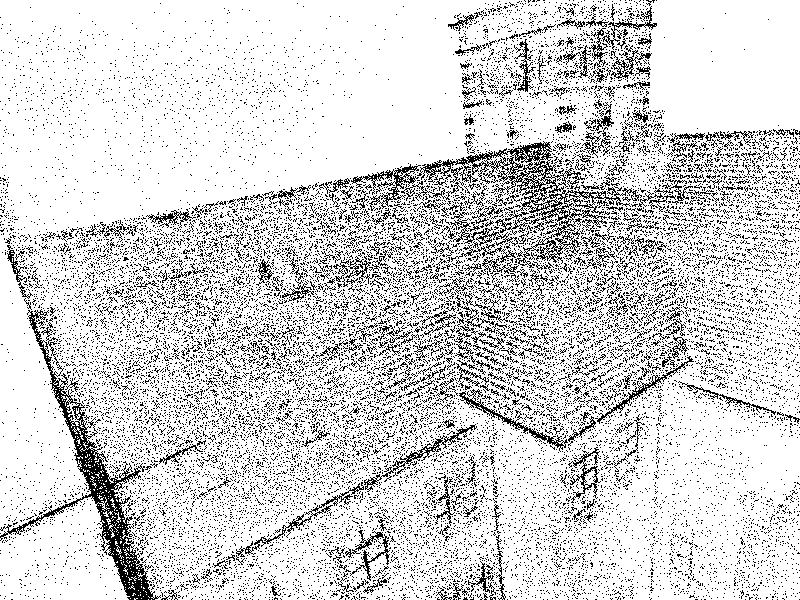}&    
    \includegraphics[width=\clonewidth]{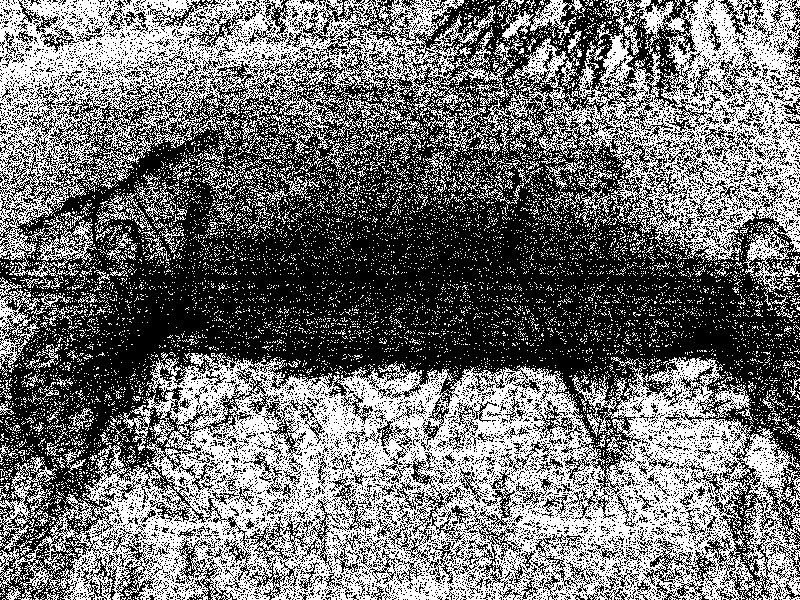}&  
    \includegraphics[width=\clonewidth]{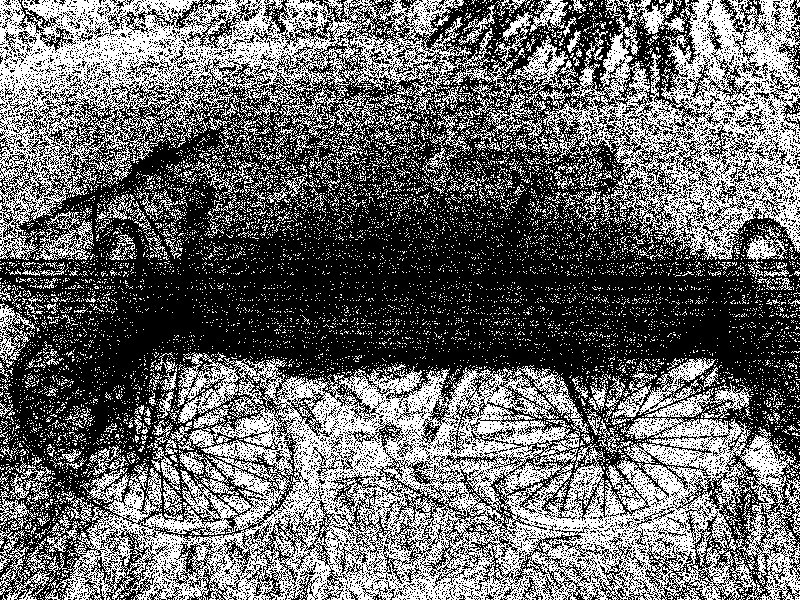} \\
    3DGS~\cite{kerbl3Dgaussians} & Ours & 3DGS & Ours~\cite{kerbl3Dgaussians} 
    \end{tabular}
    \vspace{-0.1in}
    \caption{\textbf{Comparison of clone strategy on the Mip-NeRF 360 Dataset~\cite{barron2022mipnerf360}.} 
    Our clone strategy leads to more uniformly distributed Gaussian primitives.
    }
    \label{fig:clone}
\end{figure*}

\boldparagraph{Details for Surface Extraction}
We provide a pseudo-code of our tile-based Gaussian Opacity Fields evaluation in Algorithm~\ref{alg:integrate} and Algorithm~\ref{alg:integrate_single}. Note that the evaluation takes 3D Gaussians, training views, and 3D points as input and it does not rely on the tetrahedra cells. Therefore, the same algorithm also applies to the Marching Cubes Algorithm. The pseudo-code of our Marching Tetrahedra augmented with binary search is shown in Algorithm~\ref{alg:marching_tetrahedra_binary}, the same idea could be applied to the Marching Cubes algorithm.

\begin{algorithm}
		\caption{Gaussian Opacity Field evaluation\\
			$M$, $S$, $A$: Gaussian means, covariances, and opacity\\
            $P$: position of points \\
            $V$: view configurations}
		\label{alg:integrate}
		\begin{algorithmic}
			\Function{EvaluateOpacityField}{$M$, $S$, $A$, $V$, $P$}

            \State $O \gets \mathbf{1}$ \Comment{Init Opacity}
            \ForAll{Views $v$ $\textbf{in}$ $V$}
            \State $w$, $h$ $\gets$ GetImageSize($V$)
            \State $O_v \gets$ EvaluateSingleView($w$, $h$, $M$, $S$, $A$, $V$, $P$)
            \State $O \gets$ Min($O$, $O_v$) \Comment{Take Minimal Opacity}
            \EndFor
			
			\Return $O$
			\EndFunction
			
		\end{algorithmic}
\end{algorithm}

\begin{algorithm}
		\caption{Gaussian Opacity Field evaluation for a single view\\
			$w$, $h$: width and height of training image\\
			$M$, $S$, $A$: Gaussian means, covariances, and opacity\\
            $P$: position of points \\
            $V$: view configuration of current camera}
		\label{alg:integrate_single}
		\begin{algorithmic}
			\Function{EvaluateSingleView}{$w$, $h$, $M$, $S$, $A$, $V$, $P$}
			
			\State CullGaussian($M$, $V$) \Comment{Frustum Culling}
			\State $M', S'$ $\gets$ ScreenspaceGaussians($M$, $S$, $V$) \Comment{Transform}
			\State $T_g$ $\gets$ CreateTiles($w$, $h$)
			\State $L_g$, $K_g$ $\gets$ DuplicateWithKeys($M'$, $T_g$) \Comment{Indices and Keys}
			\State SortByKeys($K_g$, $L_g$)							\Comment{Globally Sort}
			\State $R_g$ $\gets$ IdentifyTileRanges($T_g$, $K_g$)
		    \State CullPoints($P$, $V$) \Comment{Frustum Culling}
			\State $P'$ $\gets$ ScreenspacePoints($P$, $V$) \Comment{Transform}
			\State $T_p$ $\gets$ CreateTiles($w$, $h$)
			\State $L_p$, $K_p$ $\gets$ CreateWithKeys($P'$, $T_p$) \Comment{Indices and Keys}
			\State SortByKeys($K_p$, $L_p$)							\Comment{Globally Sort}
			\State $R_p$ $\gets$ IdentifyTileRanges($T_p$, $K_p$)
   
			\State $O \gets \mathbf{1}$ \Comment{Init Opacity}
			
			\ForAll{Tiles $t$ $\textbf{in}$ $I$} \Comment{$I$ is the Canvas}
			\ForAll{Pixels $i$ $\textbf{in}$ $t$}
			
			\State $r_g \gets$ GetTileRange($R_g$, $t$)
			\State $L_g' \gets$ FilterGaussians($i$, $L_p$, $r_g$, $K_g$, $M$, $S$, $A$) \Comment{Select Gaussians that contributes to pixel $i$}
   
            \State $r_p \gets$ GetTileRange($R_p$, $t$)
            \State $L_p' \gets$ FilterPoints($i$, $L_p$, $r_p$, $K_p$, $P$) \Comment{Select Points that projected to pixel $i$}

            \ForAll{Points $p$ $\textbf{in}$ $L_p'$}
            \State $O[p] \gets$ EvaluateOpacity($i$, $L_g'$, $K_g$, $M$, $S$, $A$)
            \EndFor
			\EndFor
			\EndFor

			\Return $O$
			\EndFunction
			
		\end{algorithmic}
\end{algorithm}

\begin{algorithm}
		\caption{Marching Tetrahedra with Binary Search \\
			$M$, $S$, $A$: Gaussian means, covariances, and opacity\\
            $P$, $C$: Tetrahedra Points and Cells \\
            $V$: view configurations \\
            $L$: Level set value \\
            $N$: Step of binary search}
		\label{alg:marching_tetrahedra_binary}
		\begin{algorithmic}
			\Function{MarchingTetrahedraBinarySearch}{$M$, $S$, $A$, $P$, $C$, $V$}

            \State $O \gets$ EvaluateOpacityField($M$, $S$, $A$, $V$, $P$)
            \State $F$, $P_1$, $P_2$, $O_1$, $O_2 \gets$ MarchingTetrahedra($P$, $C$, $O$, $S$)
            
            \ForAll{i $v$ $\textbf{in}$ $\{1,2,...,N\}$}
                \State $P_m \gets$ MidPoint($P_1$, $P_2$)
                \State $O_m \gets$ EvaluateOpacityField($M$, $S$, $A$, $V$, $P_m$)
                \State $P_1$, $P_2$, $O_1$, $O_2 \gets$ ChangeEndPointsAndValues($P_1$, $P_2$, $O_1$, $O_2$, $L$,$P_m$, $O_m$)
            \EndFor
			\State $V \gets$ LinearInterpolate($P_1$, $P_2$, $O_1$, $O_2$, $L$)
			\Return $F$, $V$ \Comment{return faces and vertices for triangle mesh}
			\EndFunction
			
		\end{algorithmic}
\end{algorithm}

\section{Additional Results}

\newcommand{\unbounedwidth}{0.24\textwidth}

\begin{figure*}[t]
    \centering
    \setlength{\tabcolsep}{0.1em}
    \renewcommand{\arraystretch}{0.4}
    \scriptsize
    \begin{tabular}{cccc}
    \includegraphics[width=\unbounedwidth]{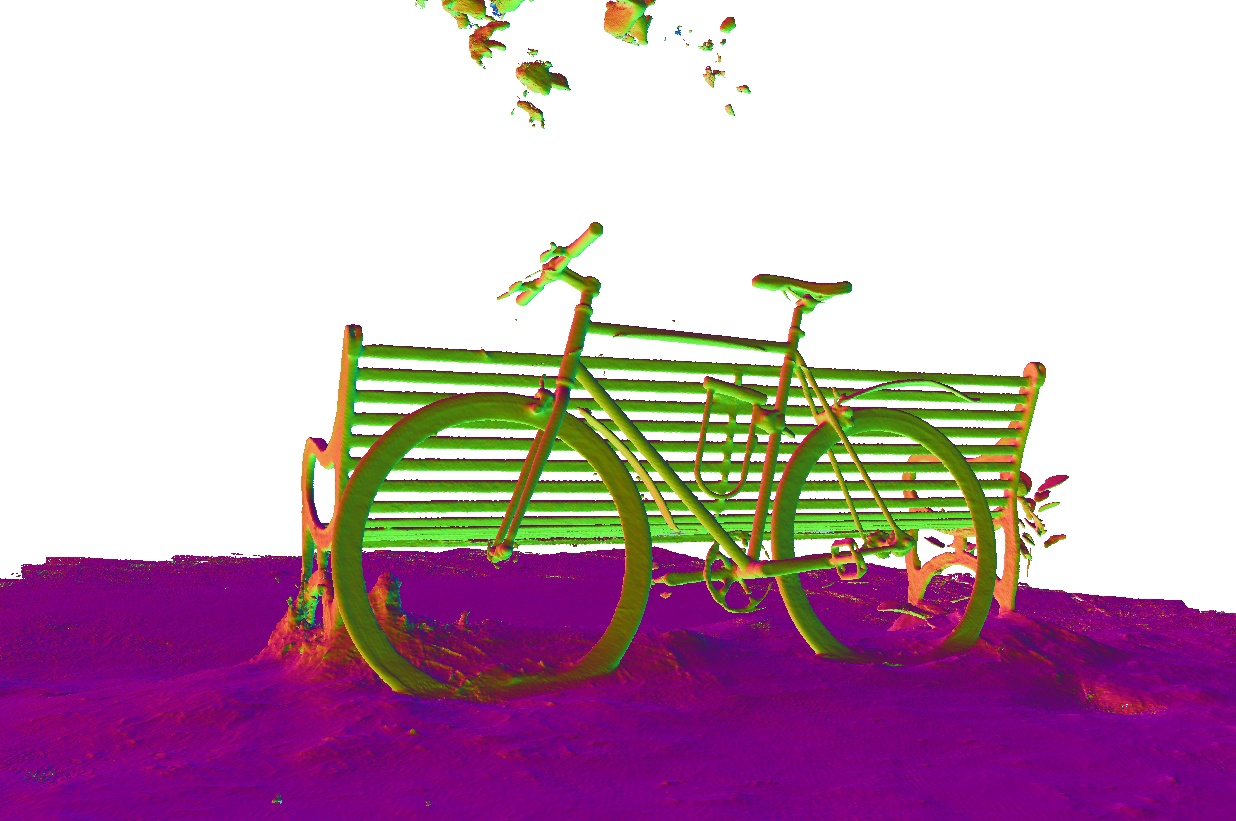}&    
    \includegraphics[width=\unbounedwidth]{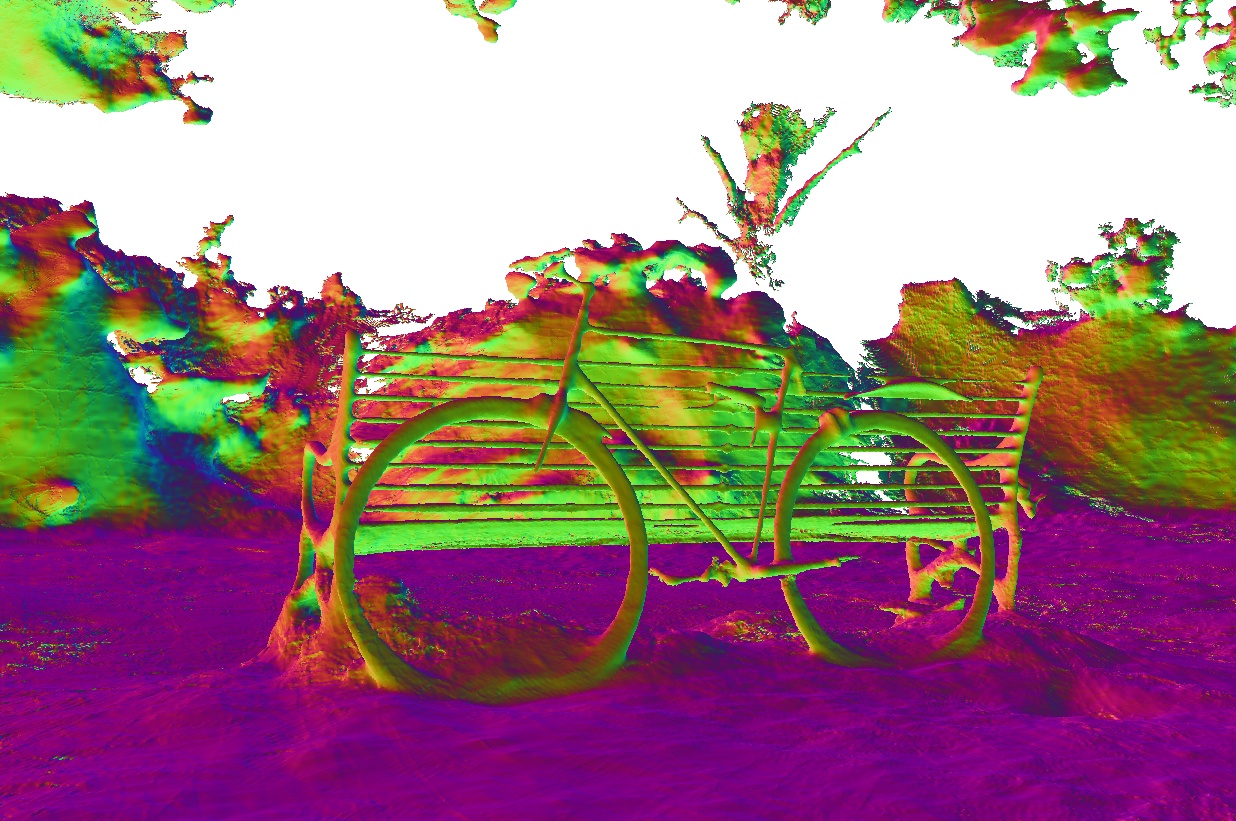}&    
    \includegraphics[width=\unbounedwidth]{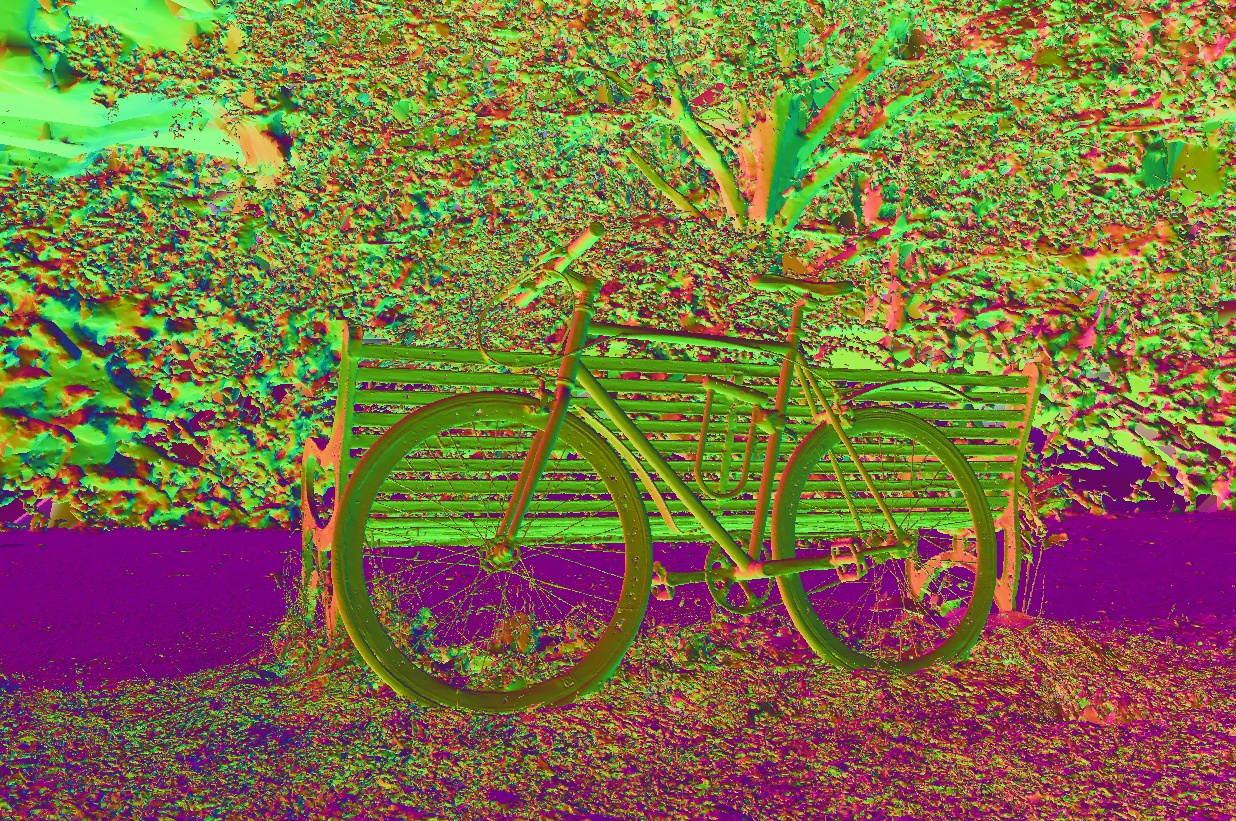}&  
    \includegraphics[width=\unbounedwidth]{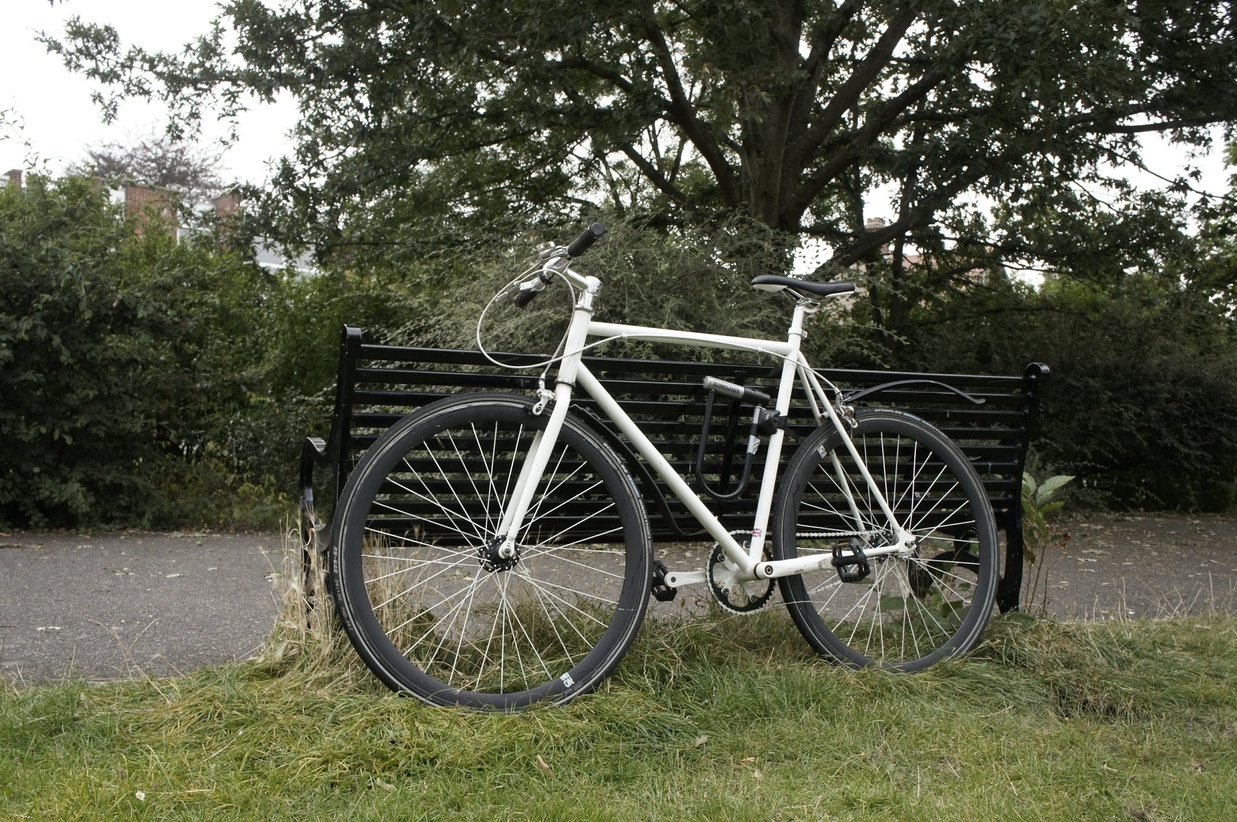} \\
    \includegraphics[width=\unbounedwidth]{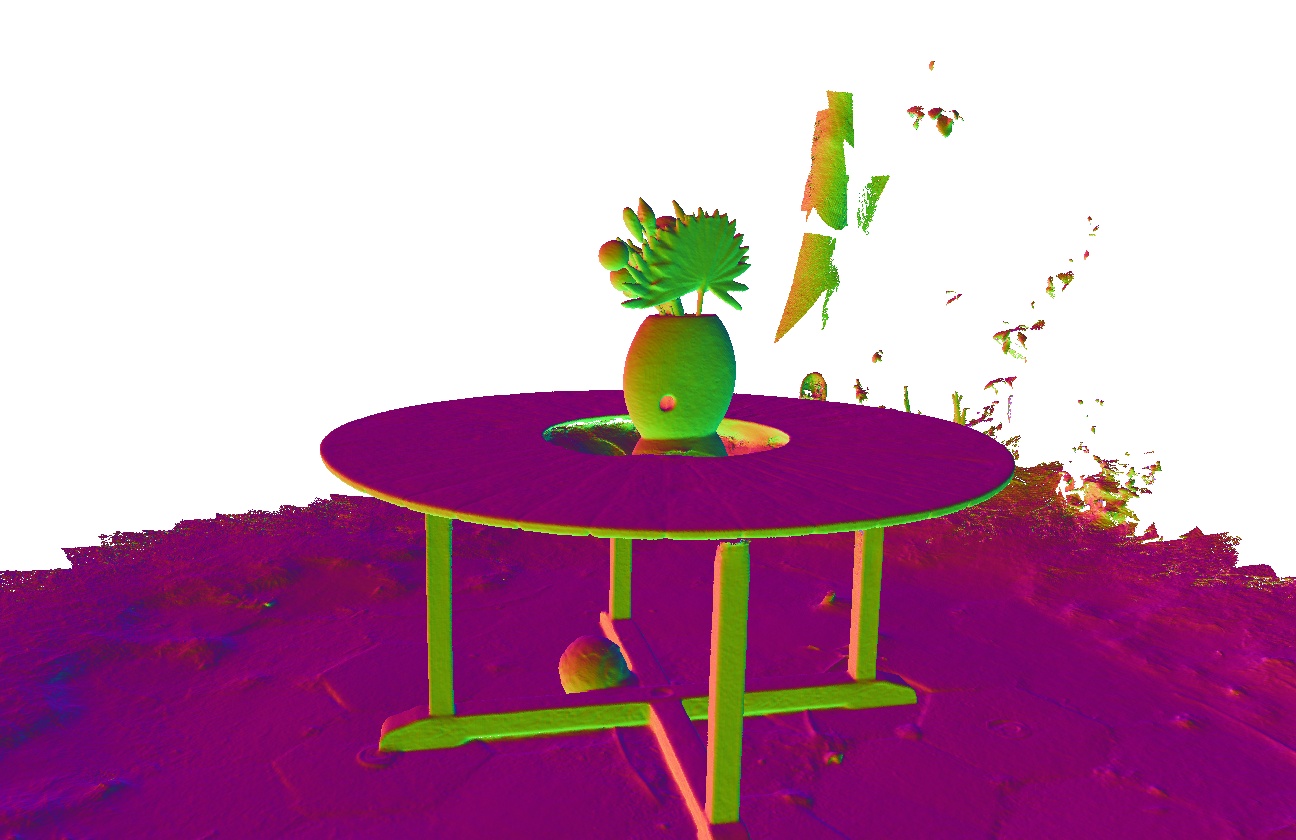}&    
    \includegraphics[width=\unbounedwidth]{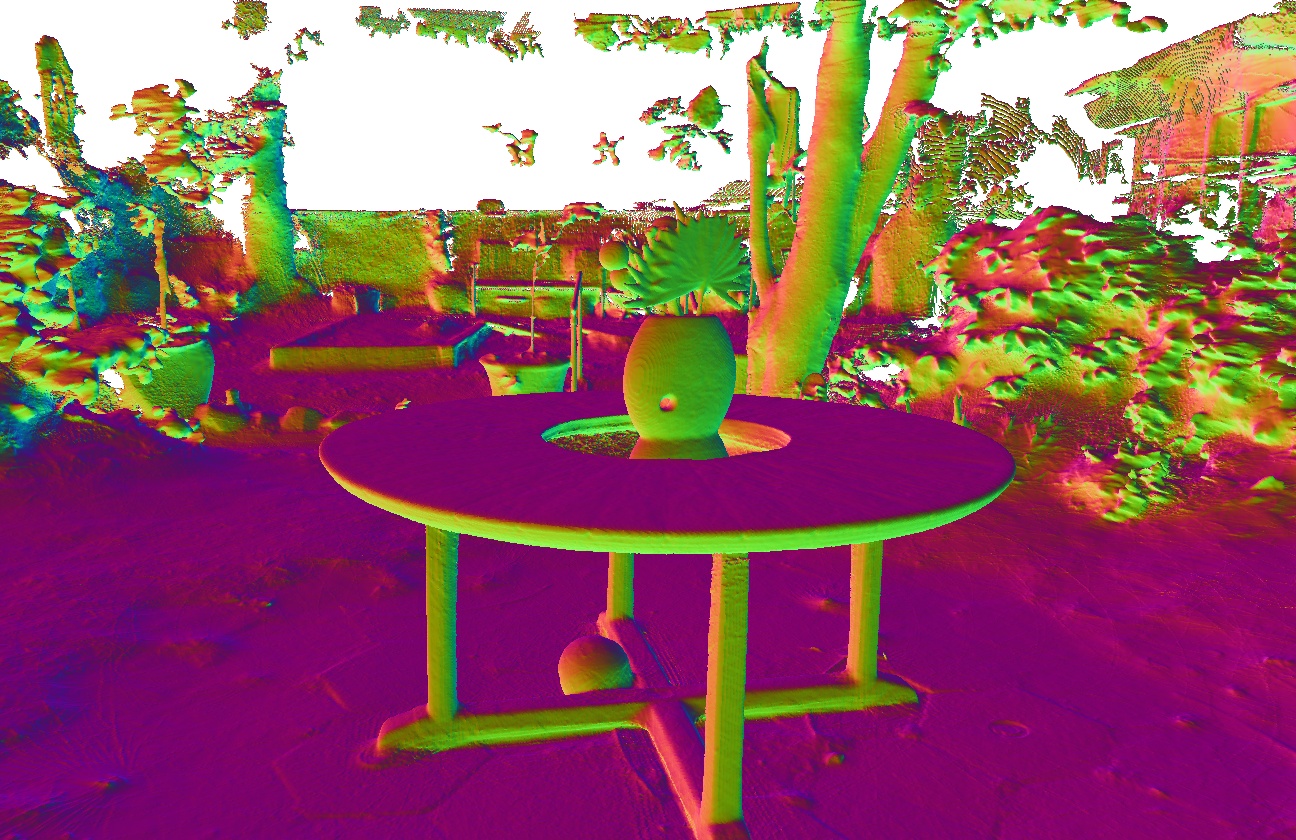}&    
    \includegraphics[width=\unbounedwidth]{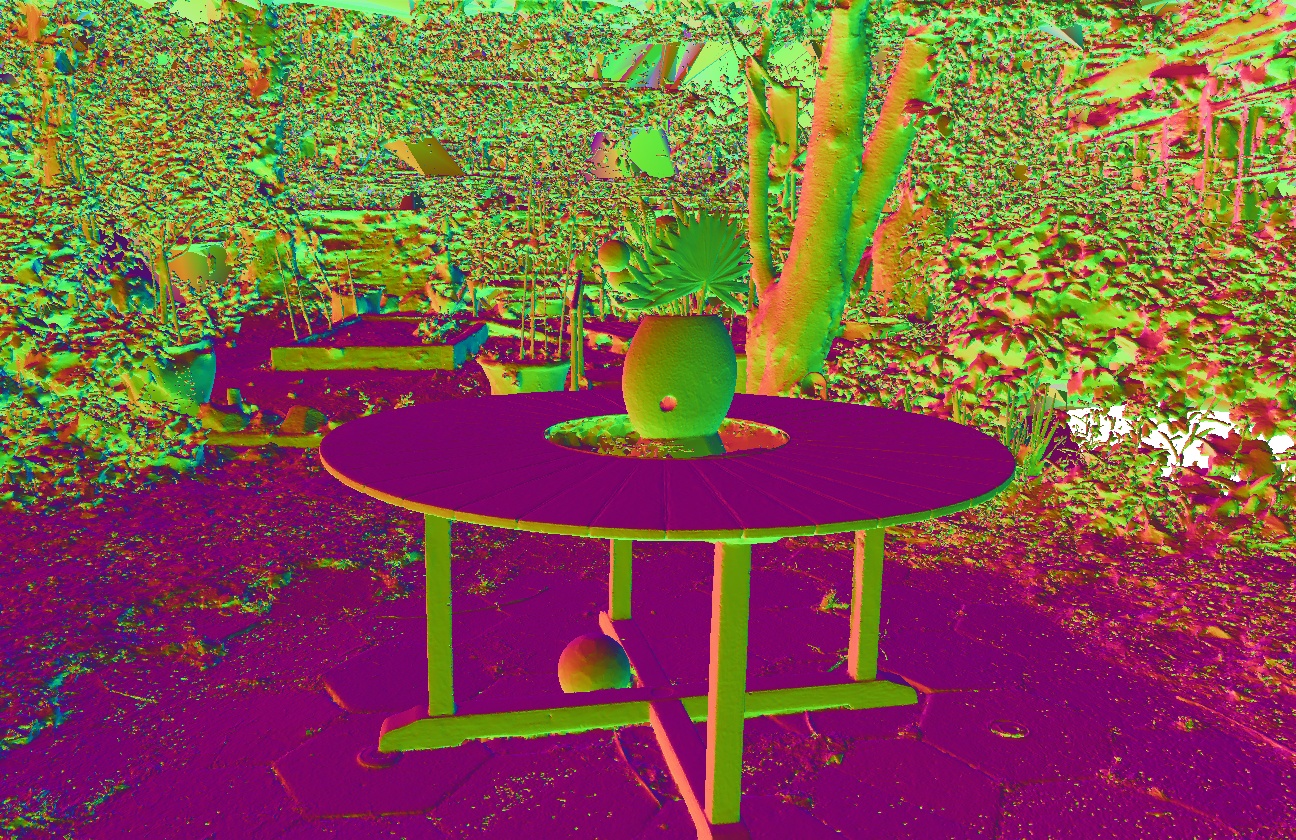}&  
    \includegraphics[width=\unbounedwidth]{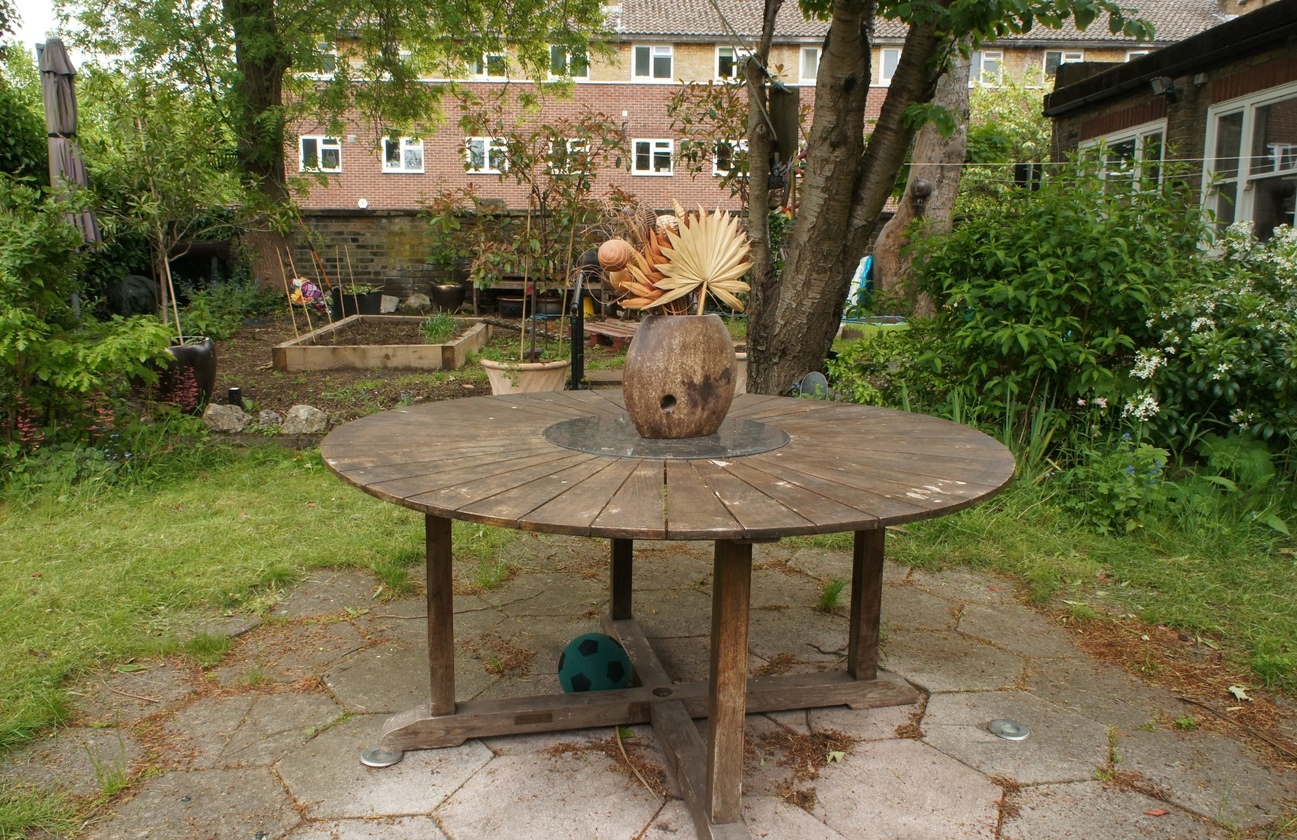} \\
    \includegraphics[width=\unbounedwidth]{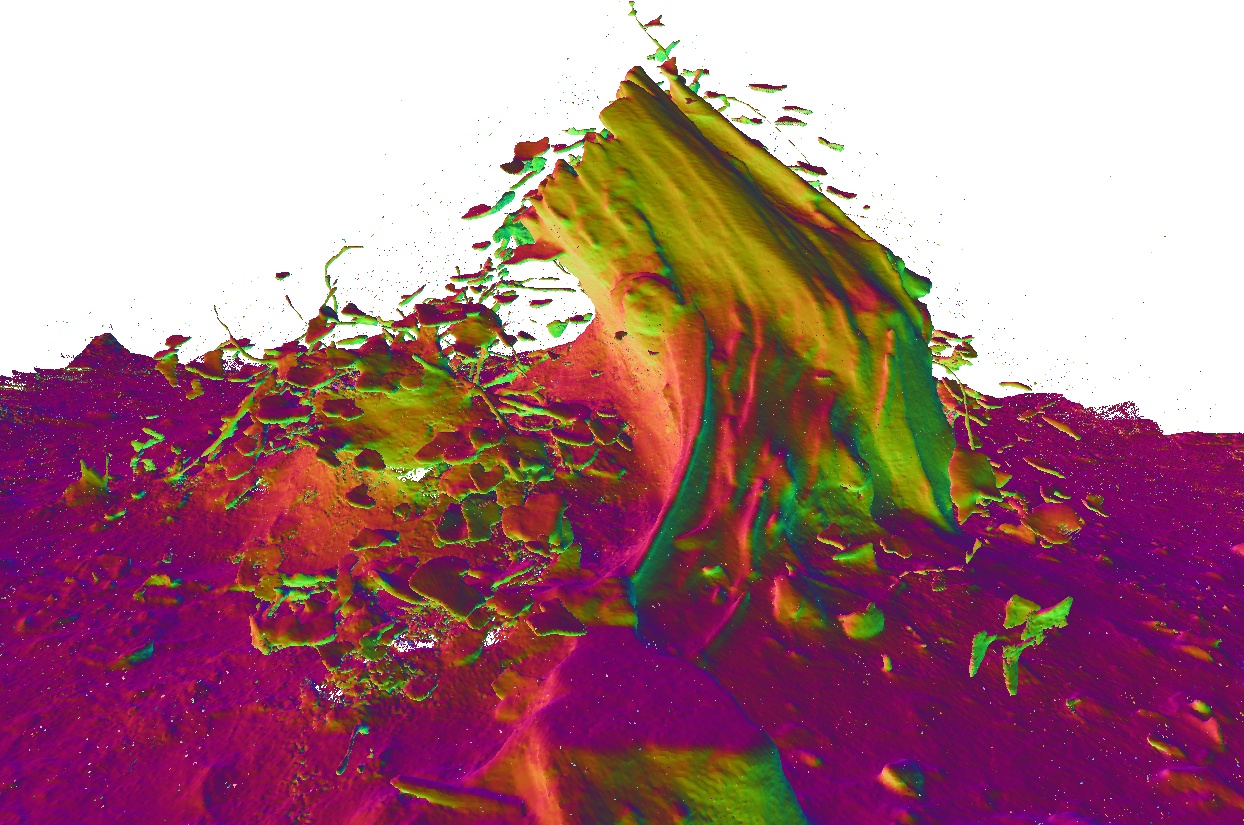}&    
    \includegraphics[width=\unbounedwidth]{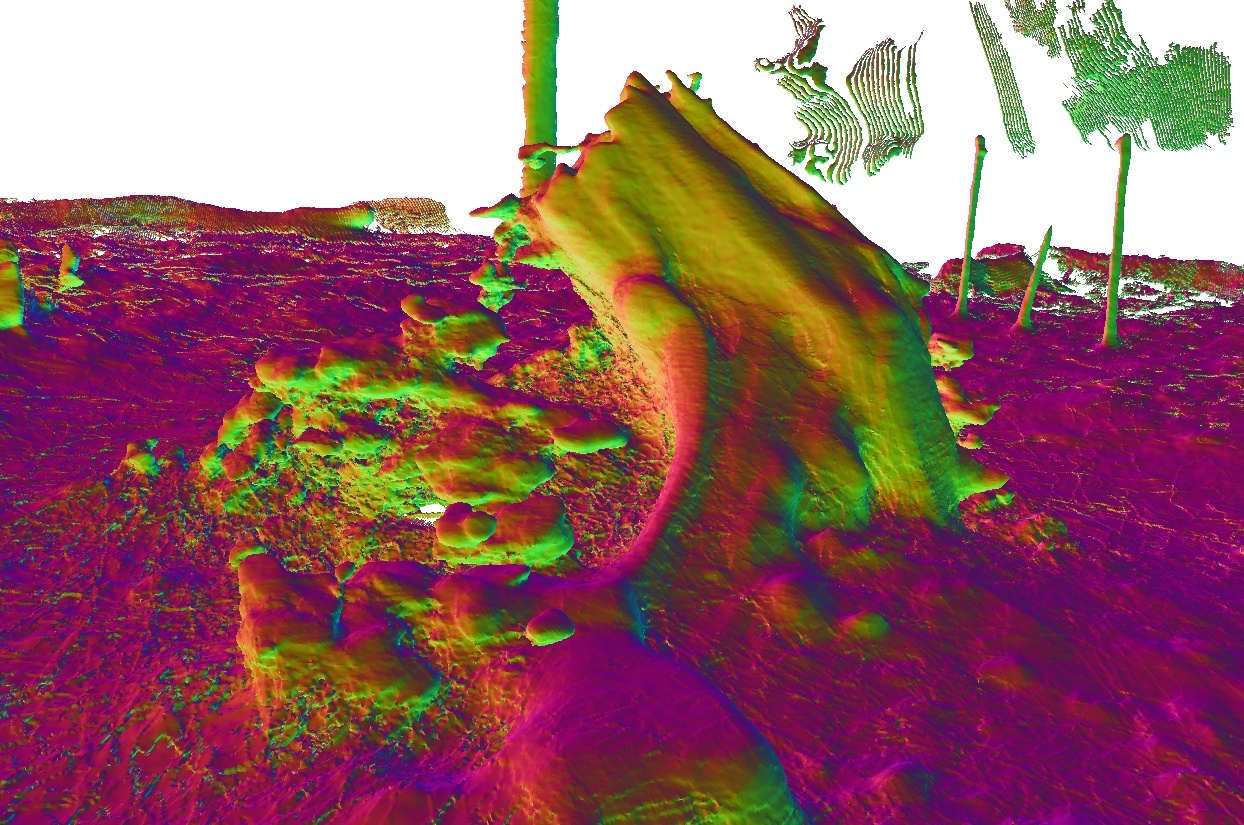}&    
    \includegraphics[width=\unbounedwidth]{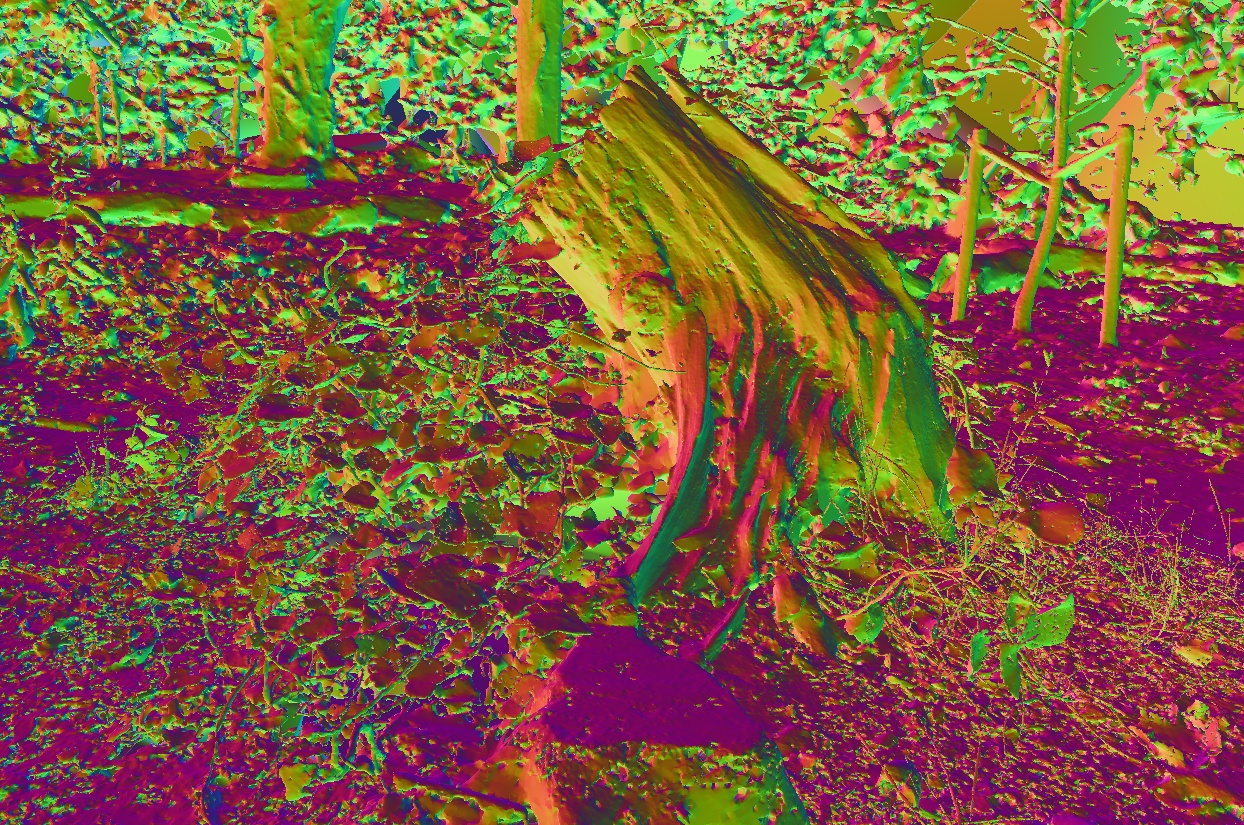}&  
    \includegraphics[width=\unbounedwidth]{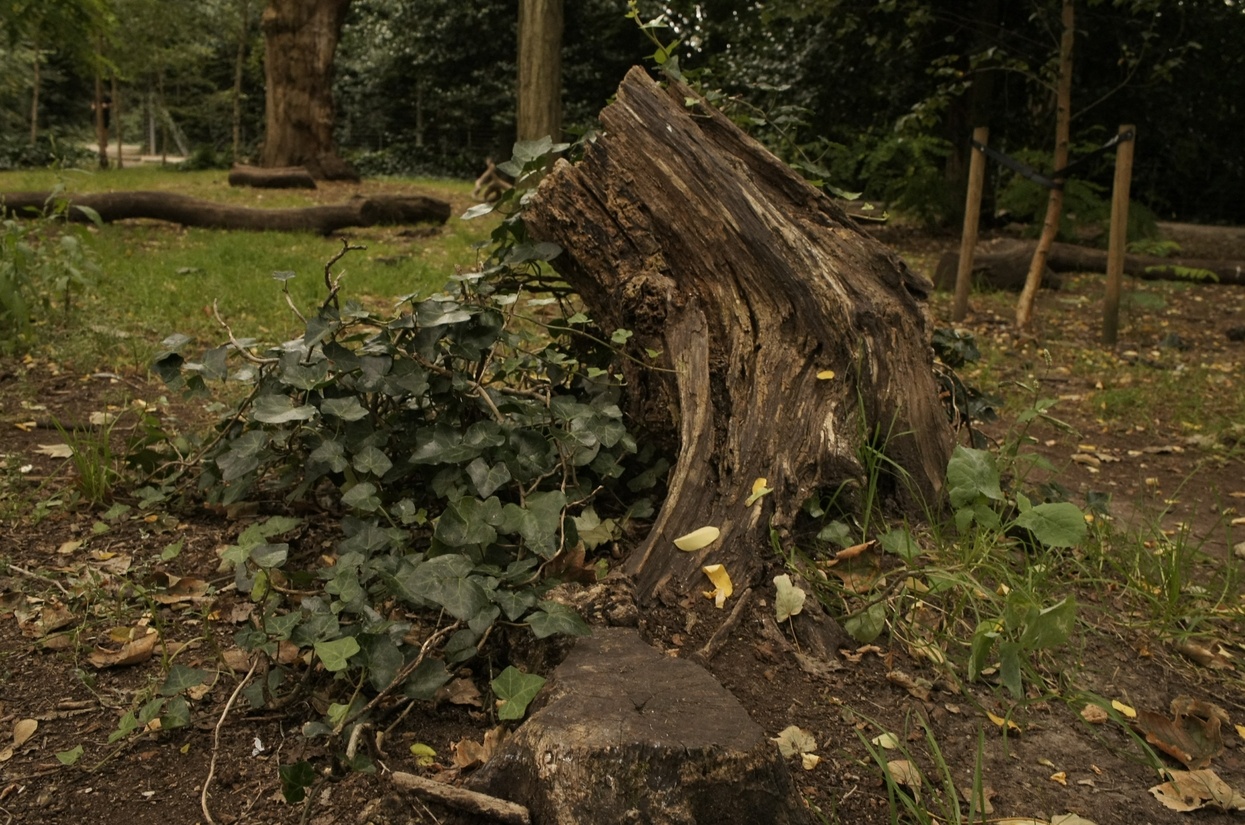} \\
    \includegraphics[width=\unbounedwidth]{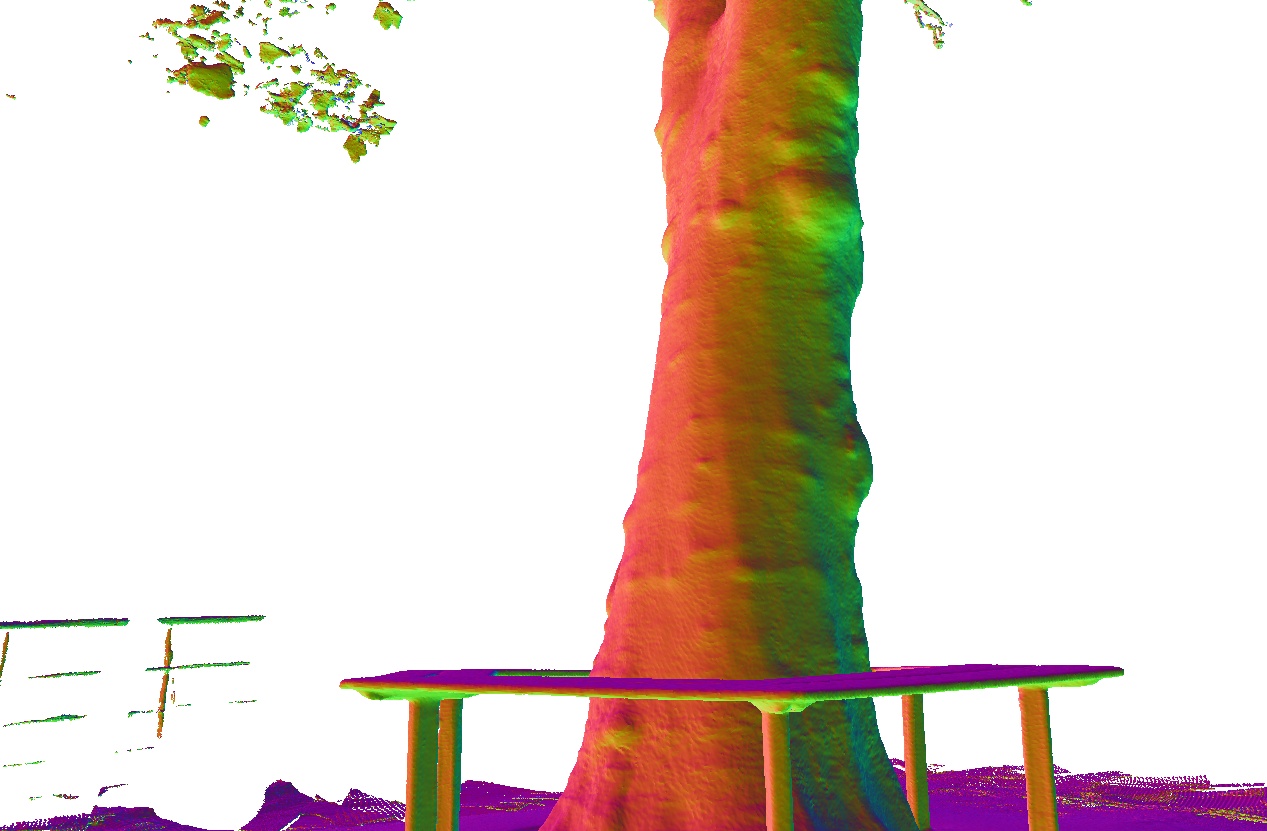}&    
    \includegraphics[width=\unbounedwidth]{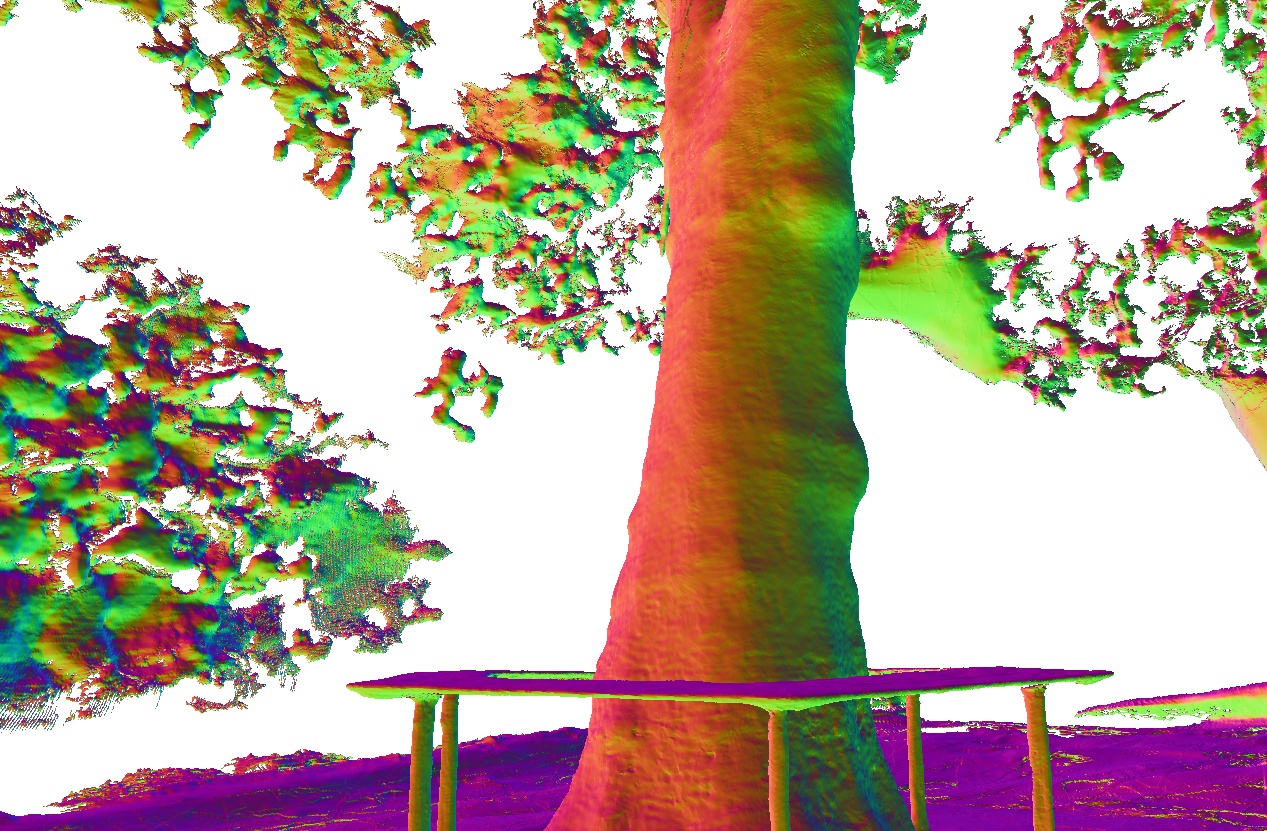}&    
    \includegraphics[width=\unbounedwidth]{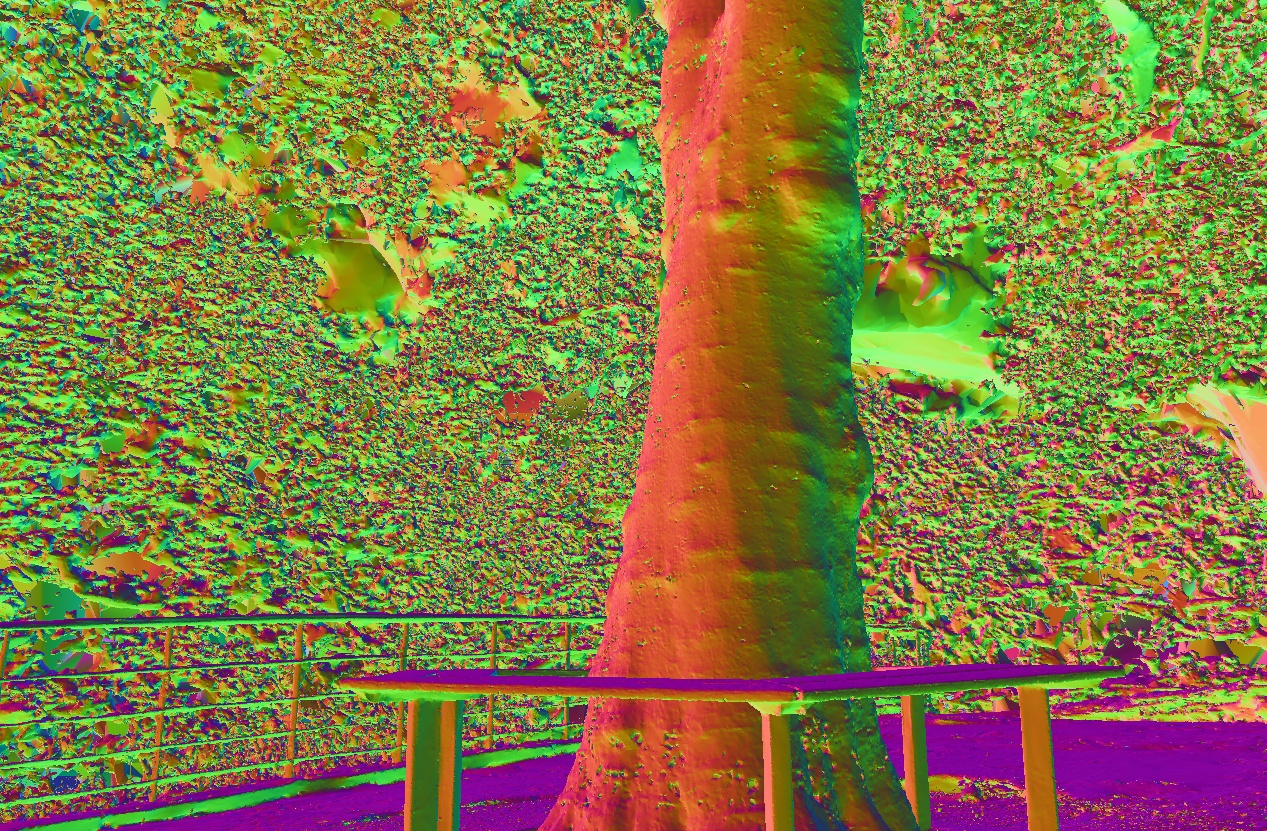}&  
    \includegraphics[width=\unbounedwidth]{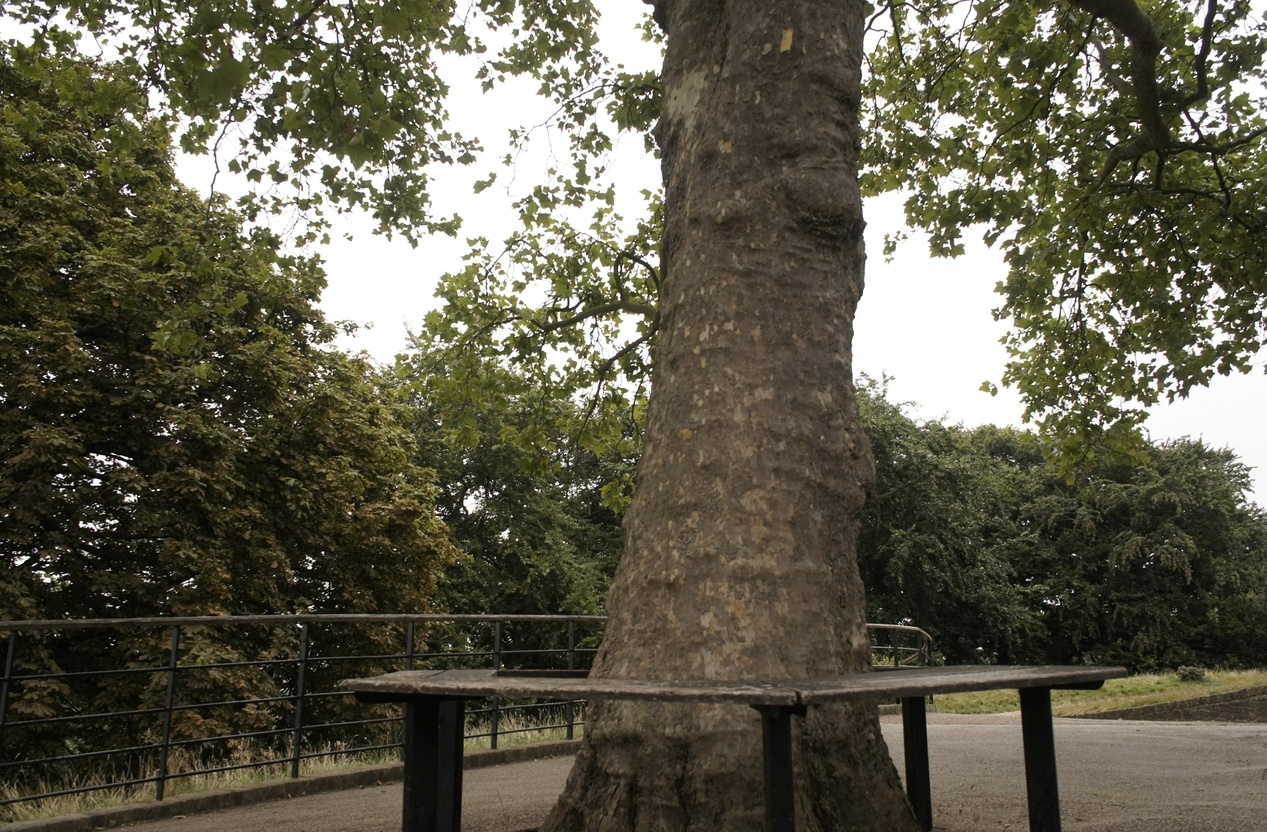} \\
    2DGS~\cite{Huang2024ARXIV} & 2DGS~\cite{Huang2024ARXIV} Unbounded & Ours & GT
    \end{tabular}
    \vspace{-0.1in}
    \caption{\textbf{Reconstructions on the Mip-NeRF 360 Dataset~\cite{barron2022mipnerf360}}. We compare GOF with both 2DGS's bounded and unbounded mesh extraction. While 2DGS's unbounded setting can reconstruct mesh for the background region, its meshes are incomplete and lack of details. By contrast, GOF can reconstruct detailed meshes both for the foreground objects and background regions.}
    \label{fig:2dgs_unbounded}
    \vspace{-0.1cm}
\end{figure*}

\boldparagraph{TSDF Fusion in Contraction Space}
In the main paper, we compare GOF with 2DGS~\cite{Huang2024ARXIV} with its default (bounded) settings. Specifically, 2DGS extracts meshes with TSDF fusion focusing on the foreground objects, resulting in missing geometry in the background regions. However, 2DGS also supports extracting meshes for the background regions by applying TSDF fusion in the contraction space~\cite{barron2022mipnerf360}, which is referred as the unbounded setting. We extract unbounded meshes for 2DGS with a grid resolution of 2048 and the resulting meshes have similar number of vertices and faces with GOF's meshes. Then we compare GOF with both 2DGS's bounded and unbounded meshes. As shown in Figure~\ref{fig:2dgs_unbounded}, while 2DGS's unbounded setting can reconstruct meshes for the background regions, the meshes on the background regions are incomplete and lack details. By contrast, our method can reconstruct detailed surfaces for both foreground objects and background regions.

\end{appendix}
\newcommand{\prowidth}{0.32\textwidth}

\begin{figure*}[t]
    \centering
    \setlength{\tabcolsep}{0.1em}
    \renewcommand{\arraystretch}{0.4}
    \scriptsize
    \begin{tabular}{ccc}
    \includegraphics[width=\prowidth]{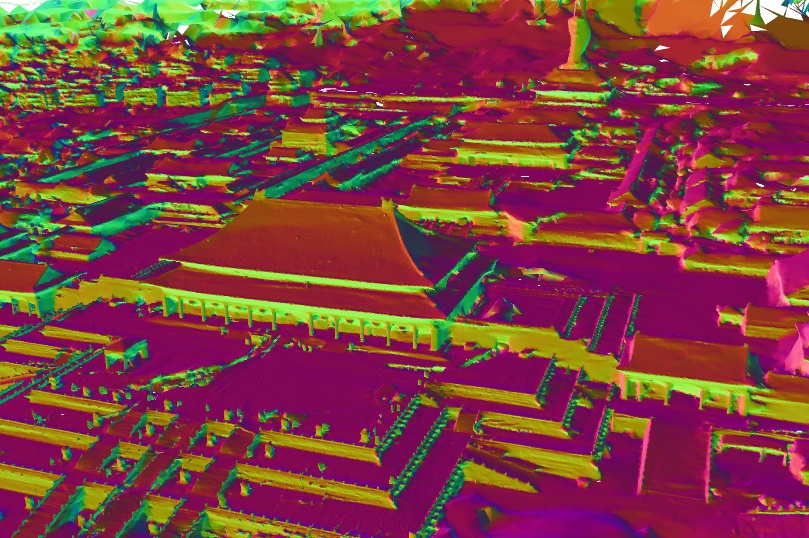}&
    \includegraphics[width=\prowidth]{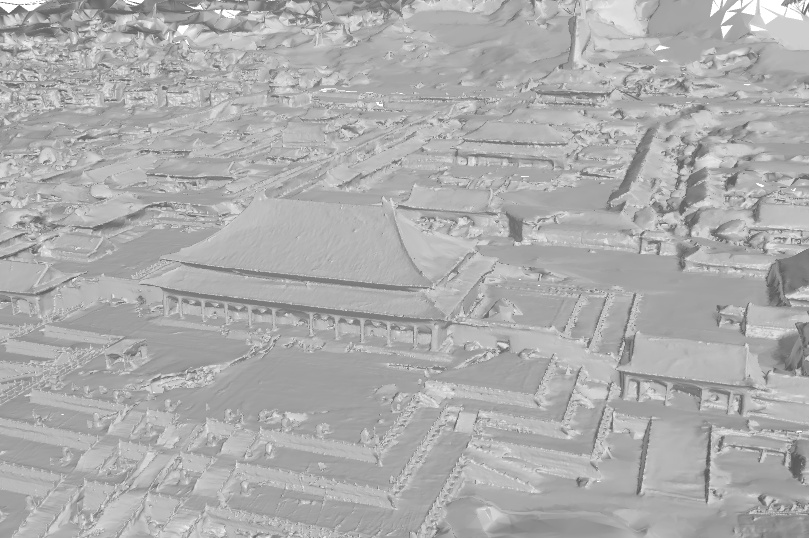}&
    \includegraphics[width=\prowidth]{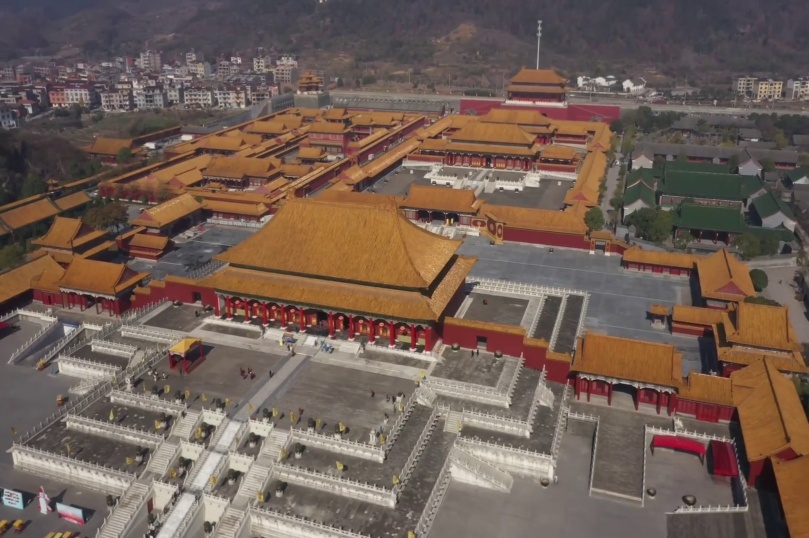}\\
    \includegraphics[width=\prowidth]{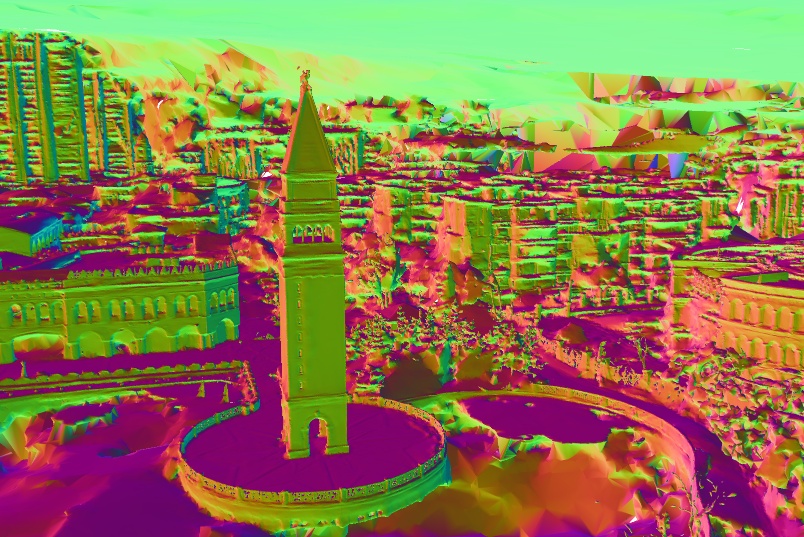}&
    \includegraphics[width=\prowidth]{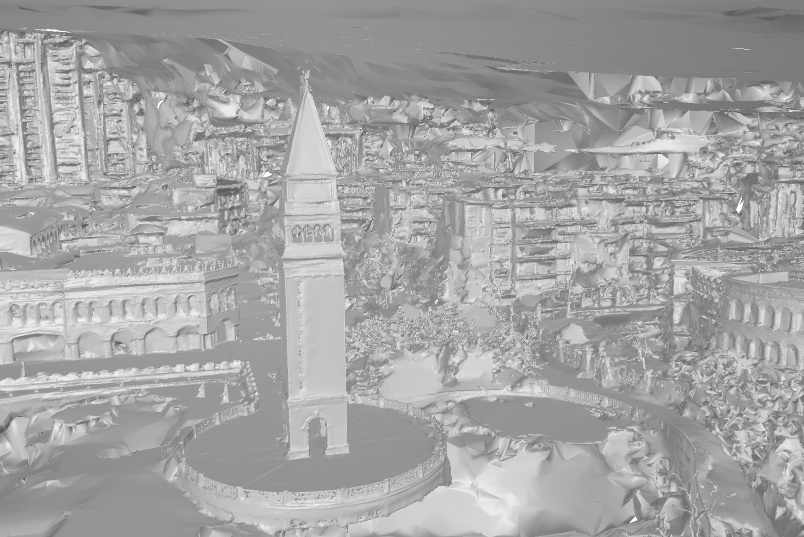}&
    \includegraphics[width=\prowidth]{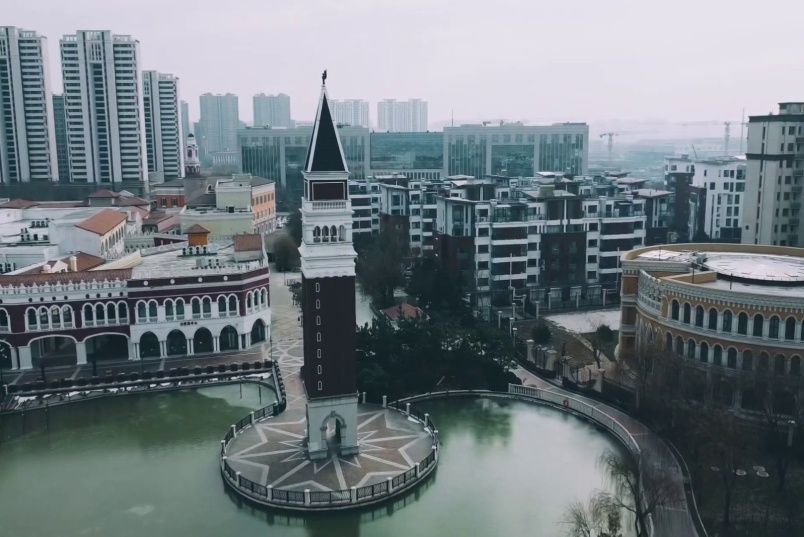}\\
    \includegraphics[width=\prowidth]{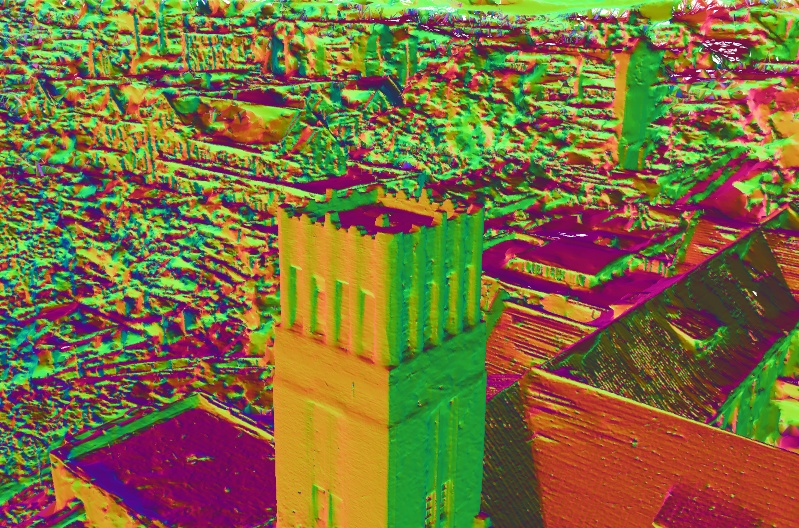}&
    \includegraphics[width=\prowidth]{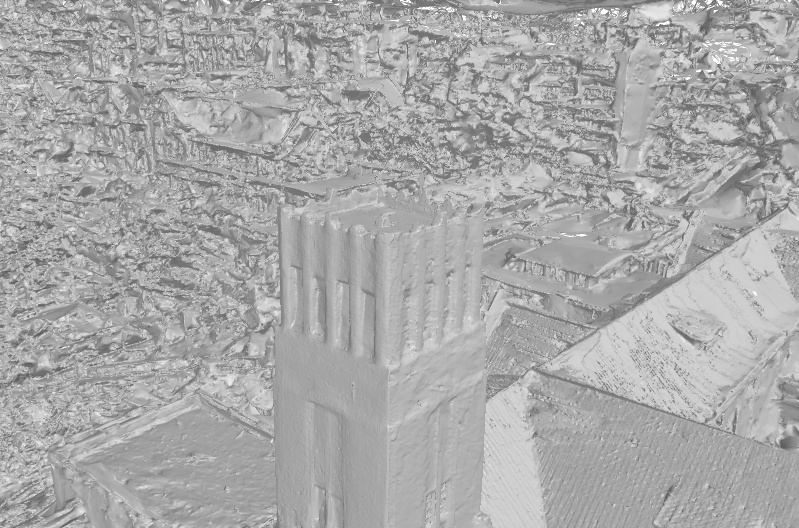}&
    \includegraphics[width=\prowidth]{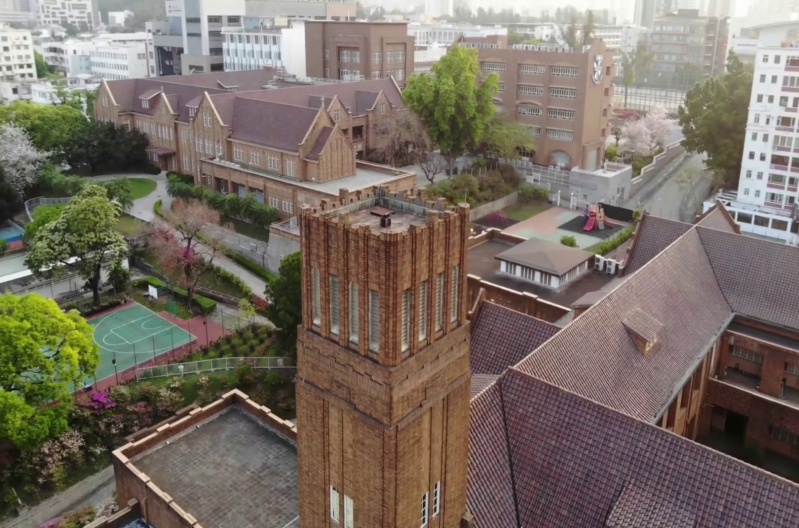}\\
    \includegraphics[width=\prowidth]{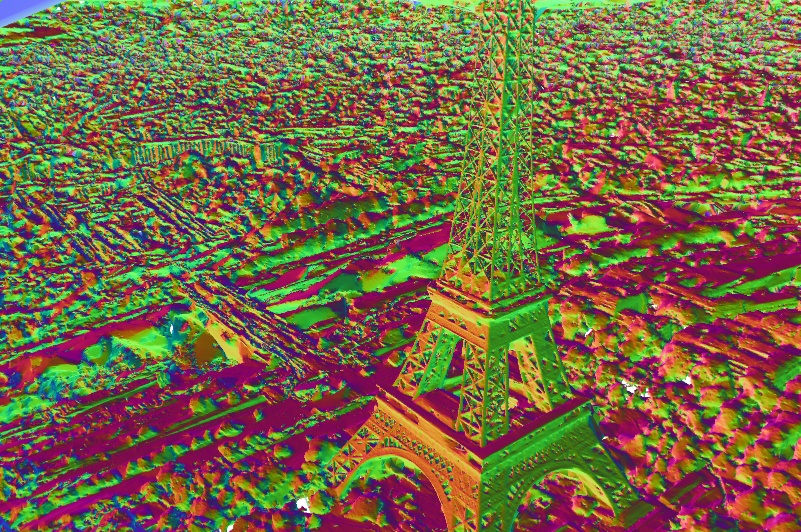}&
    \includegraphics[width=\prowidth]{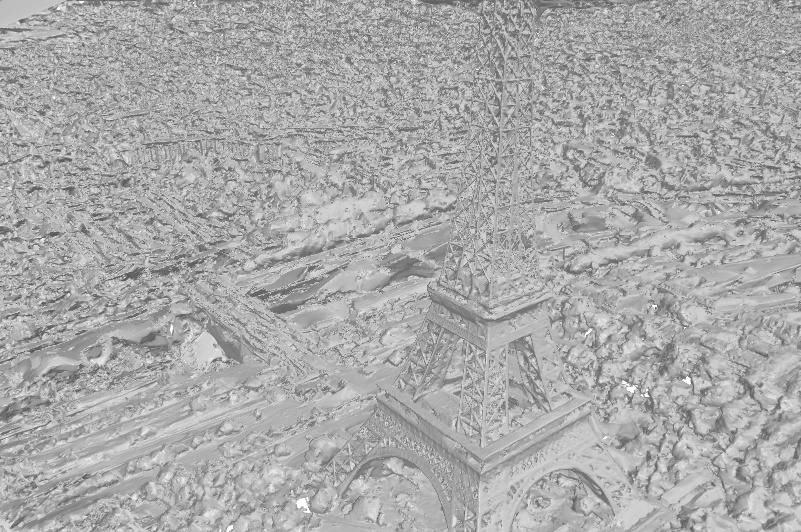}&
    \includegraphics[width=\prowidth]{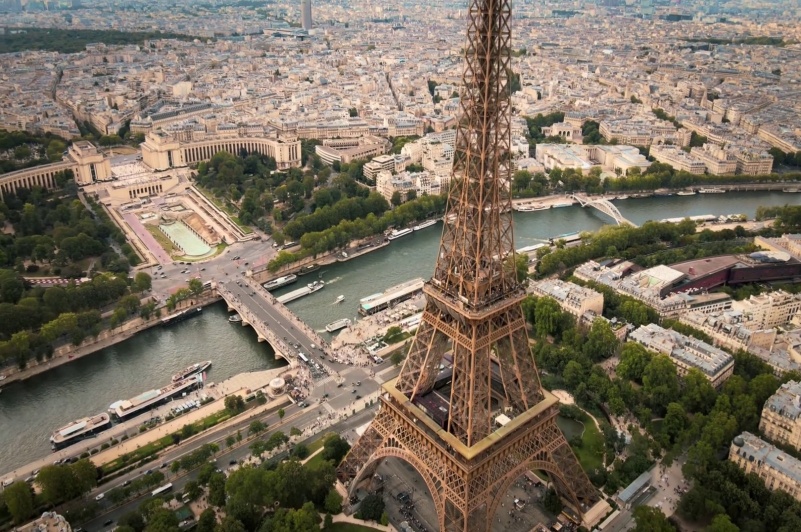}\\
    \includegraphics[width=\prowidth]{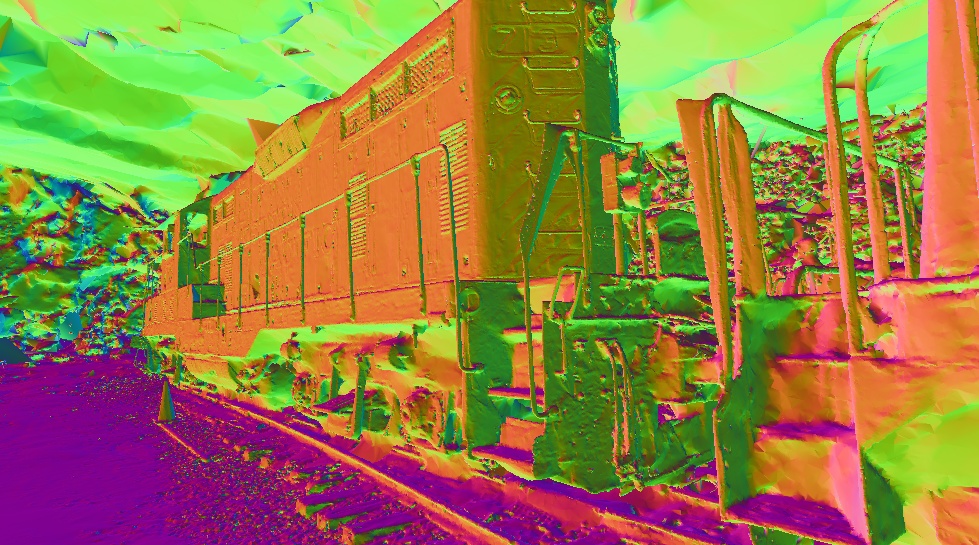}&
    \includegraphics[width=\prowidth]{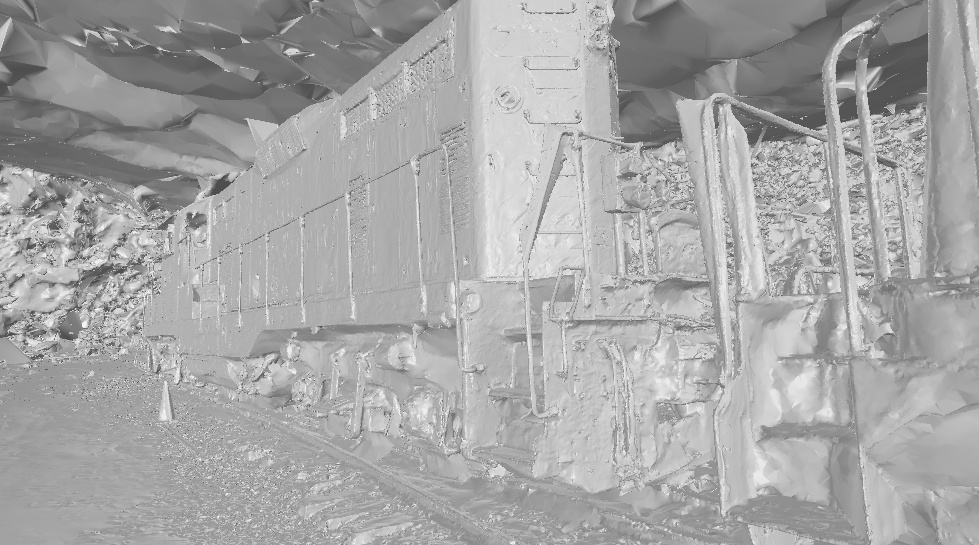}&
    \includegraphics[width=\prowidth]{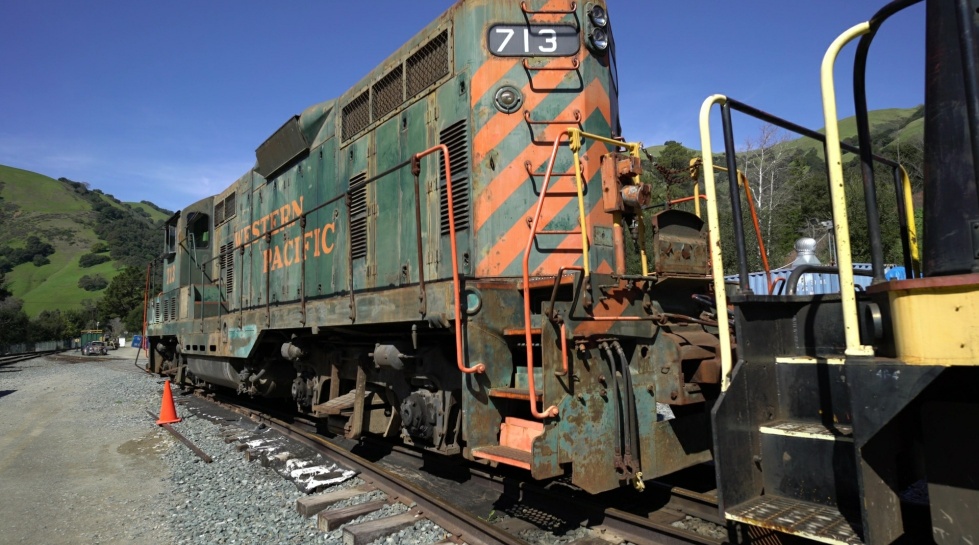}\\
    \includegraphics[width=\prowidth]{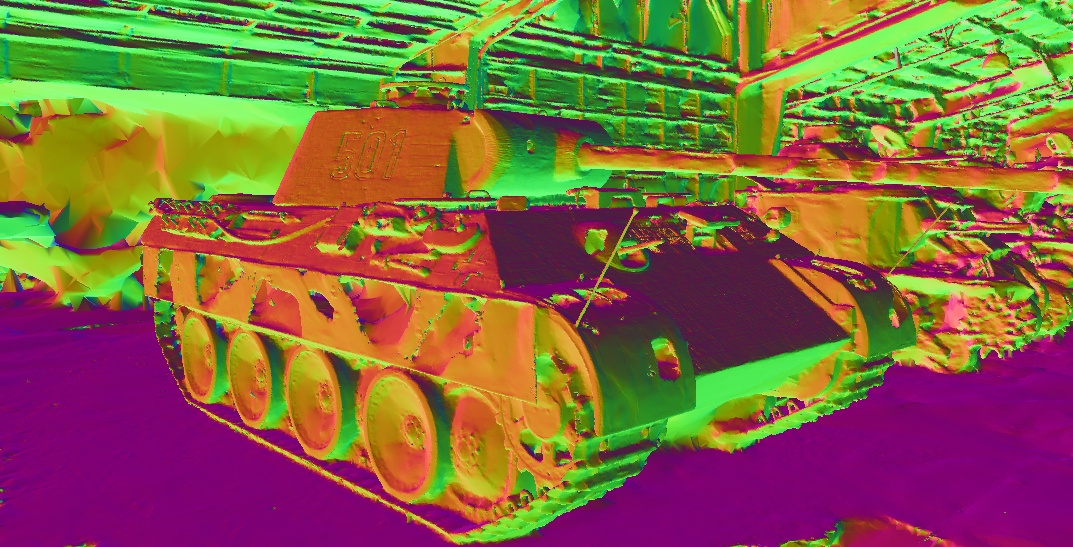}&
    \includegraphics[width=\prowidth]{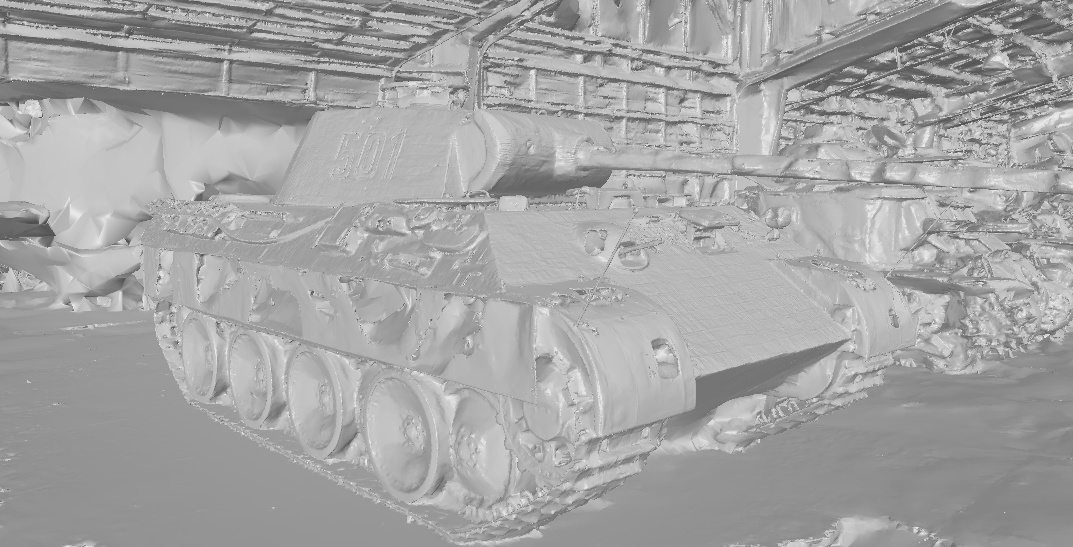}&
    \includegraphics[width=\prowidth]{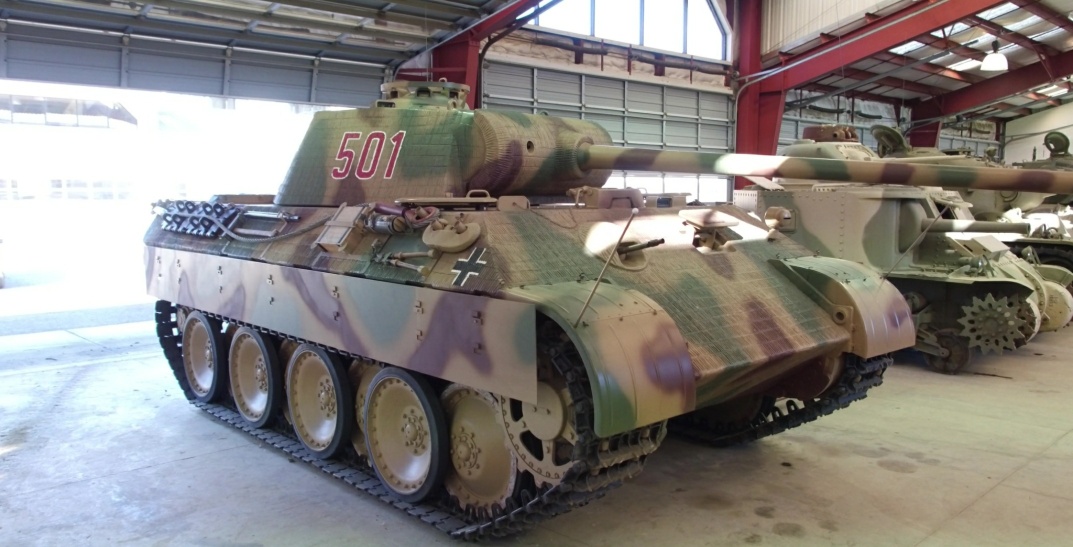}\\
    Normal & Mesh &  GT
    \end{tabular}
    \vspace{-0.15in}
    \caption{\textbf{Reconstructions on the GaussianPro~\cite{cheng2024gaussianpro} and the Tanks and Temples dataset~\cite{Knapitsch2017}}.}
    \label{fig:pro}
    \vspace{-0.1cm}
\end{figure*}

\newcommand{\ourwidth}{0.32\textwidth}

\begin{figure*}[t]
    \centering
    \setlength{\tabcolsep}{0.1em}
    \renewcommand{\arraystretch}{0.4}
    \scriptsize
    \begin{tabular}{ccc}
    \includegraphics[width=\ourwidth]{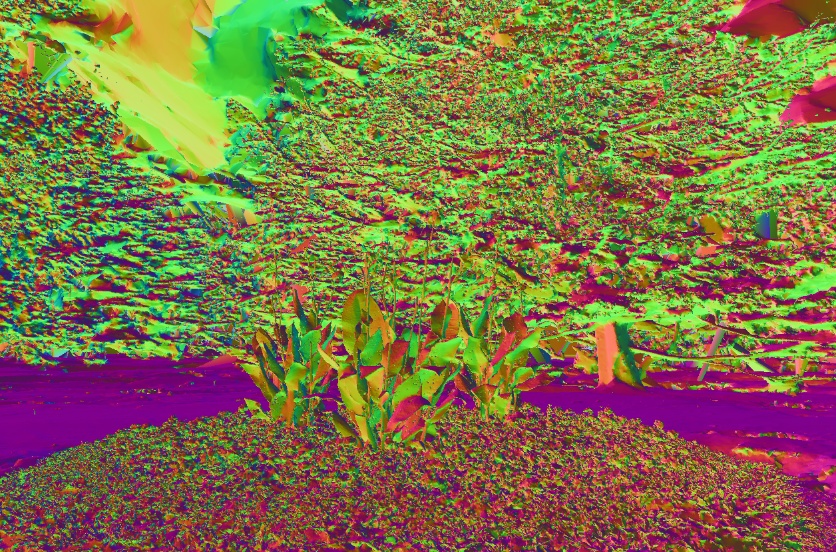}&
    \includegraphics[width=\ourwidth]{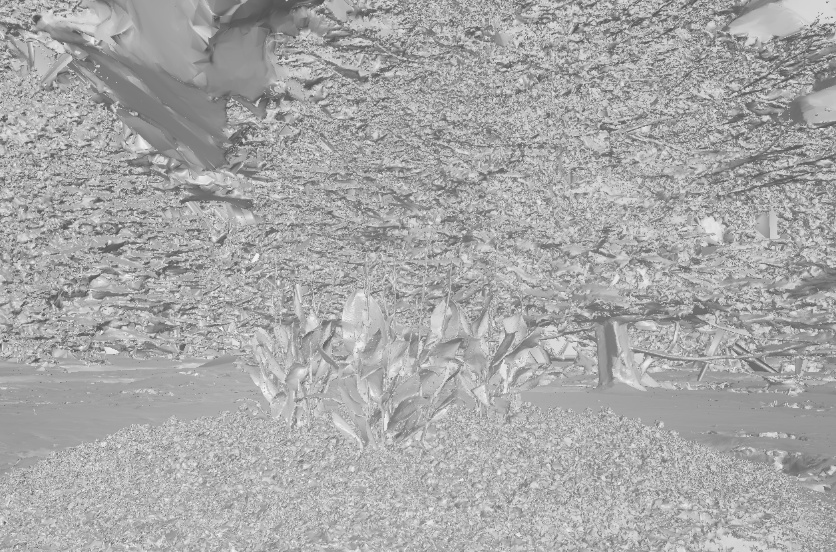}&
    \includegraphics[width=\ourwidth]{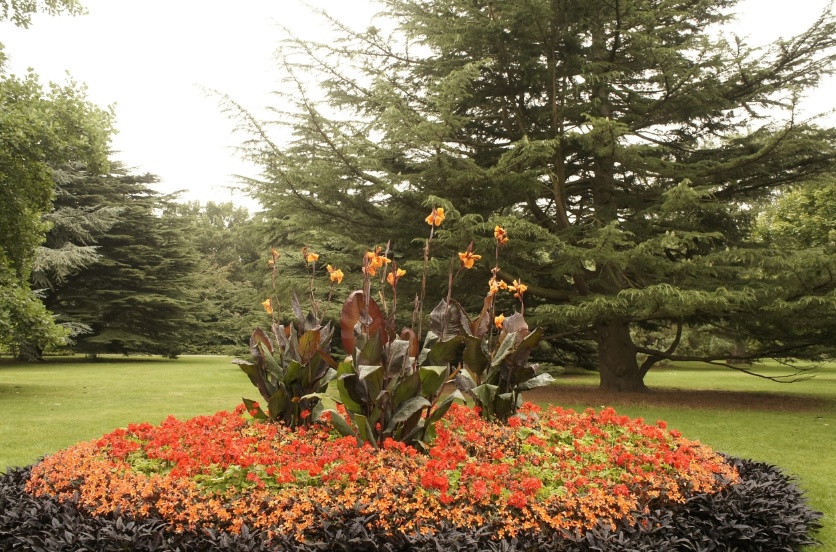}\\
    \includegraphics[width=\ourwidth]{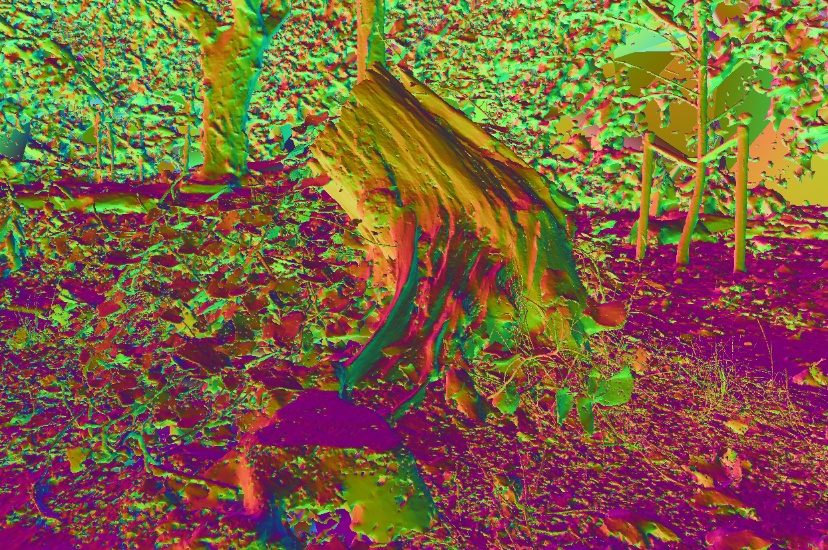}&
    \includegraphics[width=\ourwidth]{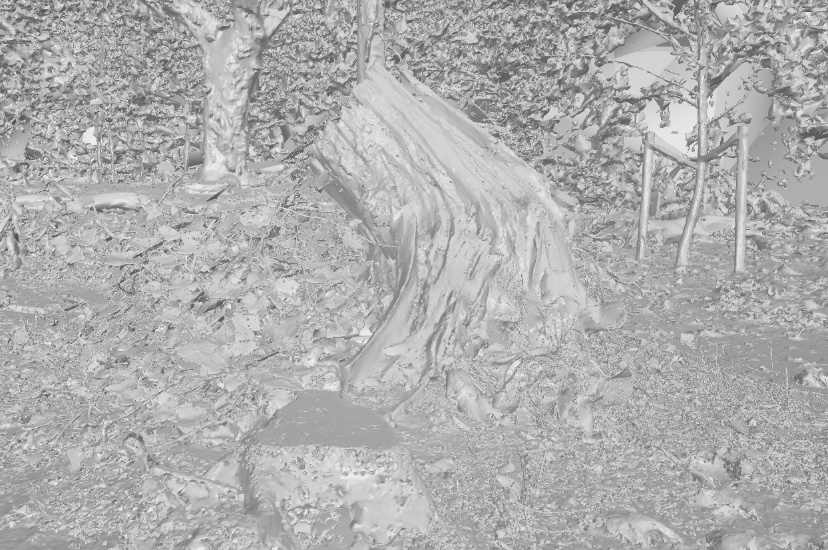}&
    \includegraphics[width=\ourwidth]{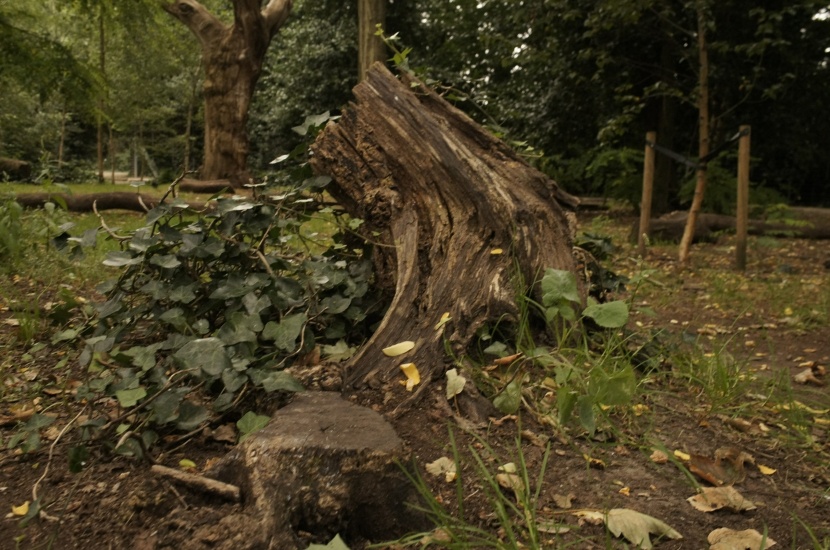}\\
    \includegraphics[width=\ourwidth]{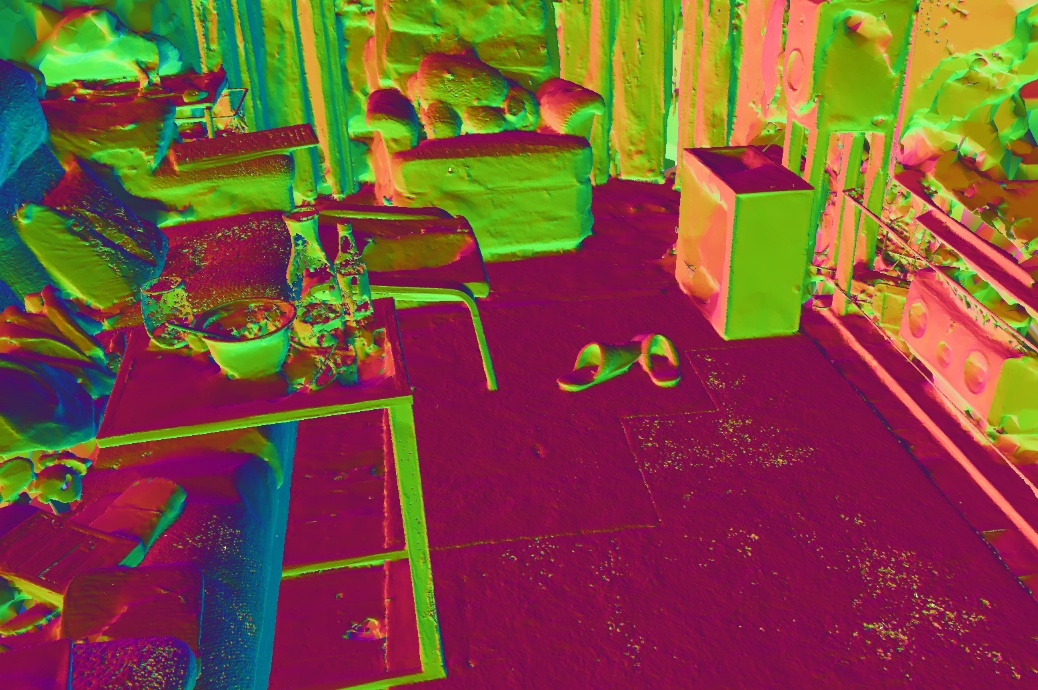}&
    \includegraphics[width=\ourwidth]{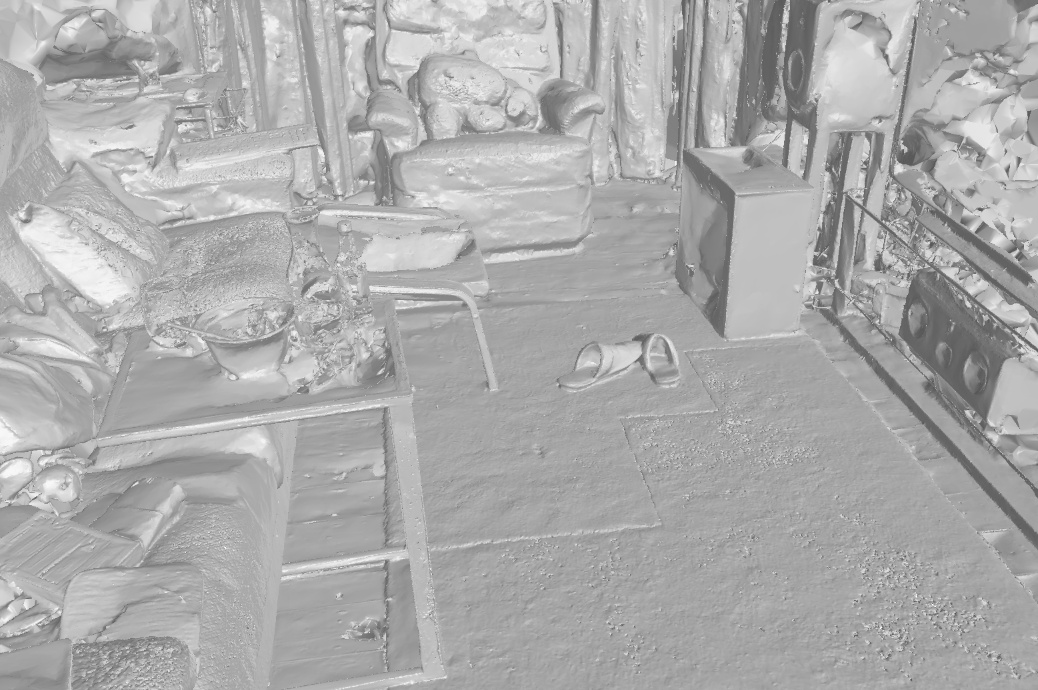}&
    \includegraphics[width=\ourwidth]{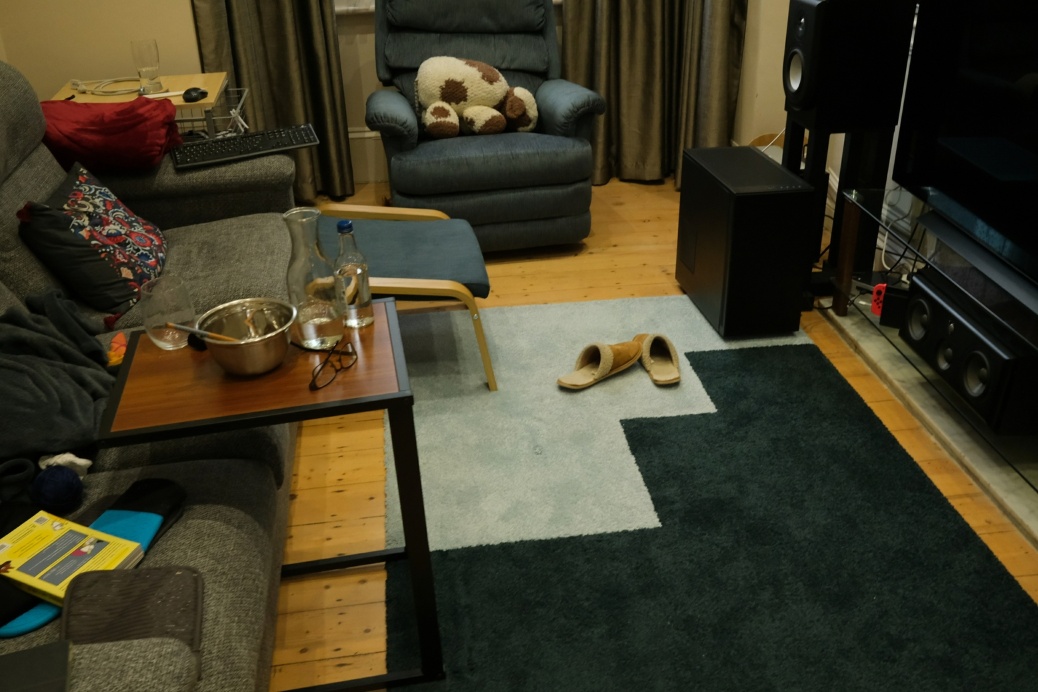}\\
    \includegraphics[width=\ourwidth]{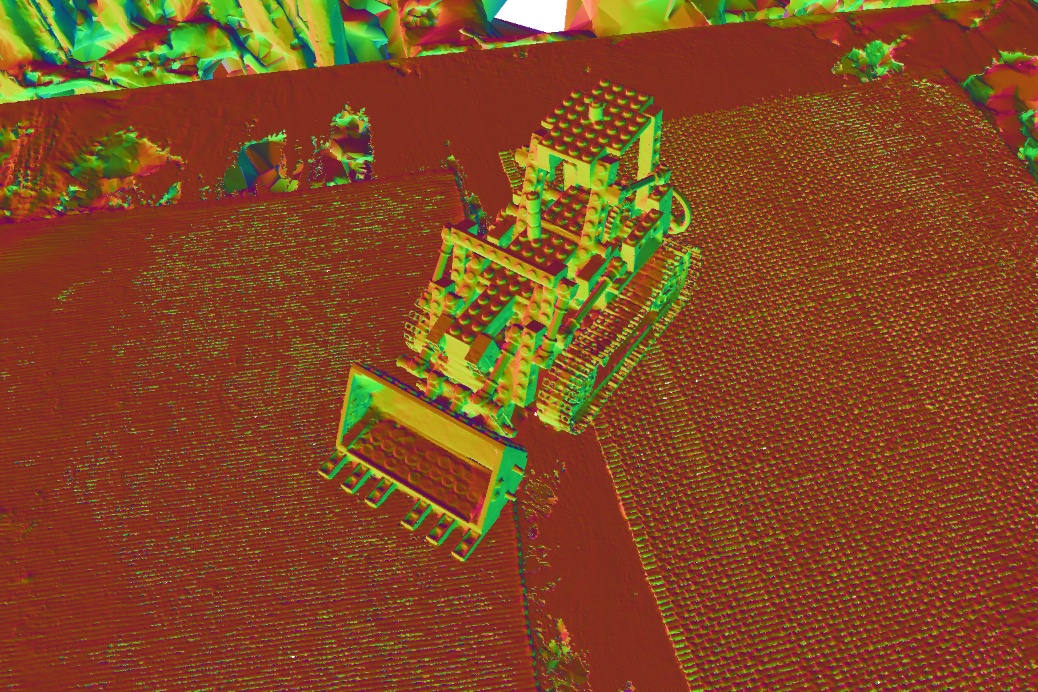}&
    \includegraphics[width=\ourwidth]{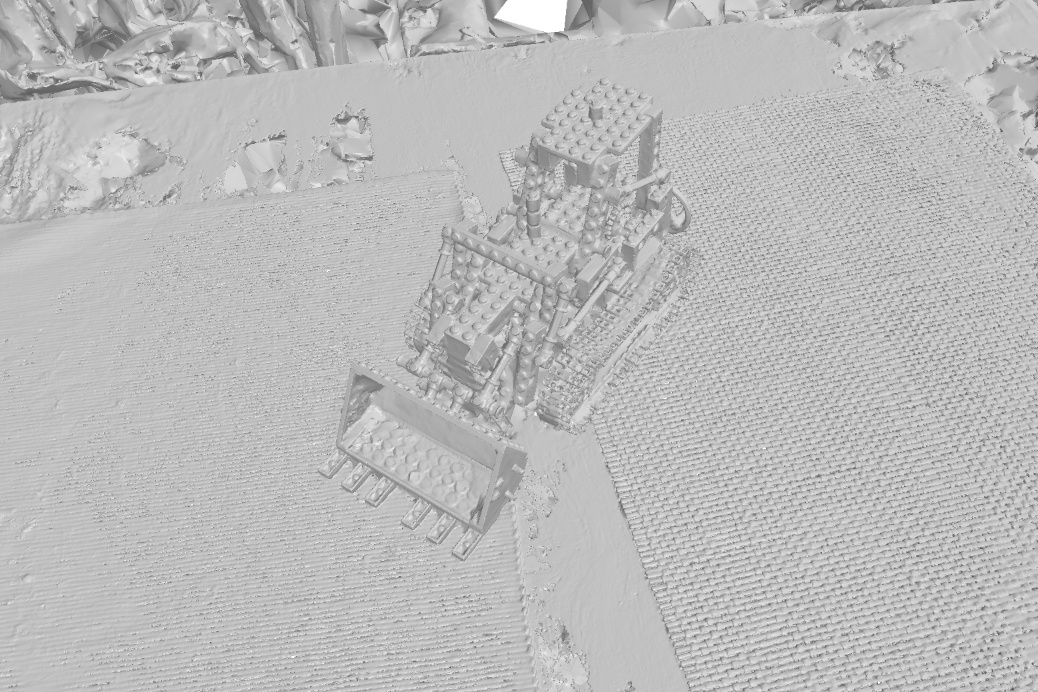}&
    \includegraphics[width=\ourwidth]{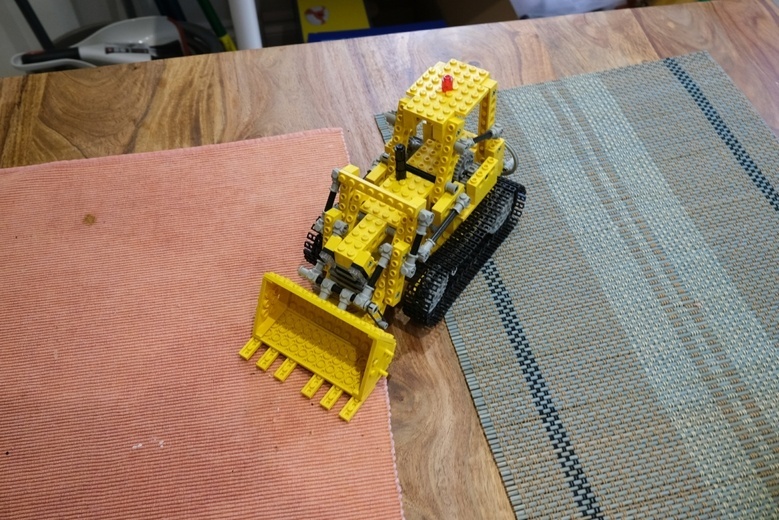}\\
    \includegraphics[width=\ourwidth]{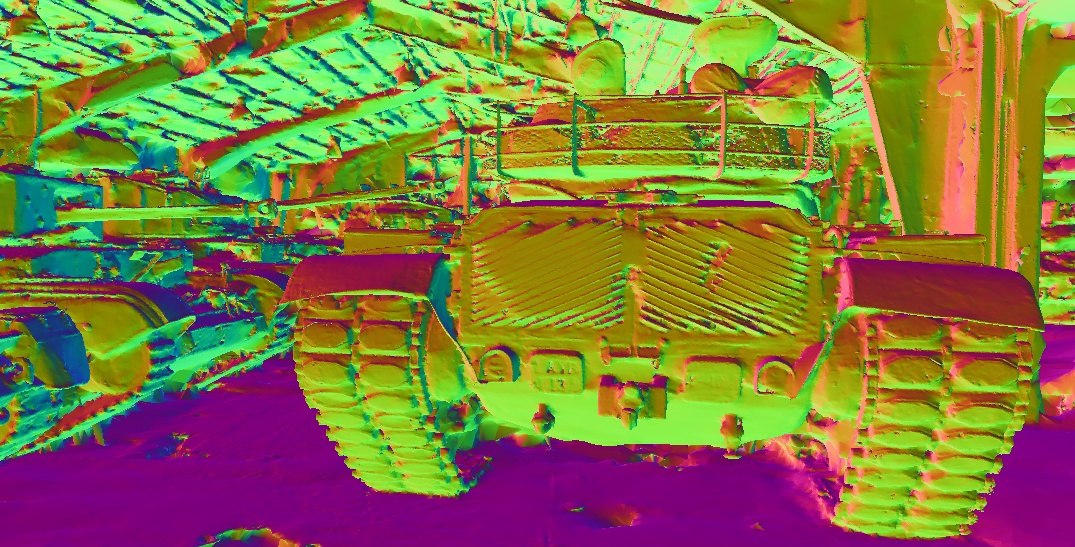}&
    \includegraphics[width=\ourwidth]{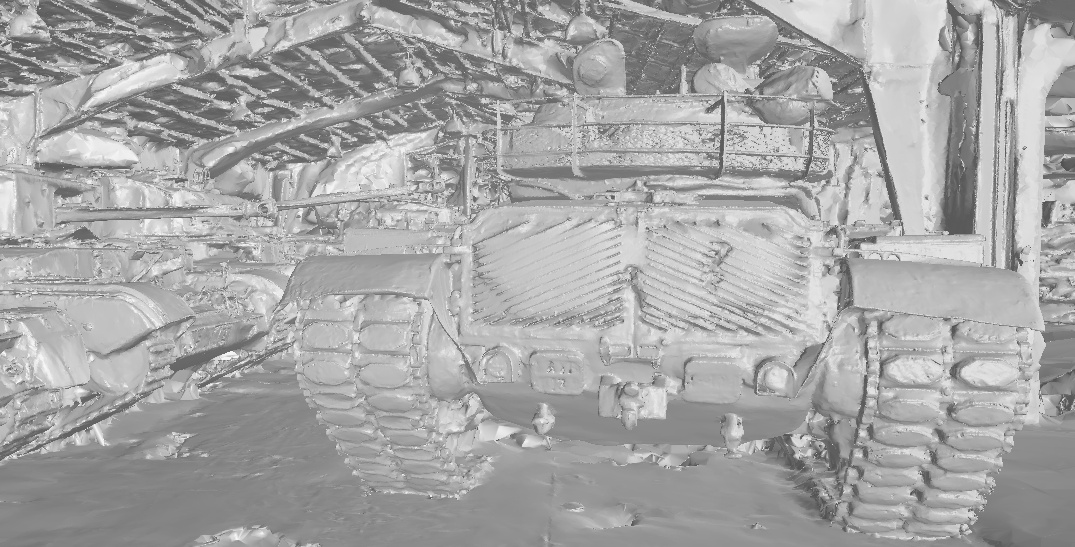}&
    \includegraphics[width=\ourwidth]{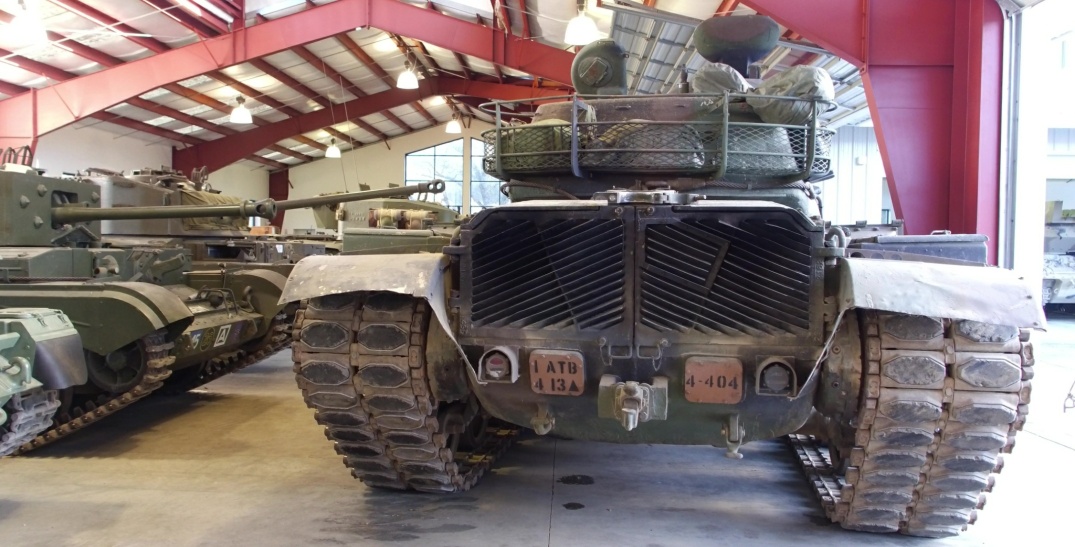}\\
    \includegraphics[width=\ourwidth]{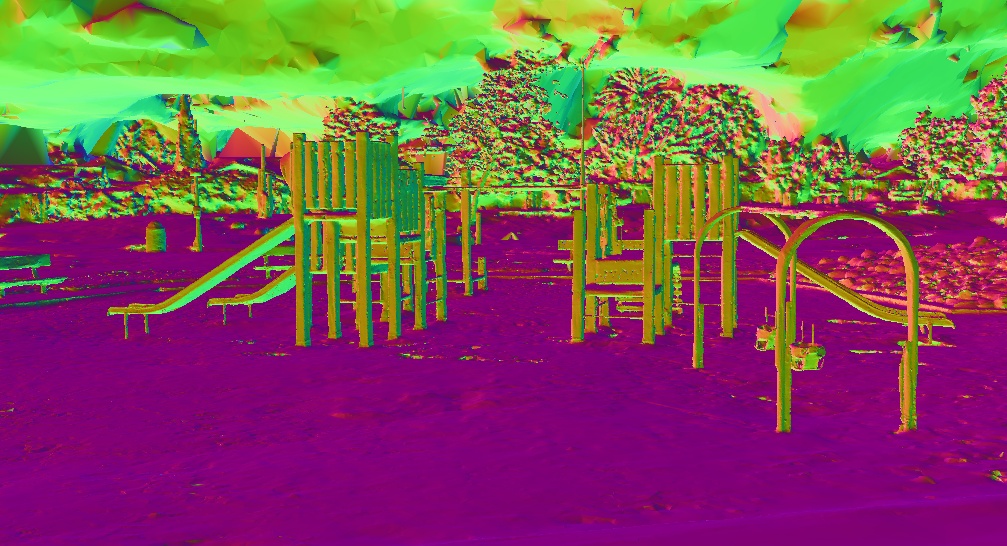}&
    \includegraphics[width=\ourwidth]{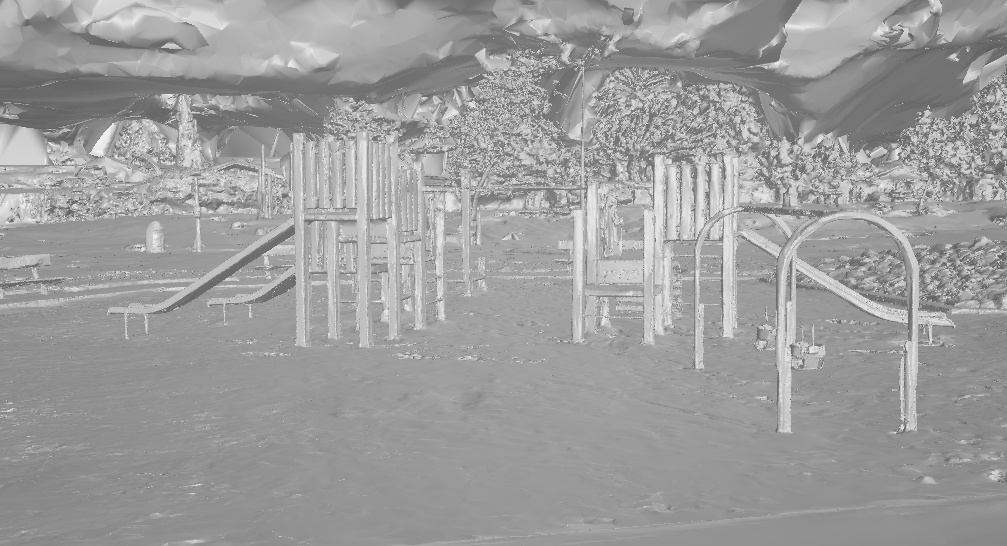}&
    \includegraphics[width=\ourwidth]{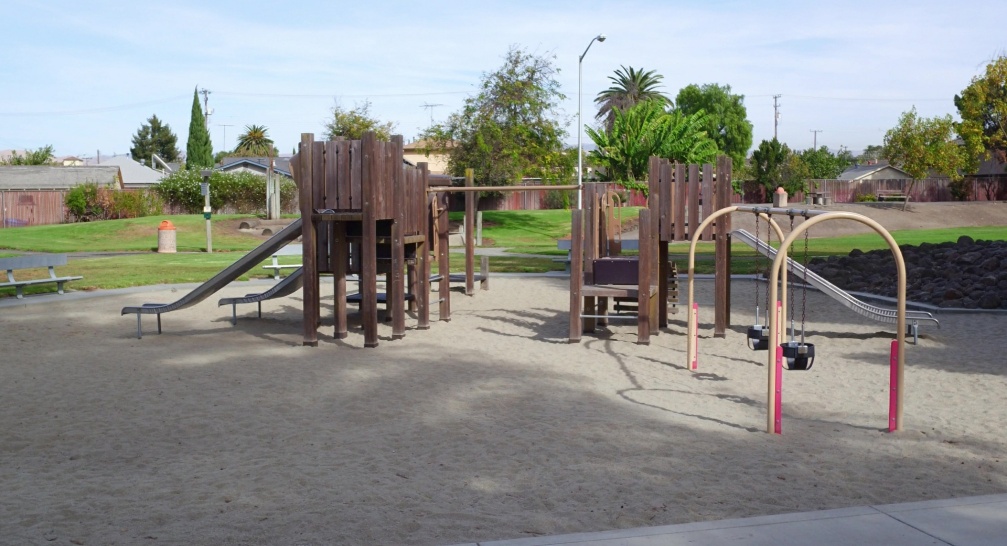}\\
    Normal & Mesh &  GT
    \end{tabular}
    \vspace{-0.15in}
    \caption{\textbf{Reconstructions on the Mip-NeRF 360~\cite{barron2022mipnerf360} and the Tanks and Temples dataset~\cite{Knapitsch2017}}. }
    \label{fig:our_one}
    \vspace{-0.1cm}
\end{figure*}

\end{document}